\documentclass{article}

\usepackage[preprint]{neurips_2023}

\usepackage[utf8]{inputenc} %
\usepackage[T1]{fontenc}    %

\usepackage[dvipsnames,svgnames]{xcolor}		%
\definecolor{MyRef}{HTML}{74787c}     %

\definecolor{lightblue}{HTML}{9A8F97}
\definecolor{darkblue}{HTML}{1A254B}
\definecolor{blue}{HTML}{2B50AA}
\definecolor{CustomRed}{HTML}{cb1338}

\definecolor{DarkBlue}{HTML}{1A254B}
\definecolor{DarkRed}{HTML}{A4243B}
\colorlet{MyRed}{blue}
\colorlet{MyBlue}{CustomRed}

\usepackage{hyperref}
\hypersetup{
colorlinks=true,
linktocpage=true,
pdfstartview=FitH,
breaklinks=true,
pdfpagemode=UseNone,
pageanchor=true,
pdfpagemode=UseOutlines,
plainpages=false,
bookmarksnumbered,
bookmarksopen=false,
bookmarksopenlevel=1,
hypertexnames=true,
pdfhighlight=/O,
urlcolor=MyRed,linkcolor=MyRef,citecolor=MyRef,	%
pdftitle={},
pdfauthor={},
pdfsubject={},
pdfkeywords={},
pdfcreator={pdfLaTeX},
pdfproducer={LaTeX with hyperref}
}

\usepackage{amsmath}
\usepackage{amssymb}
\usepackage{mathtools}
\usepackage{amsthm}
\usepackage{upgreek}
\usepackage{algpseudocodex,algorithm}
\usepackage[sort&compress,capitalize,nameinlink]{cleveref}
\usepackage{autonum}

\usepackage{url}            %
\usepackage{booktabs}       %
\usepackage{amsfonts}       %
\usepackage{nicefrac}       %
\usepackage{microtype}      %
\usepackage{xcolor}         %
\usepackage[inline]{enumitem}
\usepackage{wrapfig}
\usepackage{subcaption}

\theoremstyle{plain}
\newtheorem{theorem}{Theorem}[section]
\newtheorem{proposition}[theorem]{Proposition}
\newtheorem{lemma}[theorem]{Lemma}

\theoremstyle{definition}
\newtheorem{definition}[theorem]{Definition}
\newtheorem{assumption}[theorem]{Assumption}
\theoremstyle{remark}

\newcommand{\expeLigne}[1]{\mathbb{E}[#1]}
\newcommand{\CPELigne}[2]{\mathbb{E}[#1 \ | \ #2]}
\def \rmc {\mathrm{C}}
\def \msu {\mathsf{U}}
\def \msv {\mathsf{V}}
\def \msa {\mathsf{A}}
\def \msf {\mathsf{F}}
\def \mse {\mathsf{E}}
\def \rmD {\mathrm{D}}
\def \bfM {\mathbf{M}}
\def \mtt {\mathtt{m}}
\def \Ltt {\mathtt{L}}
\def \calA {\mathcal{A}}
\def \calB {\mathcal{B}}
\def \calR {\mathcal{R}}
\def\rset{\mathbb{R}}
\def\rmd{\mathrm{d}}
\def\bfX{\mathbf{X}}
\def\bfY{\mathbf{Y}}
\def\bfB{\mathbf{B}}

\def\Kker{\mathrm{K}}
\def\Pker{\mathrm{P}}
\def\rme{\mathrm{e}}
\def\dom{\mathcal{D}}
\newcommand{\abs}[1]{|#1|}
\newcommand{\mcb}[1]{\mathcal{B}(#1)}

\newcommand{\cl}[1]{\mathrm{cl}(#1)}
\newcommand{\inte}[1]{\mathrm{int}(#1)}
\newcommand{\ccint}[1]{[#1]}
\newcommand{\ocint}[1]{(#1]}
\newcommand{\coint}[1]{(#1]}
\newcommand{\ooint}[1]{(#1)}
\definecolor{airforceblue}{rgb}{0.36, 0.54, 0.66}
\definecolor{thistle}{rgb}{0.85, 0.75, 0.85}
\definecolor{ticklemepink}{rgb}{0.99, 0.54, 0.67}
\definecolor{thulianpink}{rgb}{0.67, 0.24, 0.43}
\definecolor{tealblue}{rgb}{0.11, 0.36, 0.43}

\newcommand{\ensembleLigne}[2]{\{#1\,:\; #2\}}
\newcommand{\cball}[2]{\bar{\mathrm{B}}(#1 \, ; #2)}
\newcommand{\ball}[2]{\mathrm{B}(#1 \, ; #2)}
\newcommand{\normLigne}[1]{\|#1\|}
\def \vareps {\varepsilon}
\def \Pbb {\mathbb{P}}
\def \nset {\mathbb{N}}
\def \Rbb {\mathbb{R}}
\def \Jbb {\mathbb{J}}

\usepackage{xspace}		%

\usepackage{acronym}		%
\newcommand{\aclip}[1]{\textit{\aclp{#1}}}		%
\newcommand{\acdef}[1]{\textit{\acl{#1}} \textup{(\acs{#1})}\acused{#1}}		%
\newcommand{\acdefp}[1]{\textit{\aclp{#1}} \textup{(\acsp{#1})}\acused{#1}}	%

\newacro{SB}{Schr\"odinger bridge}
\newacro{DSB}{diffusion Schr\"odinger bridge}
\newacro{USB}{unbalanced Schr\"odinger bridge}
\newacro{IPF}{iterative proportional fitting}
\newacro{SDE}{stochastic differential equation}
\newacro{sb}{Schr\"odinger bridge}
\newacro{KDE}{Kernel Density Estimation}

\title{Unbalanced Diffusion Schr\"odinger Bridge}

\author{%
  Matteo Pariset\thanks{Correspondence to \href{mailto:mpariset@ethz.ch}{mpariset@ethz.ch}.} \\
  ETH Zurich \\
  \And
  Ya-Ping Hsieh \\
  ETH Zurich \\
  \And
  Charlotte Bunne \\
  ETH Zurich \\
  \AND
  Andreas Krause \\
  ETH Zurich \\
  \And
  Valentin De Bortoli \\
  ENS Ulm \\
}

\begin{document}

\maketitle

\begin{abstract}
\acdefp{SB} provide an elegant framework for modeling the temporal evolution of populations in physical, chemical, or biological systems.
Such natural processes are commonly subject to changes in population size over time due to the emergence of new species or birth and death events.
However, existing neural parameterizations of \acp{SB} such as \acdefp{DSB}
are restricted to settings in which the endpoints of the stochastic process are both \emph{probability measures} and assume \emph{conservation of mass} constraints.  
To address this limitation, we introduce \emph{unbalanced} \acp{DSB} which model the temporal evolution of marginals with arbitrary finite mass.
This is achieved by deriving the time reversal of \aclip{SDE} with killing and birth terms. We present two novel algorithmic schemes that comprise a scalable objective function for training unbalanced \acp{DSB} and provide a theoretical analysis alongside challenging applications on predicting heterogeneous molecular single-cell responses to various cancer drugs and simulating the emergence and spread of new viral variants.
\end{abstract}

\section{Introduction}

Modeling the evolution of distributions is a fundamental principle that finds application in various domains such as natural sciences \citep{bunne2021learning, schiebinger2019optimal}, signal processing \citep{kolouri2017optimal}, and economics \citep{galichon2018optimal}.
These fields rely on capturing the collective dynamics of particles and face inherent challenges due to the limitations of continuous monitoring.
Observations are only typically restricted to \emph{discrete} time points, making it difficult to track individual particles along their trajectories.
For instance, in biology, assessing cellular responses to cancer drugs typically involves killing the cells, resulting in \textit{unpaired} measurements of cell populations before and after treatment. Similarly, reconstructing the dynamics of infectious diseases presents difficulties in tracing the origin and complete infection history of new viral variants.

Optimal transport (OT) \citep{santambrogio2015optimal, villani2009optimal} has recently emerged as a central tool to reconstruct population dynamics from periodic snapshots, by providing an actionable way of connecting these unpaired measurements.
The fundamental objective of OT is to flow a distribution ($\mu_0$) into another ($\mu_1$) while minimizing a user-designed cost.
In its static form, its solution yields a (stochastic) \emph{map} between two marginals, which describes the transition from the initial to the final one \citep{cuturi2013sinkhorn, schiebinger2019optimal}.

However, in many cases, the primary interest lies in accurately describing the \emph{continuous} dynamics of particles over time \citep{bunne2022proximal, tong2020trajectorynet, bunne2022recovering}.
The \emph{dynamic} OT problem thus aims at estimating a \emph{path measure}, i.e., a probability measure over the space of particle trajectories, rather than a static map between initial and final observations. This formulation, also known as the \acdef{SB} problem \citep{schrodinger1932theorie}, seeks a path measure that minimizes a specified cost function while respecting fixed initial and terminal distributions. To solve the \ac{SB} problem, recent advancements in diffusion models \citep{song2020score, ho2020denoising} have led to the emergence of powerful algorithmic approximations called \acdefp{DSB} \citep{de2021diffusion, chen2021likelihood, vargas2021solving}.

When reconstructing temporal dynamics, traditional \acp{DSB} assume that both extreme marginals are \textit{normalized} measures on the state-space. In this work, we relax this requirement by considering marginals with \textit{arbitrary} mass and formulate the \textit{unbalanced} SB problem. This increased flexibility is particularly relevant in biological applications, as living entities can experience death or proliferation between successive observations. For example, the occurrence and characteristics of cell deaths and births play a crucial role in understanding cellular responses to drugs.

In order to incorporate time-varying mass into the traditional \ac{DSB} methodology, we employ a generalized approach inspired by the work of \citet{chen2022most}. The main novelty of our work lies in the development of \emph{time-reversal} formulas for \acp{SDE} with \emph{birth and death} mechanisms. This can be seen as a rigorous extension of \ac{DSB} from the Euclidean space $\rset^d$ to its \emph{one-point compactification}, denoted as $\rset^d \cup \{\infty\}$, in which the added point ${\infty}$ serves as a "cemetery" or "coffin state."

To summarize, we make the following contributions:
\begin{enumerate*}[label=(\alph*)]
\item We derive the time-reversal of diffusions with killing terms and show they
  correspond to diffusions with \emph{birth}. Analogously, we show that time-reversing diffusions with birth yields diffusions with \emph{death}.
\item We leverage these formulas and associated loss functions to propose a framework for solving the \ac{DSB} on $\rset^d \cup \{\infty\}$.
\item We design scalable algorithms to approximately tackle these \emph{unbalanced} \acp{DSB} problems with both deaths and births.
\item We assess the performance of our algorithms on various tasks, with a particular focus on enhancing the modeling of cellular responses to cancer drugs. Notably, our experiments demonstrate a significant impact of including information on deaths and births occurring at intermediate times.
\end{enumerate*}

\begin{figure*}[t]
    \centering
    \includegraphics[width=0.9\textwidth]{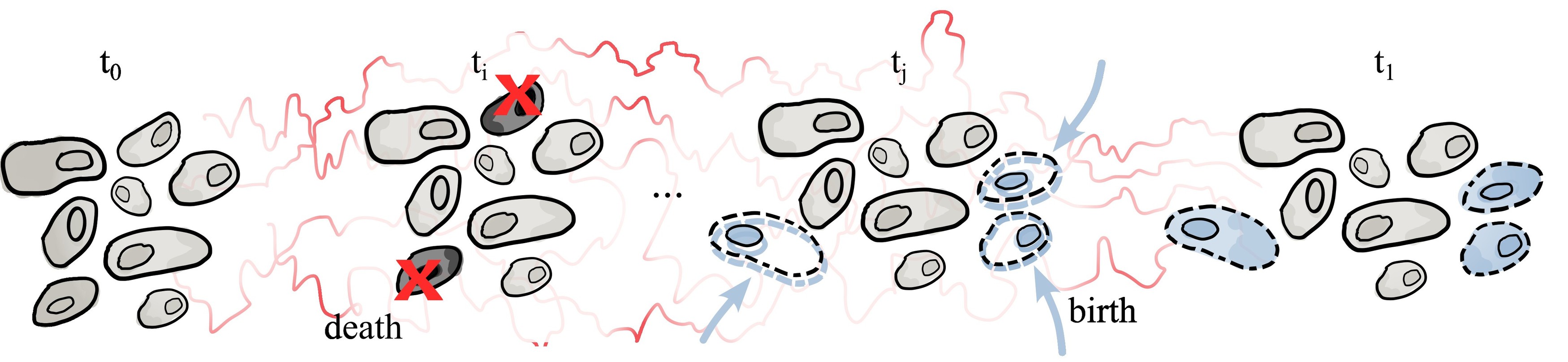}
    \caption{\emph{Unbalanced} Schr\"odinger bridge. The time evolution of a cell population encompasses alterations in transcriptomic profiles as well as natural and externally-induced death and birth events.}
    \label{fig:results_cell_class}
\end{figure*}

\section{Background}
\label{sec:background}

\def \Qbb {\mathbb{Q}}
\def \argmin {\mathrm{argmin}}
\def \KL {\mathrm{KL}}
\def \SB {\mathrm{SB}}

\paragraph{Schr\"odinger Bridge and unbalanced transport.}
The Schr\"odinger Bridge (SB) problem can be seen as the \emph{dynamical}
counterpart of the static entropy regularized OT problem. More precisely, the
Schr\"odinger Bridge $\Pbb^\SB$ is defined by
\begin{equation}
\textstyle
  \label{eq:dynamic_schrodinger_bridge}
  \Pbb^\SB = \argmin \ensembleLigne{\KL(\Pbb | \Pbb^0)}{\Pbb_0 = \mu_0, \ \Pbb_T = \mu_1} ,
\end{equation}
where $\mu_0, \mu_1$ are probability measures and $\Pbb^0$ is a reference
probability path measure, i.e.~a probability measure on the space of continuous
functions $\rmc(\ccint{0,T}, \rset^d)$. In the case where $\Pbb^0$ is given by a
Brownian motion, it can be shown that $\Pbb^\SB_{0,T}$ is the coupling between
$\mu_0$ and $\mu_1$ for the quadratic Wasserstein distance with $T^{-1}$ entropy
regularization \citep{mikami2002optimal}. The solutions of
\eqref{eq:dynamic_schrodinger_bridge} are generally intractable and one has to
resort to numerical schemes. One popular method to approximate $\Pbb^\SB$ is
the \ac{IPF} algorithm
\citep{sinkhorn1967concerning,knight2008sinkhorn,peyre2019computational,cuturi2014fast}
which defines processes $(\Pbb^n)_{n \in \nset}$ such that, for any $n \in \nset$,
\begin{equation}
  \label{eq:ipf_iteration} 
  \Pbb^{2n+1} = \argmin \ensembleLigne{\KL(\Pbb|\Pbb^{2n})}{\Pbb_T = \mu_1} , \quad \Pbb^{2n+2} = \argmin \ensembleLigne{\KL(\Pbb|\Pbb^{2n+1})}{\Pbb_0 = \mu_0} .
\end{equation}
Under mild assumptions, $\lim_{n \to +\infty} \Pbb^n = \Pbb^\SB$.  While
optimal transport deals with couplings between two (or more) \emph{probability}
measures, unbalanced optimal transport removes this \emph{conservation of mass}
constraint and tackles the problem of coupling \emph{finite} measures with
possibly different mass. These unbalanced extensions can be broadly classified
into two
families. Previous work
investigates relaxations of the classical OT formula where the \emph{hard}
marginal constraints are replaced by \emph{soft} ones \citep{chizat2018interpolating,liero2018optimal,yang2018scalable,kondratyev2016new}. Another line of work
extends the finite measures to measures of the same mass by
adding a \emph{cemetery} (or \emph{coffin}) state $\{\infty\}$ and performing
classical optimal transport on this extended space
\citep{pele2009fast,caffarelli2010free,gramfort2015fast,ekeland2010existence}.
\citet{caffarelli2010free} show the equivalence between a \emph{partial} OT
problem and the OT problem on an extended space with a coffin state which accounts for the loss of mass. We follow this line of work and denote the extended state-space as $\hat{\rset}^d = \rset^d \cup \{\infty\}$. Finally, some of the theoretical
properties of the solutions of Schr\"odinger bridge problems on $\hat{\rset}^d$ are studied by
\cite{chen2022most}.

\paragraph{Diffusion Schr\"odinger bridges.} 
In order to numerically approximate the solutions of
\eqref{eq:dynamic_schrodinger_bridge}, \citet{de2021diffusion} introduced
\acp{DSB}, which leverage recent advances in
diffusion models \citep{song2020score,ho2020denoising} to solve \ac{IPF} iterations
\eqref{eq:ipf_iteration}.  The procedure alternates between a projection on the
space of path-measures with either correct initial distribution or correct terminal
distribution. %
Assuming that $\Pbb^0$ is associated with the \ac{SDE}
$\rmd \bfX_t = f_t(\bfX_t) \rmd t + \rmd \bfB_t$, then $\Pbb^1$ is associated
with the \emph{time-reversal}
\citep{haussmann1986time,anderson1982reverse,cattiaux2021time} of this \ac{SDE}, initialized at $\mu_1$.  More precisely, we have
$(\bfY_{T-t})_{t \in \ccint{0,T}} \sim \Pbb^1$ with
\begin{equation}
  \rmd \bfY_t = \{-f_{T-t}(\bfY_t) + \nabla \log p_{T-t}(\bfY_t) \} \rmd t + \rmd \bfB_t , \qquad \bfY_0 \sim \mu_1 
\end{equation}
where $p_t$ is the density of $\Pbb^0_t$ and $(\bfB_t)_{t \geq 0}$ a
$d$-dimensional Brownian motion. The quantity $\nabla \log p_t$ is called the
\emph{Stein score} and can be estimated using score-matching techniques
\citep{hyvarinen2005estimation,vincent2011connection}. %
While the first iteration of \ac{DSB} can be understood as training a diffusion
model, subsequent iterations of \ac{DSB} go beyond that by alternating between the optimization of the forward and backward
processes.

\def\calK{\mathcal{K}}
\def\calB{\mathcal{B}}

\section{Time-Reversal of Birth and Death Processes}

To numerically address the \acl{USB} problem, it is crucial to derive formulas
for the \ac{IPF} procedure which can be easily discretized. In particular,
following the framework of \cite{de2021diffusion}, we need to derive
time-reversal formulas. %
We present a succinct overview of the core concepts and leave a thorough discussion to \crefrange{app:sec:notation}{app:sec:unbalanced_ipf}.%

\paragraph{Diffusion processes with killing.} 
A diffusion process with \emph{killing} can be intuitively characterized in terms of its \ac{SDE} representation:
\begin{equation}
\label{eq:diffusion}
   \rmd\bfX_t = b(\bfX_t) \rmd t + \rmd\bfB_t,
 \end{equation}
 (where $b: \rset^d \to \rset^d$ is the drift) by defining the \emph{killing rate}
 $k(x)\geq 0$ as $\textstyle \mathbb{P}\left( \bfX_t \textup{ dies in } [t, t+h] \right) =  k(\bfX_t)h + o(h).$
 Rigorous definitions can be obtained using either \emph{Feynman-Kac semigroups} (see \Cref{app:sec:time-reversal-killed}) or \emph{infinitesimal generators}, but we choose the latter
 approach, since it simplifies the study of birth processes. To this
 end, we consider the \emph{one-point compactification} of $\rset^d$, denoted
 $\hat{\rset}^d = \rset^d \cup \{ \infty \}$, where $\infty$ denotes the
 \emph{cemetery} state after a particle is killed. For a smooth
 function $f$ on $\hat{\rset}^d$, we define an operator $\hat{\calK}$ as
\begin{equation}
  \label{eq:inf_gen_extended_rd_main}
 \hat{\calK}(f)(x) = \left[\langle b(x), \nabla f(x) \rangle + \tfrac{1}{2} \Delta f(x) - k(x)(f(x) - f(\infty))\right] \,\mathbf{1}_{\rset^d}(x) . 
\end{equation}%
It can be shown that \eqref{eq:inf_gen_extended_rd_main} is the
\emph{infinitesimal generator} associated with the Feynman-Kac semigroup with
killing rate $k$, i.e., $\hat{\calK}(f) = \lim_{t \to 0} (1/t) (\CPELigne{f(\bfX_t)}{\bfX_0=x}-f(x))$ for a process
$(\bfX_t)_{t \geq 0}$ which is a diffusion with killing on $\hat{\rset}^d$. A rigorous treatment of
such processes is provided in \cref{prop:feynm-kac-semigr}.

  \paragraph{Time-reversal of diffusions with killing.}
  We are now ready to
  derive our main result: a time-reversal formula for diffusion processes with
  killing. %
Let $(\bfX_t)_{t\in[0,T]}$ be the diffusion process with killing rate $k$, defined by the generator \eqref{eq:inf_gen_extended_rd_main}, and $(\bfY_t)_{t \in \ccint{0,T}} \coloneqq (\bfX_{T-t})_{t \in \ccint{0,T}}$ be its \emph{time reversal}.

\begin{proposition}[Time reversal; informal]
    \label{prop:time_reversal_kill_informal}
    Under mild assumptions, the generator $\hat{\calB}$ of the time-reversed
    process $(\bfY_t)_{t \in \ccint{0,T}}$ is given for any sufficiently smooth
    function $f$ and any $x \in \hat{\rset}^d$ by 
    \begin{align}
      \label{eq:time_reversal_kill_informal}
      \textstyle \hat{\calB}(f)(t,x) &= [\langle -b(x) + \textcolor{MyRed}{\nabla \log p_{T-t}(x)}, \nabla f(x) \rangle + \tfrac{1}{2} \Delta f(x)] \;\mathbf{1}_{\rset^d}(x) \\
      & \qquad \textstyle + \int_{\rset^d} (\textcolor{MyRed}{p_{T-t}(\tilde{x})/ S_{T-t}}) k(\tilde{x})  (f(\tilde{x}) - f(\infty)) \rmd \tilde{x} \;\mathbf{1}_{\infty}(x). 
    \end{align}
    Here, $p_t$ is the probability density of $\bfX_t$ and $S_t \coloneqq \mathbb{P}[\bfX_t=\infty]$. %
  \end{proposition}
  The key terms appearing in \eqref{eq:time_reversal_kill_informal} are highlighted in \textcolor{MyRed}{blue}. Note that, if there is no killing in $\bfX_t$, i.e.,
  $k(x) \equiv 0$, then $\hat{\calB}(f)(t,x)$ reduces to the well-known
  time-reversal formula for Euclidean diffusions
  \citep{haussmann1986time,cattiaux2021time}. In the presence of killing, instead,
  \Cref{prop:time_reversal_kill_informal} leads to the following important
  observation: %
  \emph{The time-reversal of a diffusion process with killing is a diffusion
    process with \textbf{birth}.}
More precisely, the second line of \eqref{eq:time_reversal_kill_informal} suggests that the killing rate $x \mapsto k(x)$ is turned into a birth rate $x \mapsto k(x) p_t(x) / S_t$. %
As a consequence, a region with a high probability of birth in the backward process is one in which:
\begin{enumerate*}[label=(\alph*)] 
\item the killing rate is large, i.e.,~many particles die in the forward process;
\item the density $p_t$ is large, i.e.,~it is visited often by $\bfX_t$;
\item $S_t$ is small, i.e.,~most of the mass in the forward process is located somewhere in $\rset^d$ at time $t$.
\end{enumerate*}

\paragraph{Time-reversal of diffusions with birth.} 
So far, we have established that the time reversal of a diffusion process with killing is, itself, a diffusion process with birth. However, the \ac{IPF} procedure also requires to time-reverse the latter, and we, therefore, devote the rest of this section to show that: 
\emph{The time-reversal of a diffusion process with birth is a diffusion process with \textbf{killing}.} %
Consider a diffusion process of the form \eqref{eq:diffusion} with \emph{birth rate} $q(x)\geq 0$. Similar to \eqref{eq:inf_gen_extended_rd_main}, one can define such a process via the generator $\hat{\calB}$ such that, for any
sufficiently smooth $f$ and $x \in \hat{\rset}^d$,
  \begin{equation}
  \label{eq:hold1}
    \textstyle \hat{\calB}(f)(x) = (\langle b(x), \nabla f(x) \rangle + \tfrac{1}{2}\Delta f(x)) \;\mathbf{1}_{\rset^d}(x) + \int_{\rset^d} (f(\tilde{x}) - f(\infty)) q(\tilde{x}) \mathrm{d}\tilde{x} \;\mathbf{1}_{\infty}(x) . 
  \end{equation}
Then, under mild assumptions and using techniques similar to those in \cref{prop:time_reversal_kill_informal}, one can show that the generator $\hat{\calK}$ associated with the time-reversed diffusion with birth in \eqref{eq:hold1} can be expressed as, 
  \begin{align}
    \label{eq:time_reversal_birth_main}
    \hat{\calK}(f)(t,x) &= (\langle -b(x) + \textcolor{MyBlue}{\nabla \log p_{T-t}(x)}, \nabla f(x) \rangle + \tfrac{1}{2}\Delta f(x))  \;\mathbf{1}_{\rset^d}(x) \\
    & \qquad - (\textcolor{MyBlue}{S_{T-t}/ p_{T-t}(x)}) q(x)  (f(x) - f(\infty)) \;\mathbf{1}_{\rset^d}(x) . 
  \end{align}
  for any sufficiently smooth function $f$ and $x \in \hat{\rset}^d$.
  The key terms in \eqref{eq:time_reversal_birth_main} are highlighted in
  \textcolor{MyBlue}{red}.  Comparing \eqref{eq:time_reversal_birth_main} with
  \eqref{eq:inf_gen_extended_rd_main}, we conclude that the time-reversal of a
  diffusion with birth rate $q(x)$ is a diffusion with killing rate
  $x \mapsto S_{T-t} q(x) /
  p_{T-t}(x)$. %

\newcommand{\diff}[1]{\ensuremath{\operatorname{d}\!{#1}}}

\newcommand{\ip}[2]{\left \langle #1,\, #2 \right \rangle}

\newcommand{\norm}[1]{\left\lVert #1 \right\rVert}

\newcommand{\vect}[1]{\mathbf{#1}}

\newcommand{\gradx}{\nabla_\vect{x}}

\newcommand{\laplx}{\Delta_\vect{x}}

\newcommand{\hessx}{\nabla^2_\vect{x}}

\newcommand{\divx}{\gradx \cdot}

\newcommand{\tr}[1]{\text{tr}\left( #1 \right)}

\newcommand{\prob}[1]{\mathbb{P}\left(#1\right)}
\newcommand{\probd}[2]{\mathbb{P}_{#1}\left(#2\right)}

\newcommand{\ev}[1]{\mathbb{E}\left[ #1\right]}
\newcommand{\evd}[2]{\mathbb{E}_{#1}\left[ #2\right]}

\newcommand{\generator}{L}

\newcommand{\kl}[2]{D_{\text{KL}}\left( #1 \| #2\right)}

\newcommand{\indfunc}[1]{\mathbb{1}_{#1}}

\newcommand{\broadcast}[3]{{
    \IfEqCase{#1}{
        {F}{#2}
        {B}{#3}
    }[\PackageError{tree}{Unknown direction: #1}{}]
}}

\newcommand{\drift}{b}
\newcommand{\diffusivity}{\sigma}

\newcommand{\refproc}{\mathcal{R}}
\newcommand{\sbproc}{p}

\newcommand{\marg}[1]{\broadcast{#1}{\mu_0}{\mu_1}}

\newcommand{\rew}[1]{\Tilde{#1}}
\newcommand{\rmarg}[1]{\broadcast{#1}{\Tilde{\pi}_0}{\Tilde{\pi}_1}}

\newcommand{\densfact}[1]{\broadcast{#1}{\varphi}{\hat{\varphi}}}

\newcommand{\discrfact}[1]{\broadcast{#1}{\Psi}{\hat{\Psi}}}

\newcommand{\X}[1]{\broadcast{#1}{X}{\hat{X}}}

\newcommand{\Y}[1]{\broadcast{#1}{Y}{\hat{Y}}}

\newcommand{\Ynn}[1]{\Y{#1}^\params{#1}}
\newcommand{\sfYnn}{Y^\theta}
\newcommand{\sbYnn}{\hat{Y}^{\hat\theta}}

\newcommand{\Z}[1]{\broadcast{#1}{Z}{\hat{Z}}}
\newcommand{\sfZnn}{Z^\theta}
\newcommand{\sbZnn}{\hat{Z}^{\hat\theta}}

\newcommand{\Znn}[1]{\Z{#1}^\params{#1}}

\newcommand{\Qnn}{Q^\params{F}}
\newcommand{\Nnn}{N^\params{B}}

\newcommand{\params}[1]{\broadcast{#1}{\theta}{\hat\theta}}

\newcommand{\ipflossd}[1]{\mathcal{L}^\params{#1}_{\text{IPF}}}

\newcommand{\tdloss}{\mathcal{L}_{\text{TD}}}

\newcommand{\fvers}[1]{\overset{\rightarrow}{#1}}
\newcommand{\bvers}[1]{\overset{\leftarrow}{#1}}

\newcommand{\gen}{L}
\newcommand{\gensb}{S}
\newcommand{\dgen}[2]{\broadcast{#1}{\overset{\rightarrow}{\gen}}{\overset{\leftarrow}{\gen}}\left[ #2\right]}

\newcommand{\jump}{J}
\newcommand{\djump}[1]{\broadcast{#1}{\overset{\rightarrow}{\jump}}{\overset{\leftarrow}{\jump}}}

\newcommand{\ipfbef}{\rho}
\newcommand{\ipfb}{\bar{\rho}}
\newcommand{\ipff}{\bar{\bar{\rho}}}

\newcommand{\ipfloss}{\mathcal{L}_\text{IPF}}

\newcommand{\usbalgo}{\textsc{UDSB-TD}}
\newcommand{\usbferryman}{\textsc{UDSB-F}}

\newcommand{\schrodinger}{Schr\"odinger}

\newcommand{\partflow}{\Tilde{c}}
\newcommand{\normpartflow}{c}

\newcommand{\killing}{k}
\newcommand{\birth}{q}

\newcommand{\ferryman}{g}
\newcommand{\ferrymanparams}{\zeta}
\newcommand{\ferrymannn}{\ferryman^\ferrymanparams}

\newcommand{\coffin}{\infty}

\newcommand{\dotp}[2]{\langle #1, #2 \rangle}

\section{Unbalanced Iterative Proportional Fitting}
\label{sec:algo}

Equipped with the notion of time reversal, we are now ready to derive an IPF scheme for marginals with arbitrary mass. First, we recall the basics of
unbalanced SBs and then present two algorithms to approximate the unbalanced IPF numerically. 

\paragraph{Unbalanced \acp{SB}.}
For simplicity, we assume that the target measures $\mu_0, \mu_1$ satisfy
$\mu_0(\rset^d) \geq \mu_1(\rset^d)$ and their restrictions to $\rset^d$ have
smooth densities $p_0, p_1$ w.r.t. the Lebesgue measure.  We start by recalling a result from \citet{chen2022most} which extends the
\emph{Schr\"odinger equations} to the unbalanced marginals and forward processes with killing. We let the process with infinitesimal generator $\hat{\calK}^0$ defined in \eqref{eq:inf_gen_extended_rd_main} be our prior motion $\Pbb^0$ and %
aim to solve \eqref{eq:dynamic_schrodinger_bridge}. Unlike in \Cref{sec:background}, though, we consider $\Pbb^{\mathrm{SB}}$ and $\Pbb^0$ to be path measures on the
\emph{extended} space $\hat{\rset}^d$. To state the solution to
\eqref{eq:dynamic_schrodinger_bridge} in this case, we define the functions
$(\textcolor{MyBlue}{\varphi}, \textcolor{MyRed}{\hat{\varphi}}, \textcolor{MyBlue}{\Psi}, \textcolor{MyRed}{\hat{\Psi}})$ which, for any $t \in \ccint{0,T}$
and $x \in \rset^d$, satisfy:
\begin{align}
  \label{eq:system_unrolled}
  \partial_t \textcolor{MyBlue}{\varphi_t(x)} &= -\langle b(x), \nabla \textcolor{MyBlue}{\varphi_t(x)} \rangle - \tfrac{1}{2}\Delta \textcolor{MyBlue}{\varphi_t(x)} + k(x) \textcolor{MyBlue}{\varphi_t(x)} - k(x) \textcolor{MyBlue}{\Psi_t}, &\partial_t \textcolor{MyBlue}{\Psi_t} &= 0 , \\
  \partial_t \textcolor{MyRed}{\hat{\varphi_t}(x)} &= -\mathrm{div}(b \textcolor{MyRed}{\hat{\varphi_t}})(x) + \tfrac{1}{2}\Delta \textcolor{MyRed}{\hat{\varphi_t}(x)} - k(x) \textcolor{MyRed}{\hat{\varphi_t}(x)}, & \partial_t   \textcolor{MyRed}{\hat{\Psi}_t} &= \textstyle\int_{\rset^d} k(x) \textcolor{MyRed}{\hat{\varphi_t}(x)} \rmd x.
\end{align}
together with the boundary conditions:
\[
    \textcolor{MyBlue}{\varphi_0} \textcolor{MyRed}{\hat{\varphi_0}} = p_0, \quad \textcolor{MyBlue}{\varphi_1} \textcolor{MyRed}{\hat{\varphi_1}} = p_1 \qquad \text{and} \qquad
    \textcolor{MyBlue}{\Psi_0} \textcolor{MyRed}{\hat{\Psi}_0} = 0, \quad \textcolor{MyBlue}{\Psi_1} \textcolor{MyRed}{\hat{\Psi}_1} = \mu_0(\rset^d) - \mu_1(\rset^d)
\]

\begin{proposition}
  \label{prop:unbalanced_sb}
  Under mild conditions, there exists a unique solution $\Pbb^\SB$ to
  \eqref{eq:dynamic_schrodinger_bridge} on $\hat{\rset}^d$. The paths
  $(\bfX_t)_{t \in \ccint{0,T}} \sim \Pbb^\SB$ are associated with the generator $\hat{\calK}^\SB$ given,
  for any smooth $f$, $t \in \ccint{0,T}$ and
  $x \in \hat{\rset}^d$, by
  \begin{equation}
    \label{eq:killing_optimal}
\textstyle    \hat{\mathcal{K}}^\SB_t(f)(x) = \left[\langle b(x) + \nabla \log \textcolor{MyBlue}{\varphi_t}(x), \nabla f(x) \rangle + \tfrac{1}{2} \Delta f(x) - (\textcolor{MyBlue}{\Psi_t/\varphi_t(x)})k(x) (f(x) - f(\infty))\right] \mathbf{1}_{\rset^d}(x).
  \end{equation}
  In addition, $(\bfX_{T-t})_{t \in \ccint{0,T}}$ is associated with
  $\hat{\calB}^\SB$, given for any smooth $f$ and $x \in \hat{\rset}^d$, by
  \begin{align}
    \label{eq:birth_optimal} 
    \hat{\calB}^\SB_t(f)(x) &= (\langle -b(x) + \nabla \log \textcolor{MyRed}{\hat{\varphi}_{T-t}(x)}, \nabla f(x) \rangle + \tfrac{1}{2}\Delta f(x))  \;\mathbf{1}_{\rset^d}(x) \\
    & \qquad \textstyle +  \int_{\rset^d} (\textcolor{MyRed}{\hat{\varphi}_{T-t}(\tilde{x})/\hat{\Psi}_{T-t}}) k(\tilde{x})  (f(\tilde{x}) - f(\infty)) \rmd \tilde{x} \;\mathbf{1}_{\infty}(x) . 
  \end{align}
\end{proposition}
To improve readability, we highlight the control terms $(\textcolor{MyBlue}{\varphi}, \textcolor{MyBlue}{\Psi})$
of the process with killing in \textcolor{MyBlue}{red} and those of the diffusion with birth in \textcolor{MyRed}{blue}.

The first part of the proof is a consequence of \citet[Theorem
9]{chen2022most}. The generator \eqref{eq:killing_optimal} is given by
\citet[Theorem 10]{chen2022most} while the generator \eqref{eq:birth_optimal} is
obtained using \cref{prop:time_reversal_kill_informal}.
We leave the discussion on the connections between \eqref{eq:birth_optimal} and
\eqref{eq:time_reversal_kill_informal}, and \eqref{eq:killing_optimal} and
\eqref{eq:time_reversal_birth_main} to \Cref{app:sec:time-reversal-killed} and comment, instead, on the interest of this result. Define $\textcolor{MyBlue}{\chi}, \textcolor{MyRed}{\hat{\chi}}: \ \ccint{0,T} \times \hat{\rset}^d \to \rset$, for any $t \in \ccint{0,T}$ and $x \in \hat{\rset}^d$, as
\begin{equation}
  \textcolor{MyBlue}{\chi_t(\infty)} = \textcolor{MyBlue}{\Psi_t} , \quad \textcolor{MyBlue}{\chi_t(x)} = \textcolor{MyBlue}{\varphi_t(x)} \qquad \text{and} \qquad \textcolor{MyRed}{\hat{\chi}_t(\infty)} =  \textcolor{MyRed}{\hat{\Psi}_t} , \quad \textcolor{MyRed}{\hat{\chi}_t(x)} = \textcolor{MyRed}{\hat{\varphi_t}(x)} . 
\end{equation}
It holds that $\textcolor{MyBlue}{\chi}$ and $\textcolor{MyRed}{\hat{\chi}}$ satisfy a pair of \emph{Backward} and \emph{Forward Kolmogorov} equations. In \Cref{app:sec:time-reversal-killed} we show that, if $\hat{\calK}^{0,\star}$ denotes the \emph{dual} of $\hat{\calK}^0$, 
\eqref{eq:system_unrolled} is equivalent to
\begin{equation}
\textstyle
  \label{eq:system_rolled} \partial_t \textcolor{MyBlue}{\chi_t} = - \hat{\calK}^0 \textcolor{MyBlue}{\chi_t} , \qquad \partial_t \textcolor{MyRed}{\hat{\chi}_t} = \hat{\calK}^{0,\star} \textcolor{MyRed}{\hat{\chi}_t} , 
\end{equation}
This means that \eqref{eq:system_unrolled} is in fact a system of
\emph{dual} equations, even though appearing non-symmetric at first sight. This system can therefore be related to existing work such as \citet{liu2022deep}. Finally, \Cref{prop:unbalanced_sb}
reveals that the solution $\Pbb^\SB$ when the reference is a killing process is \textit{also a killing process}. In particular, \eqref{eq:killing_optimal} implies that the
difference between its generator and the one of the reference depends only on
$(\textcolor{MyBlue}{\varphi}, \textcolor{MyBlue}{\Psi})$. Analogously, the time reversal of $\Pbb^\SB$ is
a birth process which, as seen in \eqref{eq:birth_optimal}, only depends on
$(\textcolor{MyRed}{\hat{\varphi}}, \textcolor{MyRed}{\hat{\Psi}})$.

\paragraph{Unbalanced \acl{IPF}.} 
We now describe the unbalanced \ac{IPF} scheme which serves as the foundation for our proposed algorithms. We consider the sequence $(\Pbb^n)_{n \in \nset}$ of measures on $\hat{\rset}^d$, given by \eqref{eq:ipf_iteration}, and recall that $\Pbb^0$ is associated
with a diffusion \emph{with killing}.
For any $n \in \nset$, we have that:
\begin{enumerate*}[label=(\alph*)]
\item $\Pbb^{2n+1}$ is a diffusion \emph{with birth}, given by
  the time-reversal of $\Pbb^{2n}$, initialized at $\mu_1$;
\item $\Pbb^{2n+2}$ is a diffusion \emph{with killing}, given by
  the time-reversal of $\Pbb^{2n+1}$, initialized at $\mu_0$.
\end{enumerate*}
In order to extend \acp{DSB} to \emph{unbalanced} problems, we need to estimate the
parameters appearing in \eqref{eq:time_reversal_kill_informal} and
\eqref{eq:time_reversal_birth_main}. Contrary to \citet{de2021diffusion}, it is not sufficient to estimate the drift of the diffusion, since we also need to update the killing/birth rate. This motivates the following result.

\begin{proposition}
  \label{prop:ipf_form}
  Under mild assumptions, there exist
  $(\textcolor{MyBlue}{\varphi^n}, \textcolor{MyRed}{\hat{\varphi}^n}, \textcolor{MyBlue}{\Psi^n}, \textcolor{MyRed}{\hat{\Psi}^n})_{n \in \nset}$ such that,
  for any $n \in \nset$, $\Pbb^{2n}$ is initialized at $\Pbb^{2n}_0 = \mu_0$ and
  associated with $\hat{\calK}^n$ given for any $t \in \ccint{0,T}$ and
  $x \in \hat{\rset}^d$ by 
  \begin{equation}
    \label{eq:killing_step_n}
\textstyle    \hat{\calK}^n_t(f)(x) = \left[\langle b(x) + \nabla \log \textcolor{MyBlue}{\varphi^n_t(x)}, \nabla f(x) \rangle + \tfrac{1}{2} \Delta f(x) - k(x) (\textcolor{MyBlue}{\Psi^n_t / \varphi^n_t(x)}) (f(x) - f(\infty))\right] \mathbf{1}_{\rset^d}(x).
\end{equation}
Similarly, $\Pbb^{2n+1}$ is initialized at $\Pbb^{2n+1}_T = \mu_1$ and associated
with $\hat{\calB}^{n+1}$ given by
  \begin{align}
    \label{eq:birth_step_n}
    \hat{\calB}^{n+1}_t(f)(x) &= (\langle -b(x) + \nabla \log \textcolor{MyRed}{\hat{\varphi}^{n+1}_{T-t}(x)}, \nabla f(x) \rangle + \tfrac{1}{2}\Delta f(x))  \mathbf{1}_{\rset^d}(x) \\
    & \qquad \textstyle +  \int_{\rset^d} (\textcolor{MyRed}{\hat{\varphi}^{n+1}_{T-t}(\tilde{x})/\hat{\Psi}_{T-t}^{n+1}}) k(\tilde{x})   (f(\tilde{x}) - f(\infty)) \rmd \tilde{x} \mathbf{1}_{\infty}(x) .     
  \end{align}
  We have that for, any $n \in \nset$, $t \in \ccint{0,T}$ and $x \in \rset^d$
  \begin{align}
    \textstyle \log \textcolor{MyBlue}{\varphi^n_t(x)} + \log \textcolor{MyRed}{\hat{\varphi}^{n+1}_t(x)} &= \log p_t^{2n}(x) , &\textcolor{MyBlue}{\Psi^{n}_t}  \textcolor{MyRed}{\hat{\Psi}^{n+1}_t} &=\textstyle 1 - \int_{\rset^d} p_t^{2n}(x) \rmd x , \\
    \textstyle \log \textcolor{MyBlue}{\varphi_t^{n+1}(x)} + \log \textcolor{MyRed}{\hat{\varphi}^{n+1}_t(x)} &= \log p_t^{2n+1}(x) , &\textcolor{MyBlue}{\Psi^{n+1}_t}  \textcolor{MyRed}{\hat{\Psi}^{n+1}_t} &=\textstyle 1 - \int_{\rset^d} p_t^{2n+1}(x) \rmd x,   \label{eq:update}
  \end{align}
  where $p_t^n$ is the density w.r.t. the Lebesgue measure of $\Pbb^n_t$
  restricted to $\rset^d$.
\end{proposition}

Leveraging \Cref{prop:ipf_form} and in particular \eqref{eq:update}, we can
derive an iterative algorithm to compute the path measures $(\Pbb^n)_{n \in \nset}$
by updating $(\textcolor{MyBlue}{\varphi^n}, \textcolor{MyRed}{\hat{\varphi}^n}, \textcolor{MyBlue}{\Psi^n}, \textcolor{MyRed}{\hat{\Psi}^n})_{n \in
  \nset}$. 

\paragraph{Sampling from diffusion with killing or birth.} Finally, whether it be
to approximate the quantity
$(\textcolor{MyBlue}{\varphi^n}, \textcolor{MyRed}{\hat{\varphi}^n}, \textcolor{MyBlue}{\Psi^n}, \textcolor{MyRed}{\hat{\Psi}^n})_{n \in \nset}$ or to sample
from the (approximate) unbalanced Schr\"odinger bridge, we need to discretize birth and death dynamics. We start by discussing how to sample from a
diffusion with killing of the form \eqref{eq:inf_gen_extended_rd_main}. Our
approach is based on the Lie-Trotter-Kato formula \cite[Corollary 6.7,
p.33]{ethier2009markov}. Since $\hat{\calK}$, given by
\eqref{eq:inf_gen_extended_rd_main}, can be decomposed in
$\hat{\calK}_{\mathrm{cont}}$ and $\hat{\calK}_{\mathrm{disc}}$ such that
\begin{equation}
  \hat{\calK}_{\mathrm{cont}}(f)(x) = \langle b(x), \nabla f(x) \rangle + \tfrac{1}{2} \Delta f(x), \qquad  \hat{\calK}_{\mathrm{disc}}(f)(x) = - k(x)(f(x) - f(\infty)),
\end{equation}
we alternate between sampling according to $\hat{\calK}_{\mathrm{cont}}$ and to $\hat{\calK}_{\mathrm{disc}}$, where the latter describes a
\emph{pure} killing process. For small-enough step-sizes, \citet[Corollary 6.7,
p.33]{ethier2009markov} assert that this amounts to sampling according to
$\hat{\calK}$. We discretize the diffusion process using the standard Euler-Maruyama method and use coin flipping for $\hat{\calK}_{\mathrm{disc}}$ (see \Cref{app:sec:unbalanced_ipf} for details).

Sampling diffusion with birth is, however, more problematic. We can still decompose $\hat{\calB}$ from \eqref{eq:hold1} into
$\hat{\calB}_{\mathrm{cont}}$ and $\hat{\calB}_{\mathrm{disc}}$ with
\begin{equation}
  \textstyle \hat{\calB}_{\mathrm{cont}}(f)(x) = \langle b(x), \nabla f(x) \rangle + \tfrac{1}{2} \Delta f(x), \qquad  \hat{\calB}_{\mathrm{disc}}(f)(\infty) = \int_{\rset^d} q(\tilde{x})(f(\tilde{x}) - f(\infty)) \rmd \tilde{x},
\end{equation}
but $\hat{\calB}_{\mathrm{disc}}$ is a \emph{pure} birth process, which is more difficult to sample from.
We do that by considering time intervals of size
$\gamma>0$ and assuming that the birth rate can be written as $q(\tilde{x}) = \ell(\tilde{x}) p(\tilde{x})$, where $\int_{\rset^d} p(\tilde{x}) \rmd \tilde{x} = 1$. We can then sample
$\hat{X} \sim p$ and then let $X = \hat{X}$ with probability $\ell(\hat{X}) \gamma$
and $X=\infty$ otherwise, in which case the birth is rejected.

\looseness -1 When executing our \ac{IPF} scheme, the form of $\hat{\calB}_{\mathrm{disc}}^{n+1}$, given by \eqref{eq:birth_step_n}, is
\begin{align}
  \textstyle \hat{\calB}_{\mathrm{disc}}^{n+1}(f)(t,\infty) &=\textstyle \int_{\rset^d} \textcolor{MyRed}{(\hat{\varphi}_{T-t}^{n+1}(\tilde{x})/\hat{\Psi}_{T-t}^{n+1})} k(\tilde{x})   (f(\tilde{x}) - f(\infty)) \rmd \tilde{x} \\
  &\textstyle= \int_{\rset^d} p_{T-t}^{2n}(\tilde{x}) k(\tilde{x})  /(\textcolor{MyBlue}{\varphi_{T-t}^{n}(x)} \textcolor{MyRed}{\hat{\Psi}_{T-t}^{n+1}}) (f(\tilde{x}) - f(\infty)) \rmd \tilde{x},
\end{align}
\begin{wrapfigure}{r}{0.38\textwidth}
    \centering
    \includegraphics[width=0.36\textwidth]{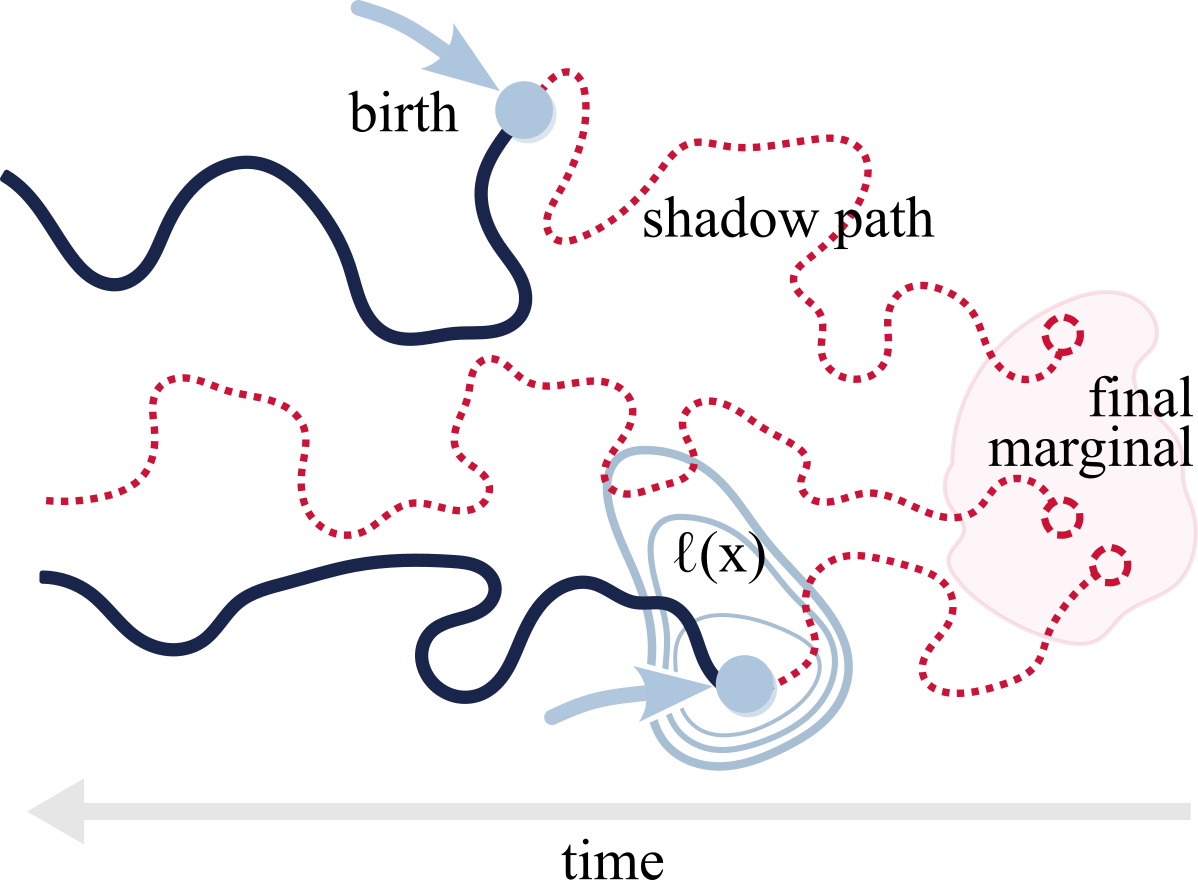}
    \captionof{figure}{Shadow trajectory sampling.}
        \label{fig:shadow_paths}
    \vspace{-20pt}
\end{wrapfigure}
where we obtained the second equality using \eqref{eq:update}. We therefore have that $\ell(\tilde{x}) = k(\tilde{x}) /(\textcolor{MyBlue}{\varphi_{T-t}^{n}}(x) \textcolor{MyRed}{\hat{\Psi}_{T-t}^{n+1}})$
and $p(\tilde{x}) = p_{T-t}^{2n}(\tilde{x})$. To sample backward paths $\Pbb^{2n+1}$, we first sample a forward trajectory $(\bfX_t)_{t \in \ccint{0,T}} \sim \Pbb^{2n}$. If the particle is at $\infty$ at time $t$, then we sample a
Bernoulli variable with parameter
$\gamma k(\bfX_{T-t}) /(\textcolor{MyBlue}{\varphi_{T-t}^{n}}(\bfX_{T-t})
\textcolor{MyRed}{\hat{\Psi}_{T-t}^{n+1}})$. The particle is born if this random variable is equal to one. In that case, we set $\bfY_{t + \gamma} = \bfX_{T-t}$ (and $\bfY_{t + \gamma}=\infty$ otherwise). After birth, backward trajectories evolve according to the continuous diffusion in $\rset^d$ described by $\hat{\calB}_{\mathrm{cont}}$. We call this method \emph{shadow trajectory sampling} and illustrate it in Fig. \ref{fig:shadow_paths} while providing the algorithm in \Cref{app:sec:unbalanced_ipf}.

\section{Unbalanced Diffusion Schr\"odinger Bridge with Temporal Difference Loss}
Equipped with \Cref{prop:ipf_form} and the sampling strategies presented in the previous paragraph, we are now ready to describe our numerical implementation of
\eqref{eq:ipf_iteration}. We parameterize
$\log \textcolor{MyBlue}{\varphi^n_t}, \log \textcolor{MyRed}{\hat{\varphi}^n_t}$ with the two networks $f_{t, \theta_n}$
and $\hat{f}_{t, \hat{\theta}_n}$, respectively. Note that, contrary to the
classical \acl{DSB} setting
\citep{de2021diffusion,chen2021likelihood}, we need to estimate
$\log \textcolor{MyBlue}{\varphi^n_t}, \log \textcolor{MyRed}{\hat{\varphi}^n_t}$ and not only
$\nabla \log \textcolor{MyBlue}{\varphi^n_t}, \nabla \log \textcolor{MyRed}{\hat{\varphi}^n_t}$.

\paragraph{Estimating $\log \varphi^n$ and $\log \hat{\varphi}^n$.}
To learn $\hat{f}_{t,\hat{\theta}^{n+1}}$, we first sample from
$\Pbb^{2n}$ using $f_{t, \theta^n}$ and
$\textcolor{MyBlue}{\Psi_t^n}$, which are assumed to be known. Once this is
done, we compute a loss on $\hat{f}_{t,\hat\theta^{n+1}}$ combining the \emph{mean-matching} (MM) loss
\citep{de2021diffusion} (which is a loss on $\nabla \hat{f}_{t,
  \hat{\theta}}$) and a \emph{temporal difference} (TD) loss
\citep{liu2022deep} (which is a loss on $\hat{f}_{t, \hat{\theta}}$). The MM loss was designed to compute iterates in classical \ac{DSB} while the TD loss was introduced in \citet{liu2022deep} in order to compute
\emph{generalized} Schr\"odinger bridges. The TD loss can be seen as a
\emph{regularizer} to the MM loss. Indeed, at equilibrium, minimizers
$\hat{f}_{t, \hat{\theta}^{\star}}$ of the MM loss satisfy $\hat{f}_{t,
  \hat{\theta}^{\star}} = \log \textcolor{MyRed}{\hat{\varphi}^{\star}_t} +
c_t$, where $c_t$ only depends on the time
$t$.  In \Cref{app:sec:algorithmic_details}, we provide further details on how the MM and TD losses can be adapted to the \emph{unbalanced} setting. So far we
have described how to update $\hat{f}_{t, \theta^{n+1}}$ given $f_{t,
  \theta^n}$ and
$\textcolor{MyBlue}{\Psi^n_t}$. The update of
$\textcolor{MyRed}{\hat{\Psi}^{n+1}_t}$, instead, leverages the closed-form expression
$\textcolor{MyRed}{\hat{\Psi}^{n+1}_t} = (1 - \int_{\rset^d} p_t^{2n}(x) \rmd
x)/\textcolor{MyBlue}{\Psi^{n}_t}$, which follows from \eqref{eq:update}. $\textcolor{MyRed}{\hat{\Psi}^{n+1}_t}$ can, in fact, be computed on the fly, by
approximating $\int_{\rset^d} p_t^{2n}(x) \rmd
x$ with the proportion of \emph{live} particles at time
$t$ in the forward process. Once we have estimated $(\hat{f}_{t,
  \hat{\theta}^{n+1}},
\textcolor{MyRed}{\hat{\Psi}^{n+1}})$ we can estimate $(f_{t, \theta^{n+1}},
\textcolor{MyBlue}{\Psi^{n+1}_t})$ in a similar fashion.

\paragraph{Updating $\Psi^n$.} We know from \eqref{eq:system_unrolled} that at
$\textcolor{MyBlue}{\Psi^n_t}$ does not depend on $t$.\footnote{This is because
  the system \eqref{eq:system_unrolled} is still valid for the \emph{iterates}
  $(\textcolor{MyBlue}{\varphi^n}, \textcolor{MyRed}{\hat{\varphi}^n},
  \textcolor{MyBlue}{\Psi^n}, \textcolor{MyRed}{\hat{\Psi}^n})_{n \in \nset}$.}
Using this observation, we consider a \emph{correction} strategy which projects
$\textcolor{MyBlue}{\Psi_t}$ on constant functions. As a consequence of
\eqref{eq:system_unrolled}, we have for any $n \in \nset$,
\begin{equation}
  \label{eq:global_psi}
  \textstyle \textcolor{MyBlue}\Psi^n = (\mu_0(\rset^d) - \mu_1(\rset^d)) / \int_0^1 \int_{\rset^d} k(\tilde{x}) \textcolor{MyRed}{\hat{\varphi}_t^n(\tilde{x})} \rmd \tilde{x} .
\end{equation}
We propose a numerical approximation of
\eqref{eq:global_psi} in 
\Cref{app:sec:algorithmic_details}.

\paragraph{Algorithm.}
We are now ready to introduce our numerical approximation of Unbalanced IPF, termed \emph{Unbalanced Diffusion
  Schr\"odinger Bridge with Temporal Difference} (UDSB-TD) and described in
\Cref{alg:usb_training}. The procedures \textsc{Update-Psi},
\textsc{Sample-forward} and \textsc{Sample-backward} are given in
\Cref{app:sec:algorithmic_details}. While our algorithm resembles the one of
\citet{liu2022deep}, we highlight some key differences:
\begin{enumerate*}[label=(\alph*)]
\item In \citet{liu2022deep}, the forward and backward processes do not
  incorporate killing and/or birth;
\item as a consequence, it is not possible to update the killing and birth rates
  to match a desired mass constraint. It can, in fact, be shown that the formulation of \citet{liu2022deep} corresponds to the \emph{reweighted} approach in
  \citet{chen2022most}, which is \emph{not} equivalent to the unbalanced SB.
\end{enumerate*}

\begin{algorithm}[h!]
    \caption{\usbalgo{} training}
    \label{alg:usb_training}
    \begin{algorithmic}[1]
        \item[] \textbf{Input:} $\theta, {\hat{\theta}}, \mu_0, \mu_1$
        \item[] \textbf{Output:} $\theta, {\hat{\theta}}$ 
        \For{epoch $n \in \{0, ..., N\}$}
            
            \State $\psi \gets $ \Call{Update-Psi}{$\theta, \hat{\theta}, \psi$} \label{alg:line:psi_update} \Comment{Update death rate}
            \State $(\bfY_t)_{t \in \ccint{0,T}} \gets $ \Call{Sample-Backward}{$\theta, \hat{\theta}, \psi$} \Comment{Sample from birth process}
            \While{reuse paths}
                \State $\mathcal{L}_\text{MM}(\theta) \gets \mathcal{L}_\text{MM}((\bfY_t)_{t \in \ccint{0,T}}; \theta),\; \mathcal{L}_\text{TD}(\theta) \gets \mathcal{L}_\text{TD}((\bfY_t)_{t \in \ccint{0,T}}; \theta)$ \Comment{Compute loss}
    
                \State Update $\theta$ using $\nabla_\theta(\mathcal{L}_\text{IPF} + \mathcal{L}_\text{TD})$ \label{alg:line:grad_update_f} \Comment{Train forward potential}
            \EndWhile
    
            \State $(\bfX_t)_{t \in \ccint{0,T}} \gets $ \Call{Sample-Forward}{$\theta, \hat{\theta}, \psi$} \Comment{Sample from kill process}
    
            \While{reuse paths}
                \State $\mathcal{L}_\text{MM}({\hat{\theta}}) \gets \mathcal{L}_\text{MM}((\bfX_t)_{t \in \ccint{0,T}}; {\hat{\theta}}),\; \mathcal{L}_\text{TD}({\hat{\theta}}) \gets \mathcal{L}_\text{TD}((\bfX_t)_{t \in \ccint{0,T}}; {\hat{\theta}})$
    
                \State Update ${\hat{\theta}}$ using $\nabla_{\hat{\theta}}(\mathcal{L}_\text{IPF} + \mathcal{L}_\text{TD})$ \label{alg:line:grad_update_b}  \Comment{Train backward potential}
            \EndWhile
        \EndFor
    \end{algorithmic}
  \end{algorithm}

\paragraph{Limitations of UDSB-TD.} Although theoretically grounded, \usbalgo{} has some limitations. Like in \citet{liu2022deep}, it needs estimates of 
$\log \textcolor{MyBlue}{\varphi^n_t}, \log
\textcolor{MyRed}{\hat{\varphi}^n_t}$, and not only of
$\nabla \log \textcolor{MyBlue}{\varphi^n_t}, \nabla \log
\textcolor{MyRed}{\hat{\varphi}^n_t}$, in order to update
$\textcolor{MyRed}{\hat{\Psi}}$ and $\textcolor{MyBlue}{\Psi}$. While the TD loss allows for that, it is less stable than the MM loss and, more importantly, requires estimates of $\rho_0, \rho_1$ which might not be available in practice.
Furthermore, the formula \eqref{eq:global_psi} used to update $\Psi$ might be numerically unstable, 
especially in high dimensions.

\paragraph{Heuristic estimation of $\Psi$.} To circumvent these limitations, we
introduce another algorithm (\usbferryman{}) that exhibits a better behavior w.r.t. the
dimension of the problem and does not require the estimation of
$\log \textcolor{MyBlue}{\varphi^n_t}, \log
\textcolor{MyRed}{\hat{\varphi}^n_t}$. While we do not prove the theoretical
validity of this new procedure, we verify empirically in
\Cref{app:sec:experiment_details} that its results are consistent with UDSB-TD in small
dimensions. Following \eqref{eq:killing_optimal}, we know that, at equilibrium,
the update on the killing rate is given by
$x \mapsto \textcolor{MyBlue}{\Psi_t/\varphi_t(x)}$. Therefore, we propose to
approximate this ratio by
$x \mapsto g_{\zeta, t}(x)$, where
$g_\zeta$ is learnable. Rewriting \eqref{eq:global_psi} at equilibrium, we have 
\begin{equation}
  \label{eq:global_psi_update}
  \textstyle \int_0^1 \int_{\rset^d} k(x) \textcolor{MyBlue}{(\Psi_t/\varphi_t(x))} p_t(x)  \rmd x  = \mu_0(\rset^d) - \mu_1(\rset^d) .
\end{equation}
This suggests considering the loss
$\mathcal{L}(\zeta) = \int_0^1\expeLigne{k(\bfX_t)g_{\zeta,t}(\bfX_t)} \rmd t  -
  \mu_0(\rset^d) + \mu_1(\rset^d)$. We again highlight that minimizing this
loss does not ensure that $g_{\zeta,t}$ is the optimal update in
\Cref{prop:unbalanced_sb}. However, using this loss we remark that we no longer
need to estimate
$\log \textcolor{MyBlue}{\varphi_t^n}, \log
\textcolor{MyRed}{\hat{\varphi}_t^n}$. Therefore, we can drop the TD loss and make the algorithm more scalable by not requiring estimates of
$\rho_0, \rho_1$. The algorithm performing the revised update of the killing
rate is presented fully in \Cref{app:sec:algorithmic_details}.

\section{Experiments}
\label{sec:experiments}
In this section, we assess the performance of our UDSB solver in two tasks: the reconstruction of simple dynamics in the plane and the modeling of cellular responses to a cancer drug. For the treatment of COVID variants spread and additional experiments comparing the two schemes discussed in \cref{sec:algo} (\usbalgo{} and \usbferryman{}), we direct the reader to \cref{app:sec:experiment_details}. Henceforth, we utilize the \usbferryman{} algorithm due to its enhanced stability and wider applicability.

\begin{figure*}[t]
    \centering
    \includegraphics[width=\textwidth]{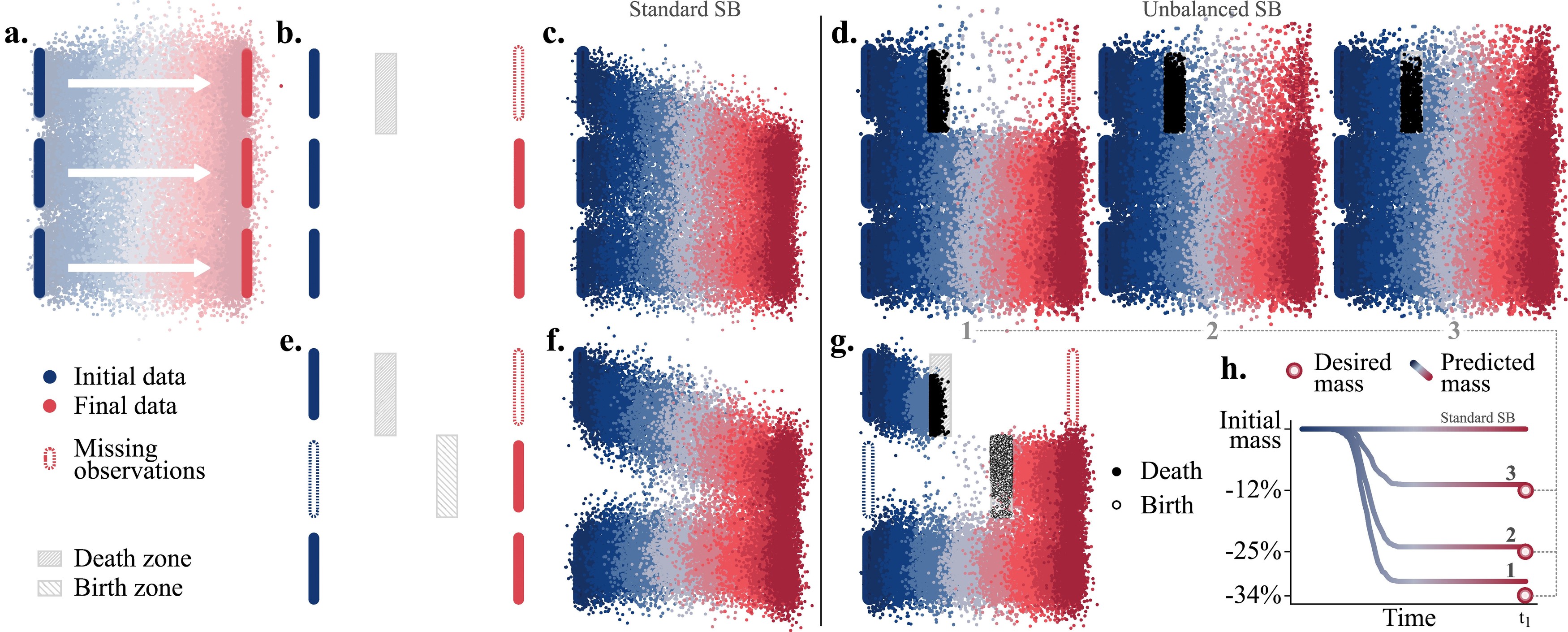}
    \caption{Synthetic datasets. (\textbf{a}) Initial (\textit{blue}) and final (\textit{red}) empirical distributions, consisting of 3 groups of points moving horizontally. When some groups are missing in observations (\textbf{b}, \textbf{e}, dashed), standard SBs fail to learn the correct dynamics (\textbf{c}, \textbf{f}). By introducing death and birth priors (\textit{gray} rectangles), unbalanced SBs recover, instead, the true evolution law (\textbf{d}, \textbf{g}). They learn valid interpolations (\textbf{d1-3}) between incomplete marginals which also reproduce the (arbitrary) mass loss observed at the end (\textbf{h}, circles).}
    \label{fig:results_toy}
\end{figure*}

\paragraph{Synthetic dynamics.}
We first consider the toy dataset displayed in Fig.~\ref{fig:results_toy}a, consisting of 3 groups of points that are known to move along the horizontal axis. Standard SBs easily reconstruct the dynamics whenever representatives of all groups appear in observed marginals. However, some clusters may not appear in the empirical distribution. Under-representation could be, for instance, attributable to the exceedingly small number of samples considered or be related to other experimental factors. We represent two of these cases: one group does not appear in the end marginal in Fig.~\ref{fig:results_toy}b, while two are missing in those of Fig.~\ref{fig:results_toy}e (top-right and middle-left).
In both cases, standard SBs fail to capture the correct dynamics, since they generate diagonal trajectories (Fig.~\ref{fig:results_toy}c, \ref{fig:results_toy}f). Orphan particles, which start at the top-left, are in fact forced to reach their closest points in the final marginal, i.e., the middle cluster on the right. This is the product of the assumption placed by SBs that all particles move continuously in the state space. It is not possible to encode the knowledge that a (known) fraction of particles should leave (or enter) the system and that some particles may be extraneous.

\looseness -1 Unbalanced SBs provide instead a natural way of identifying regions containing particles that should likely not be considered when learning the diffusion, or ones in which new particles should appear. They consist of death (and birth) priors, which we draw here as \textit{gray} rectangles. A death zone is, for instance, placed close to the orphan marginal in Fig.~\ref{fig:results_toy}b, while a birth zone stands beside the central cluster on the right of Fig.~\ref{fig:results_toy}e, since that group of particles does not have a counterpart in the initial marginal. UDSBs then learn trajectories between incomplete marginals, by only using live particles. The statuses of particles are determined by incrementally adjusting the death (and birth) priors to match a predefined amount of mass loss. For example, we consider three different amounts of mass remaining at time $t_1$ (Fig.~\ref{fig:results_toy}h) and compute UDSB trajectories for each of them (Fig.~\ref{fig:results_toy}d).
In all cases, our algorithm computes correct paths for live particles while ensuring that the number of deaths matches the mass constraint (Fig.~\ref{fig:results_toy}h).

\paragraph{Cellular dynamics.}
\begin{wraptable}{r}{.6\textwidth}
    \centering
    \vspace{-10pt}
        \begin{tabular}{lcc}
        \toprule
         & \multicolumn{2}{c}{\textbf{Cell Differentiation}} \\
        \cmidrule(lr){2-3}
        \textbf{Methods} & MMD $\downarrow$ & $\text{W}_\varepsilon \downarrow$ \\
        \midrule
         \citet{chen2021likelihood} & 1.86e-2 (0.04e-2) & 6.23 (0.02)  \\
         Ours, no deaths/births & 1.86e-2 (0.09e-2) & 6.27 (0.11) \\
         Ours & \textbf{1.75e-2} (0.11e-2) & \textbf{6.11} (0.11)  \\
         \bottomrule
        \end{tabular}
        \caption{\looseness -1\textbf{Cell evolution prediction results.} We compare our method against a baseline, a classical SB solver, by measuring the quality of the predicted end statuses of cells. We use two distributional metrics (MMD, $\text{W}_{\varepsilon}$) and report averages and standard deviations (in parentheses) over 10 runs.}
        \label{tab:results_cells}
\end{wraptable}
Unbalanced SBs have a natural application in the field of cellular biology, where the appearance or disappearance of mass has the physical meaning of cell birth or death. We examine the dataset collected by \cite{bunne2021learning}, which tracks the time evolution of single-cell markers of melanoma cells undergoing treatment with a cancer drug. 
Cell statuses are recorded at 3 different times and their evolution can be approximated by standard SBs as a continuous diffusion in their (50-dimensional) feature space. However, this model ignores that consecutive measurements capture different cell populations, owing to the death of some cells --caused either by natural or drug-induced reasons-- and to the birth of new ones.
Trajectories of dying cells are therefore artificially steered to match the final marginal while the presence of newborn cells cannot be properly taken into account.
By allowing for jumps in the trajectories of cell statuses, unbalanced SBs overcome, instead, both limitations and more accurately reproduce the measurements. They can, in fact, kill cells with statuses that are dissimilar to the ones found in subsequent measurements while also generating new cells close to observed ones.

\begin{figure*}[t]
    \centering
    \includegraphics[width=\textwidth]{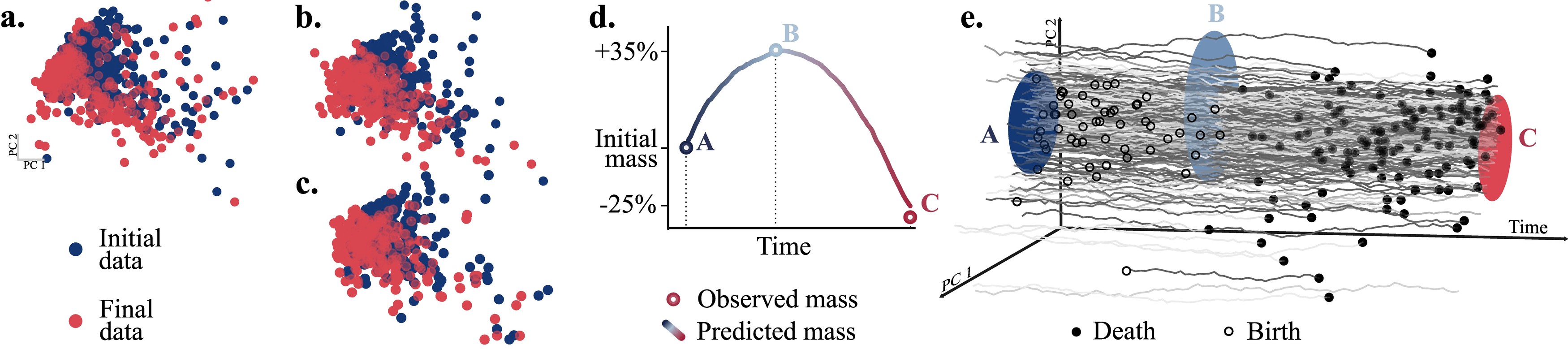}
    \caption{Single-cell response to a cancer drug. (\textbf{a}) Cell states (projected along their first two principal components) are measured at the beginning (\textit{blue}) and at the end (\textit{red}). The predictions computed by \usbferryman{} (\textbf{c}) closely matches the one of a standard SB solver (\textbf{b}) but also accounts for changes in cell population observed at intermediate times (\textbf{d}). The dynamics learned by our unbalanced SB solver (\textbf{e}) define the evolution of cell states but also describe death and birth events (\textit{black} and \textit{white} dots). Ellipses identify the $3\sigma$ region around the mean of the observed population at each timepoint.}
    \label{fig:results_cell}

\end{figure*}

In Fig. \ref{fig:results_cell}e, we plot the temporal evolution of cell trajectories computed by \usbferryman{}. Colored ellipses represent statuses less than 3 standard deviations away from empirical means. Deaths, represented by \textit{black} dots, allow removing some of the cells that are far away from observed statuses, while birth (\textit{white} dots) help introduce new cells located in the densest parts of the state-space. \usbferryman{} ensures that the number and temporal distribution of deaths/births match observations at both timepoints (Fig. \ref{fig:results_cell}d). At the same time, it learns a standard SB on live particles, which produces better quality predictions of the end marginal (\cref{tab:results_cells}) compared to the (balanced) SB solver by \cite{chen2021likelihood}. Interestingly, if we sample trajectories from \usbferryman{} but disregard deaths and births, the quality of the predictions deteriorates below the baseline, pointing to the role of the coffin state in improving the modeling.

\section{Conclusion}

In conclusion, this work addresses the important task of transporting data in the \emph{unbalanced} setting, where mass can appear or disappear, which is particularly relevant for accurately tracking population changes in various biology applications. Our key contribution is an extension of the \ac{IPF} procedure for solving diffusion models with death and birth mechanisms, enabling comprehensive modeling of dynamics with time-varying mass information. The derivation of time-reversal formulas for diffusions with killing and birth terms plays a crucial role in this extension, providing the foundation for the proposed algorithms. The resulting efficient and scalable algorithms are evaluated on various tasks, including the modeling of cellular responses to cancer drugs as well as simulating the emergence and spread of viruses. 

Furthermore, our work opens up exciting research avenues via the established connections between the versatile \ac{SB} formulation and other fields. In terms of modeling, one promising direction is the application of our approach to solving equilibria in \emph{mean-field stochastic games} where agents may enter or exit the game, as observed in traffic transportation data or epidemiology. Additionally, the exploration of the links between unbalanced SB and fields such as stochastic optimal control and Wasserstein geometry holds the promise of deepening the theoretical understanding of \ac{SB} problems on the extended, non-Euclidean spaces.

\begin{ack}
The dataset used in this paper was generated by Lucas Pelkmans, Gabriele Gut, and Jacobo Sarabia Castillo and was first published by \citet{bunne2021learning} and \citet{lubeck2022neural}.
This publication was supported by the NCCR Catalysis (grant number 180544), a National Centre of Competence in Research funded by the Swiss National Science Foundation as well as the European Union’s Horizon 2020 research and innovation programme 826121. 
\end{ack}

\bibliography{main}

\begin{thebibliography}{71}
\providecommand{\natexlab}[1]{#1}
\providecommand{\url}[1]{\texttt{#1}}
\expandafter\ifx\csname urlstyle\endcsname\relax
  \providecommand{\doi}[1]{doi: #1}\else
  \providecommand{\doi}{doi: \begingroup \urlstyle{rm}\Url}\fi

\bibitem[Anderson(1982)]{anderson1982reverse}
Brian~DO Anderson.
\newblock Reverse-time diffusion equation models.
\newblock \emph{Stochastic Processes and their Applications}, 12\penalty0
  (3):\penalty0 313--326, 1982.

\bibitem[Benton et~al.(2022)Benton, Shi, De~Bortoli, Deligiannidis, and
  Doucet]{benton2022denoising}
Joe Benton, Yuyang Shi, Valentin De~Bortoli, George Deligiannidis, and Arnaud
  Doucet.
\newblock From denoising diffusions to denoising markov models.
\newblock \emph{arXiv preprint arXiv:2211.03595}, 2022.

\bibitem[Blumenthal and Getoor(2007)]{blumenthal2007markov}
Robert~McCallum Blumenthal and Ronald~Kay Getoor.
\newblock \emph{Markov processes and potential theory}.
\newblock Courier Corporation, 2007.

\bibitem[Bradbury et~al.(2018)Bradbury, Frostig, Hawkins, Johnson, Leary,
  Maclaurin, Necula, Paszke, Vander{P}las, Wanderman-{M}ilne, and
  Zhang]{jax2018github}
James Bradbury, Roy Frostig, Peter Hawkins, Matthew~James Johnson, Chris Leary,
  Dougal Maclaurin, George Necula, Adam Paszke, Jake Vander{P}las, Skye
  Wanderman-{M}ilne, and Qiao Zhang.
\newblock {JAX}: composable transformations of {P}ython+{N}um{P}y programs,
  2018.
\newblock URL \url{http://github.com/google/jax}.

\bibitem[Brockmann and Helbing(2013)]{brockmann2013hidden}
Dirk Brockmann and Dirk Helbing.
\newblock The hidden geometry of complex, network-driven contagion phenomena.
\newblock \emph{Science}, 342\penalty0 (6164):\penalty0 1337--1342, 2013.
\newblock \doi{10.1126/science.1245200}.

\bibitem[Bunne et~al.(2021)Bunne, Stark, Gut, del Castillo, Lehmann, Pelkmans,
  Krause, and Ratsch]{bunne2021learning}
Charlotte Bunne, Stefan~G Stark, Gabriele Gut, Jacobo~Sarabia del Castillo,
  Kjong-Van Lehmann, Lucas Pelkmans, Andreas Krause, and Gunnar Ratsch.
\newblock {Learning Single-Cell Perturbation Responses using Neural Optimal
  Transport}.
\newblock \emph{bioRxiv}, 2021.

\bibitem[Bunne et~al.(2022)Bunne, Meng-Papaxanthos, Krause, and
  Cuturi]{bunne2022proximal}
Charlotte Bunne, Laetitia Meng-Papaxanthos, Andreas Krause, and Marco Cuturi.
\newblock {Proximal Optimal Transport Modeling of Population Dynamics}.
\newblock In \emph{International Conference on Artificial Intelligence and
  Statistics (AISTATS)}, 2022.

\bibitem[Bunne et~al.(2023)Bunne, Hsieh, Cuturi, and
  Krause]{bunne2022recovering}
Charlotte Bunne, Ya-Ping Hsieh, Marco Cuturi, and Andreas Krause.
\newblock {The Schr\"odinger Bridge between Gaussian Measures has a Closed
  Form}.
\newblock In \emph{International Conference on Artificial Intelligence and
  Statistics (AISTATS)}, 2023.

\bibitem[Caffarelli and McCann(2010)]{caffarelli2010free}
Luis~A Caffarelli and Robert~J McCann.
\newblock Free boundaries in optimal transport and monge-ampere obstacle
  problems.
\newblock \emph{Annals of mathematics}, pages 673--730, 2010.

\bibitem[Campbell et~al.(2022)Campbell, Benton, De~Bortoli, Rainforth,
  Deligiannidis, and Doucet]{campbell2022continuous}
Andrew Campbell, Joe Benton, Valentin De~Bortoli, Thomas Rainforth, George
  Deligiannidis, and Arnaud Doucet.
\newblock A continuous time framework for discrete denoising models.
\newblock \emph{Advances in Neural Information Processing Systems},
  35:\penalty0 28266--28279, 2022.

\bibitem[Cao and Liu(2021)]{caoCOVID19ModelingReview2021}
Longbing Cao and Qing Liu.
\newblock {{COVID-19 Modeling}}: {{A Review}}, August 2021.

\bibitem[Cattiaux et~al.(2021)Cattiaux, Conforti, Gentil, and
  L{\'e}onard]{cattiaux2021time}
Patrick Cattiaux, Giovanni Conforti, Ivan Gentil, and Christian L{\'e}onard.
\newblock Time reversal of diffusion processes under a finite entropy
  condition.
\newblock \emph{arXiv preprint arXiv:2104.07708}, 2021.

\bibitem[Chen et~al.(2021)Chen, Liu, and Theodorou]{chen2021likelihood}
Tianrong Chen, Guan-Horng Liu, and Evangelos~A Theodorou.
\newblock {Likelihood Training of {S}chr\"{o}dinger Bridge using
  Forward-Backward {SDE}s Theory}.
\newblock In \emph{arXiv Preprint}, 2021.

\bibitem[Chen et~al.(2022)Chen, Georgiou, and Pavon]{chen2022most}
Yongxin Chen, Tryphon~T Georgiou, and Michele Pavon.
\newblock The most likely evolution of diffusing and vanishing particles:
  Schrodinger bridges with unbalanced marginals.
\newblock \emph{SIAM Journal on Control and Optimization}, 60\penalty0
  (4):\penalty0 2016--2039, 2022.

\bibitem[Chizat et~al.(2018)Chizat, Peyr{\'e}, Schmitzer, and
  Vialard]{chizat2018interpolating}
Lenaic Chizat, Gabriel Peyr{\'e}, Bernhard Schmitzer, and Fran{\c{c}}ois-Xavier
  Vialard.
\newblock An interpolating distance between optimal transport and fisher--rao
  metrics.
\newblock \emph{Foundations of Computational Mathematics}, 18:\penalty0 1--44,
  2018.

\bibitem[Conforti and L{\'e}onard(2022)]{conforti2022time}
Giovanni Conforti and Christian L{\'e}onard.
\newblock Time reversal of markov processes with jumps under a finite entropy
  condition.
\newblock \emph{Stochastic Processes and their Applications}, 144:\penalty0
  85--124, 2022.

\bibitem[Cuturi(2013)]{cuturi2013sinkhorn}
Marco Cuturi.
\newblock {Sinkhorn Distances: Lightspeed Computation of Optimal Transport}.
\newblock In \emph{Advances in Neural Information Processing Systems
  (NeurIPS)}, volume~26, 2013.

\bibitem[Cuturi and Doucet(2014)]{cuturi2014fast}
Marco Cuturi and Arnaud Doucet.
\newblock Fast computation of wasserstein barycenters.
\newblock In \emph{International conference on machine learning}, pages
  685--693. PMLR, 2014.

\bibitem[Cuturi et~al.(2022)Cuturi, Meng-Papaxanthos, Tian, Bunne, Davis, and
  Teboul]{cuturi2022optimal}
Marco Cuturi, Laetitia Meng-Papaxanthos, Yingtao Tian, Charlotte Bunne, Geoff
  Davis, and Olivier Teboul.
\newblock {Optimal Transport Tools (OTT): A JAX Toolbox for all things
  Wasserstein}.
\newblock \emph{arXiv Preprint arXiv:2201.12324}, 2022.

\bibitem[De~Bortoli et~al.(2021)De~Bortoli, Thornton, Heng, and
  Doucet]{de2021diffusion}
Valentin De~Bortoli, James Thornton, Jeremy Heng, and Arnaud Doucet.
\newblock {Diffusion {S}chr\"odinger Bridge with Applications to Score-Based
  Generative Modeling}.
\newblock In \emph{Advances in Neural Information Processing Systems
  (NeurIPS)}, volume~35, 2021.

\bibitem[Dynkin and Dynkin(1965)]{dynkin1965markov}
Borisovich Dynkin and Evgenij Dynkin.
\newblock \emph{Markov processes}.
\newblock Springer, 1965.

\bibitem[Ekeland(2010)]{ekeland2010existence}
Ivar Ekeland.
\newblock Existence, uniqueness and efficiency of equilibrium in hedonic
  markets with multidimensional types.
\newblock \emph{Economic Theory}, pages 275--315, 2010.

\bibitem[Ethier and Kurtz(2009)]{ethier2009markov}
Stewart~N Ethier and Thomas~G Kurtz.
\newblock \emph{Markov processes: characterization and convergence}.
\newblock John Wiley \& Sons, 2009.

\bibitem[Fishman et~al.(2023)Fishman, Klarner, De~Bortoli, Mathieu, and
  Hutchinson]{fishman2023diffusion}
Nic Fishman, Leo Klarner, Valentin De~Bortoli, Emile Mathieu, and Michael
  Hutchinson.
\newblock Diffusion models for constrained domains.
\newblock \emph{arXiv preprint arXiv:2304.05364}, 2023.

\bibitem[Friedman(2012)]{friedman2012stochastic}
Avner Friedman.
\newblock \emph{Stochastic differential equations and applications}.
\newblock Courier Corporation, 2012.

\bibitem[Galichon(2018)]{galichon2018optimal}
Alfred Galichon.
\newblock \emph{Optimal transport methods in economics}.
\newblock Princeton University Press, 2018.

\bibitem[Gramfort et~al.(2015)Gramfort, Peyr{\'e}, and
  Cuturi]{gramfort2015fast}
Alexandre Gramfort, Gabriel Peyr{\'e}, and Marco Cuturi.
\newblock Fast optimal transport averaging of neuroimaging data.
\newblock In \emph{Information Processing in Medical Imaging: 24th
  International Conference, IPMI 2015, Sabhal Mor Ostaig, Isle of Skye, UK,
  June 28-July 3, 2015, Proceedings 24}, pages 261--272. Springer, 2015.

\bibitem[Gretton et~al.(2012)Gretton, Borgwardt, Rasch, Sch{\"o}lkopf, and
  Smola]{gretton2012kernel}
Arthur Gretton, Karsten~M Borgwardt, Malte~J Rasch, Bernhard Sch{\"o}lkopf, and
  Alexander Smola.
\newblock {A Kernel Two-Sample Test}.
\newblock \emph{Journal of Machine Learning Research}, 13, 2012.

\bibitem[Haussmann and Pardoux(1986)]{haussmann1986time}
Ulrich~G Haussmann and Etienne Pardoux.
\newblock Time reversal of diffusions.
\newblock \emph{The Annals of Probability}, pages 1188--1205, 1986.

\bibitem[Ho et~al.(2020)Ho, Jain, and Abbeel]{ho2020denoising}
Jonathan Ho, Ajay Jain, and Pieter Abbeel.
\newblock {Denoising Diffusion Probabilistic Models}.
\newblock In \emph{Advances in Neural Information Processing Systems
  (NeurIPS)}, 2020.

\bibitem[Hyv{\"a}rinen and Dayan(2005)]{hyvarinen2005estimation}
Aapo Hyv{\"a}rinen and Peter Dayan.
\newblock Estimation of non-normalized statistical models by score matching.
\newblock \emph{Journal of Machine Learning Research}, 6\penalty0 (4), 2005.

\bibitem[Karatzas et~al.(1991)Karatzas, Karatzas, Shreve, and
  Shreve]{karatzas1991brownian}
Ioannis Karatzas, Ioannis Karatzas, Steven Shreve, and Steven~E Shreve.
\newblock \emph{Brownian motion and stochastic calculus}, volume 113.
\newblock Springer Science \& Business Media, 1991.

\bibitem[Karlin and Taylor(1981)]{karlin1981second}
Samuel Karlin and Howard~E Taylor.
\newblock \emph{A second course in stochastic processes}.
\newblock Elsevier, 1981.

\bibitem[Kelley(2017)]{kelley2017general}
John~L Kelley.
\newblock \emph{General topology}.
\newblock Courier Dover Publications, 2017.

\bibitem[Kermack and McKendrick(1927)]{kermack1927contribution}
W.~O. Kermack and A.~G. McKendrick.
\newblock A contribution to the mathematical theory of epidemics.
\newblock \emph{Proceedings of the Royal Society of London. Series A,
  Containing Papers of a Mathematical and Physical Character}, 115\penalty0
  (772):\penalty0 700--721, 1927.
\newblock ISSN 0950-1207, 2053-9150.
\newblock \doi{10.1098/rspa.1927.0118}.

\bibitem[Knight(2008)]{knight2008sinkhorn}
Philip~A Knight.
\newblock The sinkhorn--knopp algorithm: convergence and applications.
\newblock \emph{SIAM Journal on Matrix Analysis and Applications}, 30\penalty0
  (1):\penalty0 261--275, 2008.

\bibitem[Kolouri et~al.(2017)Kolouri, Park, Thorpe, Slepcev, and
  Rohde]{kolouri2017optimal}
Soheil Kolouri, Se~Rim Park, Matthew Thorpe, Dejan Slepcev, and Gustavo~K
  Rohde.
\newblock Optimal mass transport: Signal processing and machine-learning
  applications.
\newblock \emph{IEEE signal processing magazine}, 34\penalty0 (4):\penalty0
  43--59, 2017.

\bibitem[Kondratyev et~al.(2016)Kondratyev, Monsaingeon, and
  Vorotnikov]{kondratyev2016new}
Stanislav Kondratyev, L{\'e}onard Monsaingeon, and Dmitry Vorotnikov.
\newblock {A new optimal transport distance on the space of finite Radon
  measures}.
\newblock \emph{Adv. Differential Equations}, 21, 2016.

\bibitem[Liero et~al.(2018)Liero, Mielke, and Savar{\'e}]{liero2018optimal}
Matthias Liero, Alexander Mielke, and Giuseppe Savar{\'e}.
\newblock {Optimal entropy-transport problems and a new Hellinger--Kantorovich
  distance between positive measures}.
\newblock \emph{Inventiones mathematicae}, 211\penalty0 (3), 2018.

\bibitem[Lin et~al.(2023)Lin, Akin, Rao, Hie, Zhu, Lu, Smetanin, Verkuil,
  Kabeli, Shmueli, dos Santos~Costa, Fazel-Zarandi, Sercu, Candido, and
  Rives]{lin2023evolutionary}
Zeming Lin, Halil Akin, Roshan Rao, Brian Hie, Zhongkai Zhu, Wenting Lu, Nikita
  Smetanin, Robert Verkuil, Ori Kabeli, Yaniv Shmueli, Allan dos Santos~Costa,
  Maryam Fazel-Zarandi, Tom Sercu, Salvatore Candido, and Alexander Rives.
\newblock Evolutionary-scale prediction of atomic-level protein structure with
  a language model.
\newblock \emph{Science}, 379\penalty0 (6637):\penalty0 1123--1130, 2023.
\newblock \doi{10.1126/science.ade2574}.

\bibitem[Lipman et~al.(2022)Lipman, Chen, Ben-Hamu, Nickel, and
  Le]{lipman2022flow}
Yaron Lipman, Ricky~TQ Chen, Heli Ben-Hamu, Maximilian Nickel, and Matt Le.
\newblock Flow matching for generative modeling.
\newblock \emph{arXiv preprint arXiv:2210.02747}, 2022.

\bibitem[Liu et~al.(2022)Liu, Chen, So, and Theodorou]{liu2022deep}
Guan-Horng Liu, Tianrong Chen, Oswin So, and Evangelos~A Theodorou.
\newblock Deep generalized schr$\backslash$" odinger bridge.
\newblock \emph{arXiv preprint arXiv:2209.09893}, 2022.

\bibitem[Liu et~al.(2023)Liu, Vahdat, Huang, Theodorou, Nie, and
  Anandkumar]{liu20232}
Guan-Horng Liu, Arash Vahdat, De-An Huang, Evangelos~A Theodorou, Weili Nie,
  and Anima Anandkumar.
\newblock I2sb: Image-to-image schrodinger bridge.
\newblock \emph{arXiv preprint arXiv:2302.05872}, 2023.

\bibitem[Lou and Ermon(2023)]{lou2023reflected}
Aaron Lou and Stefano Ermon.
\newblock Reflected diffusion models.
\newblock \emph{arXiv preprint arXiv:2304.04740}, 2023.

\bibitem[L{\"u}beck et~al.(2022)L{\"u}beck, Bunne, Gut, del Castillo, Pelkmans,
  and Alvarez-Melis]{lubeck2022neural}
Frederike L{\"u}beck, Charlotte Bunne, Gabriele Gut, Jacobo~Sarabia del
  Castillo, Lucas Pelkmans, and David Alvarez-Melis.
\newblock {Neural Unbalanced Optimal Transport via Cycle-Consistent
  Semi-Couplings}.
\newblock \emph{arXiv preprint arXiv:2209.15621}, 2022.

\bibitem[McInnes et~al.(2020)McInnes, Healy, and Melville]{mcinnes2020umap}
Leland McInnes, John Healy, and James Melville.
\newblock Umap: Uniform manifold approximation and projection for dimension
  reduction, 2020.

\bibitem[Mikami(2002)]{mikami2002optimal}
Toshio Mikami.
\newblock Optimal control for absolutely continuous stochastic processes and
  the mass transportation problem.
\newblock \emph{Electron. Commun. Probab.}, 7, 2002.

\bibitem[Nirenberg(1953)]{nirenberg1953strong}
Louis Nirenberg.
\newblock A strong maximum principle for parabolic equations.
\newblock \emph{Communications on Pure and Applied Mathematics}, 6\penalty0
  (2):\penalty0 167--177, 1953.

\bibitem[Nixon et~al.(2022)Nixon, Jindal, Parker, Reich, Ghobadi, Lee,
  Truelove, and Gardner]{nixonEvaluationProspectiveCOVID192022}
Kristen Nixon, Sonia Jindal, Felix Parker, Nicholas~G Reich, Kimia Ghobadi,
  Elizabeth~C Lee, Shaun Truelove, and Lauren Gardner.
\newblock An evaluation of prospective {{COVID-19}} modelling studies in the
  {{USA}}: From data to science translation.
\newblock \emph{The Lancet Digital Health}, 4\penalty0 (10):\penalty0
  e738--e747, October 2022.
\newblock ISSN 25897500.
\newblock \doi{10.1016/S2589-7500(22)00148-0}.

\bibitem[Oksendal(2013)]{oksendal2013stochastic}
Bernt Oksendal.
\newblock \emph{Stochastic differential equations: an introduction with
  applications}.
\newblock Springer Science \& Business Media, 2013.

\bibitem[Pedregosa et~al.(2011)Pedregosa, Varoquaux, Gramfort, Michel, Thirion,
  Grisel, Blondel, Prettenhofer, Weiss, Dubourg, Vanderplas, Passos,
  Cournapeau, Brucher, Perrot, and Duchesnay]{scikit-learn}
F.~Pedregosa, G.~Varoquaux, A.~Gramfort, V.~Michel, B.~Thirion, O.~Grisel,
  M.~Blondel, P.~Prettenhofer, R.~Weiss, V.~Dubourg, J.~Vanderplas, A.~Passos,
  D.~Cournapeau, M.~Brucher, M.~Perrot, and E.~Duchesnay.
\newblock Scikit-learn: Machine learning in {P}ython.
\newblock \emph{Journal of Machine Learning Research}, 12:\penalty0 2825--2830,
  2011.

\bibitem[Pele and Werman(2009)]{pele2009fast}
Ofir Pele and Michael Werman.
\newblock Fast and robust earth mover's distances.
\newblock In \emph{2009 IEEE 12th international conference on computer vision},
  pages 460--467. IEEE, 2009.

\bibitem[Peyré and Cuturi(2019)]{peyre2019computational}
Gabriel Peyré and Marco Cuturi.
\newblock {Computational Optimal Transport}.
\newblock \emph{Foundations and Trends in Machine Learning}, 11\penalty0 (5-6),
  2019.
\newblock ISSN 1935-8245.

\bibitem[Protter and Weinberger(2012)]{protter2012maximum}
Murray~H Protter and Hans~F Weinberger.
\newblock \emph{Maximum principles in differential equations}.
\newblock Springer Science \& Business Media, 2012.

\bibitem[Saharia et~al.(2022)Saharia, Chan, Saxena, Li, Whang, Denton,
  Ghasemipour, Gontijo~Lopes, Karagol~Ayan, Salimans,
  et~al.]{saharia2022photorealistic}
Chitwan Saharia, William Chan, Saurabh Saxena, Lala Li, Jay Whang, Emily~L
  Denton, Kamyar Ghasemipour, Raphael Gontijo~Lopes, Burcu Karagol~Ayan, Tim
  Salimans, et~al.
\newblock Photorealistic text-to-image diffusion models with deep language
  understanding.
\newblock \emph{Advances in Neural Information Processing Systems},
  35:\penalty0 36479--36494, 2022.

\bibitem[Santambrogio(2015)]{santambrogio2015optimal}
Filippo Santambrogio.
\newblock {Optimal Transport for Applied Mathematicians}.
\newblock \emph{Birk{\"a}user, NY}, 55\penalty0 (58-63):\penalty0 94, 2015.

\bibitem[Schiebinger et~al.(2019)Schiebinger, Shu, Tabaka, Cleary, Subramanian,
  Solomon, Gould, Liu, Lin, Berube, et~al.]{schiebinger2019optimal}
Geoffrey Schiebinger, Jian Shu, Marcin Tabaka, Brian Cleary, Vidya Subramanian,
  Aryeh Solomon, Joshua Gould, Siyan Liu, Stacie Lin, Peter Berube, et~al.
\newblock {Optimal-Transport Analysis of Single-Cell Gene Expression Identifies
  Developmental Trajectories in Reprogramming}.
\newblock \emph{Cell}, 176\penalty0 (4), 2019.

\bibitem[Schr{\"o}dinger(1932)]{schrodinger1932theorie}
Erwin Schr{\"o}dinger.
\newblock Sur la th{\'e}orie relativiste de l'{\'e}lectron et
  l'interpr{\'e}tation de la m{\'e}canique quantique.
\newblock In \emph{Annales de l'institut Henri Poincar{\'e}}, volume~2, 1932.

\bibitem[Shi et~al.(2023)Shi, De~Bortoli, Campbell, and
  Doucet]{shi2023diffusion}
Yuyang Shi, Valentin De~Bortoli, Andrew Campbell, and Arnaud Doucet.
\newblock Diffusion schrodinger bridge matching.
\newblock \emph{arXiv preprint arXiv:2303.16852}, 2023.

\bibitem[Sinkhorn and Knopp(1967)]{sinkhorn1967concerning}
Richard Sinkhorn and Paul Knopp.
\newblock Concerning nonnegative matrices and doubly stochastic matrices.
\newblock \emph{Pacific Journal of Mathematics}, 21\penalty0 (2):\penalty0
  343--348, 1967.

\bibitem[Somnath et~al.(2023)Somnath, Pariset, Hsieh, Martinez, Krause, and
  Bunne]{somnath2023aligned}
Vignesh~Ram Somnath, Matteo Pariset, Ya-Ping Hsieh, Maria~Rodriguez Martinez,
  Andreas Krause, and Charlotte Bunne.
\newblock Aligned diffusion schrodinger bridges.
\newblock \emph{arXiv preprint arXiv:2302.11419}, 2023.

\bibitem[Song et~al.(2021{\natexlab{a}})Song, Durkan, Murray, and
  Ermon]{song2021maximum}
Yang Song, Conor Durkan, Iain Murray, and Stefano Ermon.
\newblock {Maximum Likelihood Training of Score-Based Diffusion Models}.
\newblock In \emph{Advances in Neural Information Processing Systems
  (NeurIPS)}, 2021{\natexlab{a}}.

\bibitem[Song et~al.(2021{\natexlab{b}})Song, Sohl-Dickstein, Kingma, Kumar,
  Ermon, and Poole]{song2020score}
Yang Song, Jascha Sohl-Dickstein, Diederik~P Kingma, Abhishek Kumar, Stefano
  Ermon, and Ben Poole.
\newblock {Score-Based Generative Modeling through Stochastic Differential
  Equations}.
\newblock In \emph{International Conference on Learning Representations
  (ICLR)}, volume~9, 2021{\natexlab{b}}.

\bibitem[Stroock and Varadhan(1997)]{stroock1997multidimensional}
Daniel~W Stroock and SR~Srinivasa Varadhan.
\newblock \emph{Multidimensional diffusion processes}, volume 233.
\newblock Springer Science \& Business Media, 1997.

\bibitem[Sznitman(1998)]{sznitman1998brownian}
Alain-Sol Sznitman.
\newblock \emph{Brownian motion, obstacles and random media}.
\newblock Springer Science \& Business Media, 1998.

\bibitem[Tong et~al.(2020)Tong, Huang, Wolf, Van~Dijk, and
  Krishnaswamy]{tong2020trajectorynet}
Alexander Tong, Jessie Huang, Guy Wolf, David Van~Dijk, and Smita Krishnaswamy.
\newblock {TrajectoryNet: A Dynamic Optimal Transport Network for Modeling
  Cellular Dynamics}.
\newblock In \emph{International Conference on Machine Learning (ICML)}, 2020.

\bibitem[Vargas et~al.(2021)Vargas, Thodoroff, Lawrence, and
  Lamacraft]{vargas2021solving}
Francisco Vargas, Pierre Thodoroff, Neil~D Lawrence, and Austen Lamacraft.
\newblock {Solving {S}chr\"odinger Bridges via Maximum Likelihood}.
\newblock \emph{Entropy}, 23\penalty0 (9), 2021.

\bibitem[Villani(2009)]{villani2009optimal}
C{\'e}dric Villani.
\newblock \emph{Optimal transport: old and new}, volume 338.
\newblock Springer, 2009.

\bibitem[Vincent(2011)]{vincent2011connection}
Pascal Vincent.
\newblock A connection between score matching and denoising autoencoders.
\newblock \emph{Neural computation}, 23\penalty0 (7):\penalty0 1661--1674,
  2011.

\bibitem[Watson et~al.(2022)Watson, Juergens, Bennett, Trippe, Yim, Eisenach,
  Ahern, Borst, Ragotte, Milles, et~al.]{watson2022broadly}
Joseph~L Watson, David Juergens, Nathaniel~R Bennett, Brian~L Trippe, Jason
  Yim, Helen~E Eisenach, Woody Ahern, Andrew~J Borst, Robert~J Ragotte, Lukas~F
  Milles, et~al.
\newblock Broadly applicable and accurate protein design by integrating
  structure prediction networks and diffusion generative models.
\newblock \emph{bioRxiv}, pages 2022--12, 2022.

\bibitem[Yang and Uhler(2019)]{yang2018scalable}
Karren~D Yang and Caroline Uhler.
\newblock {Scalable Unbalanced Optimal Transport using Generative Adversarial
  Networks}.
\newblock \emph{International Conference on Learning Representations (ICLR)},
  2019.

\end{thebibliography}

\newpage
\appendix
\onecolumn

\section*{\Large Appendix}

In these appendices, we give more details on the stochastic processes we
investigate, better describe their links with the Schr\"odinger bridge problem, and present additional experiments. Notations are defined in
\Cref{app:sec:notation}. In \Cref{app:sec:exist-uniq-kill}, we prove the
existence of diffusions with killing under mild assumptions, by using general theoretical
results from \cite{ethier2009markov}. We then prove time-reversal formulas (\Cref{app:sec:time-reversal-killed}). Discretization schemes of diffusions with killing, as well as the link with their time reversal, are discussed in \Cref{app:sec:unbalanced_ipf}. \cref{app:sec:related-work} recalls related literature and compares it to this work. \cref{app:sec:algorithmic_details} examines instead the implementation of \usbalgo{} and supplies additional details on the more scalable heuristic estimation of $\Psi$, which we use in \usbferryman{}.
Finally, \cref{app:sec:experiment_details} includes additional experiments and details datasets, prior processes, models, hyper-parameters, and evaluation metrics.

\section{Notation}
\label{app:sec:notation}

We denote $\rmc_b^k(\rset^d, \rset)$ the set of functions which are $k$
differentiable and bounded. Similarly, we denote $\rmc_b^k(\rset^d, \rset)$ the
set of functions which are $k$ differentiable and compactly supported. The set
$\rmc_0^k(\rset^d, \rset)$ denotes the functions which are $k$ differentiable
and vanish when $\normLigne{x} \to +\infty$. For any $\msa \subset \rset^d$, we
denote $\cl{\msa}$ its closure. Similarly, we denote $\inte{\msa}$ its
interior. Finally, we define
$\partial \msa = \cl{\msa} \cap \inte{\msa}^\mathrm{c}$, the topological
boundary of $\msa$. We denote by $\cball{a}{r}$ the closed ball with center
$a \in \rset^d$ and radius $r > 0$. We denote by $\ball{a}{r}$ the associated
open ball. We denote $\hat{\rset}^d = \rset^d \cup \{ \infty \}$, the one-point
compactification of $\rset^d$. We refer to \citet{kelley2017general} for details
on this construction. We simply note that $f \in \rmc(\hat{\rset}^d)$, if
$f \in \rmc(\rset^d)$ and $f -f(\infty) \in \rmc_0(\rset^d)$ and that
$f \in \rmc^k(\hat{\rset}^d)$ for any $k \in \nset$ if the restriction of $f$ to
$\rset^d$ is in $\rmc^k(\rset^d)$ and $f \in \rmc(\hat{\rset}^d)$. We recall
that in our context, $\{ \infty \}$ will play the role of a \emph{cemetery}
state.  The space of right continuous with left limit functions on
$\coint{0,+\infty}$ with valued in $\mse$ where $\mse$ is a topological space is
denoted $\rmD(\coint{0,+\infty}, \mse)$, we refer to
\citet[p.116]{ethier2009markov} for more details on the topology of this
space. For a measure $\mu$ on a measurable space $\mse$ and $f$ a measurable
function on $\mse$ such that $\int_{\mse}\abs{f}(x) \rmd \mu(x) < +\infty$, we
denote $\mu(f) = \int_{\mse} f(x) \rmd \mu(x)$.

Next, we define a \emph{solution of the martingale problem} associated with
$\calA$, in the sense of \citet{ethier2009markov}. We only consider the case
where the solutions are right continuous. We assume that $\mse$ is a metric
space and let $\calA \subset \msf(\mse) \times \msf(\mse)$, where $\msf(\mse)$
is the space of real-valued measurable functions on $\mse$. A $\mse$-valued
right continuous stochastic process $(\bfX_t)_{t \geq 0}$ is a solution of the
martingale associated with $\calA$ if for any $(f,g) \in \calA$,
$(f(\bfX_t) - \int_0^t g(\bfX_s) \rmd s)_{t \geq 0}$ is a martingale with
respect to its own filtration. In layman terms, in the case where $\calA$ is a
function, this means that $f(\bfX_t)$ can be written as
\begin{equation}
  \textstyle f(\bfX_t) = \int_0^t \calA(f)(\bfX_s) \rmd s + \bfM_t ,
\end{equation}
where $(\bfM_t)_{t \geq 0}$ is a martingale. The notion of \emph{solution of a
  martingale problem} is associated with the notion of \emph{weak} solution to a
Stochastic Differential Equation (SDE). We refer to
\citet{stroock1997multidimensional} for an extensive study of SDEs from the point
of view of martingale problems.

\section{Existence of Diffusions with Killing}
\label{app:sec:exist-uniq-kill}

In this section, we prove the existence of killed diffusions
using the theory of infinitesimal generators. We will leverage general results
from \citep{ethier2009markov}. We begin by introducing the \emph{infinitesimal
  generator} we are going to study. Let
$\calA: \ \rmc_b^2(\rset^d, \rset) \to \rmc_c(\rset^d, \rset)$ be given for any
$f \in \rmc^2_c(\rset^d, \rset)$ and $x \in \rset^d$ by
\begin{equation}
  \label{eq:inf_gen_rd}
  \calA (f)(x) = \tfrac{1}{2} \Delta f(x) + \langle b(x), \nabla f(x) \rangle - k(x) f(x) ,
\end{equation}
with $b \in \rmc(\rset^d, \rset^d)$ called the \emph{drift function} and
$k \in \rmc(\rset^d, \rset_+)$ called the \emph{killing rate}. The regularity
assumptions are summarized in the following hypothesis.
\begin{assumption}
  \label{assum:positivity_killing}
  $b \in \rmc(\rset^d, \rset^d)$ and $k \in \rmc(\rset^d, \rset_+)$.
\end{assumption}
Contrary to a
classical diffusion associated with a It\^o Stochastic Differential Equation
(SDE), \eqref{eq:inf_gen_rd} incorporates a zero order term $x \mapsto -k(x) f(x)$. The fact
that $k$ is non-negative is key to the rest of our study. Before stating our
main result, we begin by stating a \emph{maximum principle} for $\cal A$.

\subsection{A Maximum Principle} In order to apply \cite[Theorem
5.4]{ethier2009markov}, we need to ensure that $\cal A$ satisfies a maximum
principle. Such results are ubiquitous in the analysis of Partial Differential
Equations (PDEs). In what follows, we use an extension of the Hopf maximum
principle, see \cite[Theorem 5]{protter2012maximum}. For completeness, we
provide the proof of this well-known result in our setting following
\cite{protter2012maximum}. We start with the following lemma.

\begin{lemma}
  \label{lemma:weak-maximum-principle}
  Assume \textup{\Cref{assum:positivity_killing}}. 
  Let $\msu \subset \rset^d$ be an open set and
  $f \in \rmc^2(\cl{\msu}, \rset)$. Assume that for any $x\in \msu$, we have
  $\calA (f)(x) > 0$.  Assume that there exists $x_0 \in \cl{\msu}$ such
  that $f(x_0) = \sup \ensembleLigne{f(x)}{x \in \cl{\msu}}$. Then
  $x_0 \in \partial \msu$.
\end{lemma}

\begin{proof}
  First, assume that $x_0 \in \msu$. Then, we have $\nabla f(x_0) =0 $ and
  $\nabla^2 f(x_0) \prec 0$ and therefore, $\Delta f(x_0) \leq 0$. Since
  $k \in \rmc(\rset^d, \rset_+)$, we get that $\calA (f)(x_0) \leq 0$, which is
  a contradiction. Hence $x_0 \in \partial \msu$.
\end{proof}

This lemma is also called the \emph{weak maximum principle}. Equipped with
\Cref{lemma:weak-maximum-principle}, we are now ready to prove the main result
of this section \cite[Theorem 6]{protter2012maximum}.

\begin{proposition}
  \label{prop:strong-maximum-principle}
  Assume \textup{\Cref{assum:positivity_killing}}. 
  Let $\msu \subset \rset^d$ be an open bounded connected set and
  $f \in \rmc^2(\cl{\msu}, \rset)$. Assume that for any $x \in \msu$,
  $\calA (f)(x) \geq 0$. Assume that there exists $x_0 \in \msu$ such that
  $f(x_0) = \sup \ensembleLigne{f(x)}{x \in \cl{\msu}}$. Then $f$ is constant on
  $\msu$. 
\end{proposition}

\begin{proof}
  In this proof, we are going to proceed by contradiction. We construct a
  function and a domain which contradicts
  \Cref{lemma:weak-maximum-principle}. First, we start by identifying an open
  set in $\msu$ such that at least one element of the boundary is a maximizer of
  $f$ and every element in the interior of the open set is \emph{not} a
  maximizer.

  Let $x_0 \in \msu $ such that
  $f(x_0) = M = \sup \ensembleLigne{f(x)}{x \in \cl{\msu}}$ and $x_1 \in \msu$
  such that $f(x_1) < M$. Since $\msu$ is connected, there exists
  $\upgamma \in \rmc([0,1], \msu)$ such that $\upgamma(0) = x_0$ and
  $\upgamma(1) = x_1$. Since $\upgamma([0,1])$ is compact and $\msu^\mathrm{c}$
  is closed, there exists $\delta > 0$ such that for any $t \in [0,1]$ and
  $x \not \in \msu$, $\| \upgamma(t) - x \| > \delta$. Denote
  $t_0 = \sup \ensembleLigne{f(\upgamma(t))=M}{t \in \ccint{0,1}}$. Note that
  $\ensembleLigne{f(\upgamma(t))=M}{t \in \ccint{0,1}}$ is not empty since
  $0 \in \ensembleLigne{f(\upgamma(t))=M}{t \in \ccint{0,1}}$. Let
  $t^\star \in \ocint{t_0, 1}$ such that
  $\| \upgamma(t^\star) - \upgamma(t_0) \| \leq \delta / 2$. In what follows, we
  denote $x^\star = \upgamma(t^\star)$.

  Next, we define $R \in \rmc(\ccint{0,\delta/2}, \rset)$ given for any
  $s \in \ccint{0, \delta/2}$ by
  $R(s) = \sup \ensembleLigne{f(x^\star + s z)}{\| z \| \leq 1}$, we emphasize
  that $\cball{x^\star}{\delta/2} \subset \msu$ and therefore $R$ is
  well-defined. Since $\upgamma(t_0) \in \cball{x^\star}{\delta/2}$ and
  $f(\upgamma(t_0)) = M$, we have $R(\delta/2) = M$. Note that $R$ is
  non-decreasing and denote
  $s^\star = \inf \ensembleLigne{s \in \ccint{0, \delta/2}}{R(s)=M}$. By
  continuity, there exists $x_1 \in \msu$ such that
  $\| x_1 - x^\star \| = s^\star$ and $f(x_1) = M$. In addition, by definition
  of $s^\star$, for any $x \in \ball{x^\star}{s^\star}$, $f(x) < M$.

  In the rest of the proof, we first present the easier case, where we assume
  that there exists a \emph{unique} $x_1 \in \msu$ such that
  $\| x_1 - x^\star \| = s^\star$ and $f(x_1) = M$. The general case will
  require one more construction. Let $r_1 = s^\star / 2$ and note that
  $\cball{x_1}{r_1} \subset \msu$. The set $\msv = \ball{x_1}{r_1}$ is the open
  set on which we are going to apply \Cref{lemma:weak-maximum-principle}. We are
  now going to define a function $\tilde{f} \in \rmc^2(\cl{\msv}, \rset)$ such
  that
  \begin{enumerate*}[label=(\alph*)]
  \item for any $x \in \msv$, $\calA(\tilde{f})(x) > 0$.
  \item $\tilde{f}$ admits a maximizer in $\msv$.
  \end{enumerate*}
  This will contradict \Cref{lemma:weak-maximum-principle}. For any $\alpha >0$,
  we introduce the following auxiliary function
  $w_\alpha \in \rmc^2(\ball{x^\star}{\delta}, \rset)$ given for any
  $x \in \ball{x^\star}{\delta}$ by
  \begin{equation}
    w_\alpha(x) = \exp[-\alpha \normLigne{x - x^\star}^2/2] - \exp[-\alpha (s^\star)^2/2].
  \end{equation}
  Since $\msv \subset \ball{x^\star}{\delta}$ we have that $w_\alpha$ is
  well-defined on $\msv$. In addition, using this result and the fact that for
  any $x \in \msv$,
  $\normLigne{x - x^\star} \geq \normLigne{x_1 - x^\star} - \normLigne{x_1 - x}
  \geq s^\star / 2$ we have for any $x \in \msv$ and $\alpha >0$
  \begin{align}
    \calA(w_\alpha)(x) &\geq \exp[-\alpha \normLigne{x - x^\star}^2/2](\alpha^2 \normLigne{x-x^\star}^2 + \alpha (d - C\normLigne{x-x^\star}) - C) \\
    &\geq \exp[-\alpha \normLigne{x - x^\star}^2/2](\alpha^2 (s^\star)^2/4 + \alpha (d - C\delta/2) - C) . 
  \end{align}
  where $C = \sup \ensembleLigne{\| b(x) \| + k(x)}{x \in \cl{\msu}}$. Hence,
  there exists $\alpha_0 > 0$ such that for any $x \in \msv$,
  $\calA(w_{\alpha_0})(x) > 0$. In addition, for any $x \in \partial \msv$, if
  $\normLigne{x-x^\star} \leq s^\star$, we have
  $w_{\alpha_0}(x) \leq 1 - \exp[-\alpha (s^\star)^2/2]$. If
  $\normLigne{x-x^\star} > s^\star$, then $w_{\alpha_0}(x) < 0$.

  We let
  $\xi = \inf \ensembleLigne{f(x)}{x \in \partial V \cap
    \cball{x^\star}{s^\star}}$. If $x \in \ball{x^\star}{s^\star}$ then
  $f(x) < M$. Similarly, because we have assume that there exists a unique
  $x_1 \in \msu$ such that $\normLigne{x_1 - x^\star} =s^\star$ and $f(x_1) = M$
  and $x_1 \not \in \partial \msv$, we have that if
  $x \in \partial \ball{x^\star}{s^\star} \cap \partial \msv$, $f(x) <
  M$. Hence, $\xi < M$. Let $\vareps > 0$ such that
  $\xi + \vareps(1 - \exp[-\alpha (s^\star)^2/2]) < M$ and we get that for any
  $x \in \cball{x^\star}{s^\star} \cap \partial \msv$,
  $f(x) + \vareps w_{\alpha_0}(x) < M$. In addition, for any
  $x \in \partial \msv \cap \cball{x^\star}{s^\star}^{\mathrm{c}}$, we have
  $f(x) + \vareps w_{\alpha_0}(x) \leq M - \vareps w_{\alpha_0}(x) <
  M$. Therefore, we get that for any $x \in \partial \msv$,
  $f(x) + \vareps w_{\alpha_0}(x) < M$. In addition,
  $f(x_1) + \vareps w_{\alpha_0}(x_1) = M$, since $w_{\alpha_0}(x_1) = 0$. In
  addition, we have that for any $x \in \msv$,
  $\calA (f + \vareps w_{\alpha_0})(x) > 0$. We have a contradiction with
  \Cref{lemma:weak-maximum-principle} upon letting
  $\tilde{f} = f + \vareps w_{\alpha_0}$. As emphasize above, we cannot conclude
  since we made the additional assumption that $x_1$ was the \emph{unique}
  element such that $\normLigne{x_1 - x^\star} = s^\star$ and $f(x_1) = M$.

  In the rest of the proof, we treat the general case which requires one more
  construction. Let $\tilde{s} = s^\star / 2$ and consider
  $\tilde{x} \in [x^\star, x_1]$ such that
  $\normLigne{\tilde{x} - x_1} = s^\star / 2$. The ball
  $\ball{\tilde{x}}{\tilde{s}}$ is going to replace $\ball{x^\star}{s^\star}$ in
  the proof above. Next, we consider $r_1 = \tilde{s} / 2$ and note that
  $\cball{x_1}{r_1} \subset \msu$. The set $\msv = \ball{x_1}{r_1}$ is the open
  set on which we are going to apply \Cref{lemma:weak-maximum-principle}. We are
  now going to define a function $\tilde{f} \in \rmc^2(\cl{\msv}, \rset)$ such
  that
  \begin{enumerate*}[label=(\alph*)]
  \item for any $x \in \msv$, $\calA(\tilde{f})(x) > 0$.
  \item $\tilde{f}$ admits a maximizer in $\msv$.
  \end{enumerate*}
  This will contradict \Cref{lemma:weak-maximum-principle}. For any $\alpha >0$,
  we introduce the following auxiliary function
  $w_\alpha \in \rmc^2(\ball{\tilde{x}}{\delta}, \rset)$ given for any
  $x \in \ball{\tilde{x}}{\delta}$ by
  \begin{equation}
    w_\alpha(x) = \exp[-\alpha \normLigne{x - \tilde{x}}^2/2] - \exp[-\alpha \tilde{s}^2/2] ,
  \end{equation}
  Since $\msv \subset \ball{\tilde{x}}{\delta}$ we have that $w_\alpha$ is
  well-defined on $\msv$. In addition, using this result and the fact that for
  any $x \in \msv$,
  $\normLigne{x - \tilde{x}} \geq \normLigne{x_1 - \tilde{x}} - \normLigne{x_1 - x}
  \geq \tilde{s} / 2$ we have for any $x \in \msv$ and $\alpha >0$
  \begin{align}
    \calA(w_\alpha)(x) &\geq \exp[-\alpha \normLigne{x - \tilde{x}}^2/2](\alpha^2 \normLigne{x-\tilde{x}}^2 + \alpha (d - C\normLigne{x-\tilde{x}}) - C) \\
    &\geq \exp[-\alpha \normLigne{x - \tilde{x}}^2/2](\alpha^2 (\tilde{s})^2/4 + \alpha (d - C\delta/2) - C) . 
  \end{align}
  where $C = \sup \ensembleLigne{\| b(x) \| + k(x)}{x \in \cl{\msu}}$. Hence,
  there exists $\alpha_0 > 0$ such that for any $x \in \msv$,
  $\calA(w_{\alpha_0})(x) > 0$. In addition, for any $x \in \partial \msv$, if
  $\normLigne{x-\tilde{x}} \leq \tilde{s}$, we have
  $w_{\alpha_0}(x) \leq 1 - \exp[-\alpha (\tilde{s})^2/2]$. If
  $\normLigne{x-\tilde{x}} > \tilde{s}$, then $w_{\alpha_0}(x) < 0$.

  We let
  $\xi = \inf \ensembleLigne{f(x)}{x \in \partial V \cap
    \cball{\tilde{x}}{\tilde{s}}}$. If $x \in \cball{\tilde{x}}{\tilde{s}}$ and $x \neq x_1$ then
  $f(x) < M$. Hence, $\xi < M$. Let $\vareps > 0$ such that
  $\xi + \vareps(1 - \exp[-\alpha (\tilde{s})^2/2]) < M$ and we get that for any
  $x \in \cball{\tilde{x}}{\tilde{s}} \cap \partial \msv$,
  $f(x) + \vareps w_{\alpha_0}(x) < M$. In addition, for any
  $x \in \partial \msv \cap \cball{\tilde{x}}{\tilde{s}}^{\mathrm{c}}$, we have
  $f(x) + \vareps w_{\alpha_0}(x) \leq M - \vareps w_{\alpha_0}(x) <
  M$. Therefore, we get that for any $x \in \partial \msv$,
  $f(x) + \vareps w_{\alpha_0}(x) < M$. In addition,
  $f(x_1) + \vareps w_{\alpha_0}(x_1) = M$, since $w_{\alpha_0}(x_1) = 0$. In
  addition, we have that for any $x \in \msv$,
  $\calA (f + \vareps w_{\alpha_0})(x) > 0$. We have a contradiction with
  \Cref{lemma:weak-maximum-principle} upon letting
  $\tilde{f} = f + \vareps w_{\alpha_0}$, which concludes the proof.
\end{proof}

In \Cref{prop:strong-maximum-principle}, we prove a maximum principle for the
\emph{elliptic} infinitesimal generator \eqref{eq:inf_gen_rd}. In order to
define the backward Kolmogorov equation associated with the process, we need to
consider the \emph{parabolic} infinitesimal generator instead, given for any
$f \in \rmc^2_c(\ooint{0,+\infty} \times \rset^d, \rset)$ and $t > 0$, $x \in \rset^d$ by
\begin{equation}
  \label{eq:inf_gen_rd_parabolic}
  \calA (f)(x) = \partial_t f(t,x) + \tfrac{1}{2} \Delta f(t,x) + \langle b(t,x), \nabla f(t,x) \rangle - k(x) f(t,x) .
\end{equation}
It can be shown that the generator \eqref{eq:inf_gen_rd_parabolic} also
satisfies a strong version of the maximum principle \citep{nirenberg1953strong}.

\subsection{Infinitesimal Generator}
Using \Cref{prop:strong-maximum-principle}, we are ready to define diffusion processes
with killing. The notion of \emph{solution to a martingale problem} are recalled
in \Cref{app:sec:notation}. We also denote $\hat{\calA}$, the extension of $\calA$ given in
\eqref{eq:inf_gen_rd} to the one point compactification $\hat{\rset}^d$ and defined as
follows: for any $f \in \rmc(\hat{\rset}^d)$ and $x \in \rset^d$, we have
\begin{equation}
  \label{eq:inf_gen_extended_rd}
  \hat{\calA}(f)(x) = \calA(f - f(\infty))(x) , \qquad \hat{\calA}(f)(\infty) = 0 . 
\end{equation}
The first equation can be rewritten for any  $f \in \rmc(\hat{\rset}^d)$ and $x \in \rset^d$
\begin{equation}
  \label{eq:inf_gen_rd_compact}
  \calA (f)(x) = \tfrac{1}{2} \Delta f(x) + \langle b(x), \nabla f(x) \rangle - k(x) (f(x) - f(\infty)).
\end{equation}
We have the following theorem.

\begin{theorem}
  \label{thm:existence_killed_diffusions}
  Assume \textup{\Cref{assum:positivity_killing}}. 
  For any probability measure $\nu$ on $\hat{\rset}^d$ exists
  $(\bfX_t)_{t \geq 0} \in \rmD(\coint{0,+\infty}, \hat{\rset}^d)$ solution to
  the martingale problem associated with $\hat{\calA}$ with initial condition
  $\nu$.
\end{theorem}

\begin{proof}
  The space $\rset^d$ is locally compact and separable. The generator $\calA$ is
  a linear operator and $\rmc^2_c(\rset^d, \rset)$ is dense in
  $\rmc_c(\rset^d, \rset)$. In addition, using
  \Cref{prop:strong-maximum-principle}, we have that $\calA$ satisfies the
  positive maximum principle. We conclude the proof upon applying \cite[Theorem
  5.4, p.199]{ethier2009markov}.
\end{proof}

Using \cite[Section 4, p.182]{ethier2009markov} it is also possible to show the
uniqueness (in some sense) of the process under mild assumptions. 

\section{Time Reversal of Diffusions with Killing}
\label{app:sec:time-reversal-killed}

Now that we have established the existence of a killed diffusion in
\Cref{app:sec:exist-uniq-kill}, we study its time-reversal. The main goal of this
section is to identify its associated infinitesimal generator. Before stating
these results, we provide some results on the \emph{extended} generator
\eqref{eq:inf_gen_rd_compact}. In particular, we derive the associated forward
and backward Kolmogorov equations.

\subsection{Feynman-Kac Semigroups}

We start by giving a useful representation of killed diffusions using
Feynman-Kac semigroups. We will consider the following assumptions.

\begin{assumption}
  \label{assum:regularity}
  $b \in \rmc(\rset^d, \rset)$, $k$ is bounded and there exists $\Ltt \geq 0$
  such that for any $x, y \in \rset^d$,
  $\normLigne{b(x)-b(y)} \leq \Ltt \normLigne{x-y}$.
\end{assumption}

\begin{assumption}
  \label{assum:existence_pde}
  For every $f \in \rmc(\rset^d, \rset)$ and $T > 0$, there exists
  $v \in \rmc(\ccint{0,T} \times \rset^d, \rset)$ and
  $v \in \rmc^{1,2}(\coint{0,T}\times \rset^d, \rset)$ such that for any
  $t \in \coint{0,T}$ and $x \in \rset^d$, we have $\calA(v)(t,x)=0$ with $\calA$
  given by \eqref{eq:inf_gen_rd_parabolic} and $v(0,x) = f(x)$. In addition,
  there exist $C \geq 0$ and $\mtt \in \ccint{0, 1/(2Td)}$ such that for any
  $x \in \rset^d$,
  $\sup \ensembleLigne{\|v(t,x)\|}{t \in \ccint{0,T}} \leq C \exp[\mtt
  \normLigne{x}^2]$.
\end{assumption}

Note that conditions on the existence and regularity of solutions to the
parabolic equation $\calA(v)=0$ and $\calA$ given by
\eqref{eq:inf_gen_rd_parabolic} have been established in the literature under
various assumptions on the coefficients of $\calA$, see
\citep{friedman2012stochastic} or \cite[Theorem 13.16]{dynkin1965markov}. The
following theorem is called the \emph{Feynman-Kac} representation theorem and
draws an explicit link between SDEs and PDEs. It can be seen as a generalization
of the backward Kolmogorov equation, see \cite[Theorem
7.6]{karatzas1991brownian}.

\begin{proposition}[Feynman-Kac formula]
  \label{prop:feynan-kac-formula}
  Assume \textup{\Cref{assum:positivity_killing}},
  \textup{\Cref{assum:regularity}} and \textup{\Cref{assum:existence_pde}}.  For
  any $f \in \rmc(\rset^d, \rset)$ and $T > 0$, there exists a unique solution
  $v \in \rmc(\ccint{0,T} \times \rset^d, \rset)$ and
  $v \in \rmc^{1,2}(\coint{0,T}\times \rset^d, \rset)$ to $\calA(v) =0$. In
  addition, we have that for any $t \in \ccint{0,T}$ and $x \in \rset^d$
  \begin{equation}
    \label{eq:feynman-kac-formula}
    \textstyle v(t,x) = \expeLigne{f(\bfX_T)\exp[-\int_t^{T} k(\bfX_s^0) \rmd s]} ,
  \end{equation}
  where $(\bfX_t)_{t \geq 0}$ is the unique (strong) solution to
  $\rmd \bfX_s^0 = b(\bfX_s^0) \rmd s + \rmd \bfB_s$, $\bfX_t^0=x$ and $(\bfB_s)_{s \geq 0}$ is a
  $d$-dimensional Brownian motion. 
\end{proposition}

The Feynman-Kac formula in \Cref{prop:feynan-kac-formula} will be used to establish
a backward Kolmogorov equation for \emph{killed} diffusions. We conclude this
section by giving an explicit form of the semigroup of killed diffusions under
assumptions on the killing rate. More precisely, we are going to show that these
diffusions can be seen as a \emph{reweighting} of the original unconstrained
diffusions. To do so, we need to precise the notion of semigroup. We start with
the notion of Markov kernel.

\begin{definition}
  Let $\mse, \msf$ be metric spaces.
  $\Kker: \mse \times \mcb{\msf} \to \ccint{0,1}$ is called a Markov kernel if
  for any $x \in \mse$, $\Kker(x, \cdot)$ is a probability measure and for any
  $\msa \in \mcb{\msf}$ we have that $\Kker(\cdot, \msa)$ is measurable.
\end{definition}

We are now ready to define the notion of semigroup. We refer to \cite[Chapter
4]{ethier2009markov} for a discussion.

\begin{definition}
  Let $(\Pker_t)_{t \geq 0}$ be a collection of Markov kernels with
  $\mse = \msf$ such that for any $x \in \mse$, $\Pker_0(x, \cdot) = \updelta_x$
  and for any $s, t \geq 0$, $x \in \mse$ and $\msa \in \mcb{\mse}$ we have
  \begin{equation}
    \textstyle \Pker_{t+s}(x, \msa) = \int_{\mse} \Pker_t(y, \msa) \Pker_s(x, \rmd y) .
  \end{equation}
  A semigroup on $\mse$ is said to be strongly continuous on
  $\rmc_0(\mse)$ if for any $f \in \rmc_0(\mse)$,
  $\lim_{t \to 0} \Pker_t(f) = f$ uniformly and
  $\Pker_t(f) \in \rmc_0(\mse)$.
\end{definition}

The notion of semigroup is intrinsically linked with the one of infinitesimal
generator. In particular, we define the generator associated with a semigroup
$(\Pker_t)_{t \geq 0}$ as
\begin{equation}
  \calA(f) = \lim_{t \to 0} (\Pker_t(f) - f)/t .
\end{equation}
The \emph{domain} of $\calA$, denoted $\dom(\calA)$, is the space of functions
for which the limit is well-defined. This definition of the generator justifies
the notation ``$\Pker_t = \rme^{t \cal A}$''.

In \Cref{thm:existence_killed_diffusions}, we have shown the existence of
(Markov) killed diffusions based on mild assumptions on the generator. The next
result goes further and gives an \emph{explicit} representation of the
associated semigroup. The following proposition is adapted from \cite[Theorem
1.1]{sznitman1998brownian}.

\begin{proposition}
  \label{prop:feynm-kac-semigr}
  Assume \textup{\Cref{assum:positivity_killing}} and
  \textup{\Cref{assum:regularity}}.  Let $(\bfX_t^0)_{t \geq 0}$ be the unique
  (strong) solution to $\rmd \bfX_t^0 = b(\bfX_t^0) \rmd t + \rmd \bfB_t$ and define
  $(\Pker_t)_{t \geq 0}$ such that for any $t \geq 0$ and
  $f \in \rmc_0(\rset^d)$
  \begin{equation}
    \label{eq:feynman-kac-semigroup}
    \textstyle \Pker_t(f) = \expeLigne{f(\bfX_t^0) \exp[-\int_0^t k(\bfX_s^0) \rmd s]} . 
  \end{equation}
  Then $(\Pker_t)_{t \geq 0}$ can be extended into a strongly continuous
  semigroup on $\hat{\rset}^d$. We have that $(\Pker_t)_{t \geq 0}$ admits a
  generator $\hat{\calA}$ with $\rmc_c^2(\rset^d) \subset \dom(\hat{\calA})$ and $\hat{\calA}$
  given by \eqref{eq:inf_gen_extended_rd}. In addition, we have the following
  \textbf{perturbation formulas} for any $t \geq 0$, $x\in \rset^d$ and $f \in \rmc_0(\rset^d)$\textup{:}
  \begin{align}
    \label{eq:perturbation_identities}
    \Pker_t(x, f) &= \textstyle \Pker_t^0(x,f) - \int_0^t \Pker_s^0(x, k \Pker_{t-s}(f)) \rmd s , \\
    &= \textstyle \Pker_t^0(x,f) - \int_0^t \Pker_s(x, k \Pker_{t-s}^0(f)) \rmd s , 
  \end{align}
  where $(\Pker_t^0)_{t \geq 0}$ is the semigroup associated with
  $(\bfX_t^0)_{t \geq 0}$. Finally, there exists a Markov process
  $(\bfX_t)_{t \geq 0}$ on $\hat{\rset}^d$ associated with the $\hat{\rset}^d$
  extension of $(\Pker_t)_{t \geq 0}$.
\end{proposition}

\begin{proof}
  We first begin by proving \eqref{eq:perturbation_identities}. We have for any
  $t \geq 0$
  \begin{align}
    \textstyle \exp[-\int_0^t k(\bfX_s^0) \rmd s] &= \textstyle 1 - \int_0^t k(\bfX^0_s) \exp[-\int_s^t k(\bfX^0_u) \rmd u] \rmd s \\
    &= \textstyle  1 - \int_0^t k(\bfX^0_s) \exp[-\int_0^s k(\bfX^0_u) \rmd u] \rmd s .
  \end{align}
  Hence, for any $t \geq 0$, $x\in \rset^d$ and $f \in \rmc_0(\rset^d)$, we have
  \begin{align}
    \Pker_t(x,f) &= \textstyle \expeLigne{f(\bfX_t^0)} - \int_0^t \expeLigne{f(\bfX_t^0) k(\bfX^0_s) \exp[-\int_s^t k(\bfX^0_u) \rmd u]} \rmd s \\
                 &= \textstyle \expeLigne{f(\bfX_t^0)} - \int_0^t \expeLigne{k(\bfX^0_s) \CPELigne{f(\tilde{\bfX}_{t-s}^0)  \exp[-\int_0^{t-s} k(\tilde{\bfX}^0_u) \rmd u]}{\tilde{\bfX}_0^0 = \bfX_s^0}} \rmd s \\
    &= \textstyle \expeLigne{f(\bfX_t^0)} - \int_0^t \expeLigne{k(\bfX^0_s) \Pker_{t-s}(\bfX^0_s, f)} \rmd s ,
  \end{align}
  which concludes the proof of \eqref{eq:perturbation_identities}. Since
  $k \in \rmc_b(\rset^d, \rset)$, we have for any $f \in \rmc_0(\rset^d)$ and
  $x \in \rset^d$, $(s, t) \in \Pker_s(x, k \Pker_{t-s}^0(f))$ which is
  continuous and therefore we get that for any $f \in \rmc_0(\rset^d)$ and
  $x \in \rset^d$
  \begin{equation}
    \label{eq:derivative_perturbation}
    \textstyle \lim_{t \to 0} \tfrac{1}{t} \int_0^t \Pker_s^0(x, k \Pker_{t-s}(f)) = k(x) f(x) . 
  \end{equation}
  We extend $(\Pker_t)_{t \geq 0}$ to $\hat{\rset}^d$ by denoting for any
  $f \in \rmc(\hat{\rset}^d), x\in \rset^d$ and $t \geq 0$
  \begin{equation}
    \textstyle \hat{\Pker}_t(x, f) = \Pker_t(x, f) + f(\infty) (1 - \exp[-\int_0^t k(\bfX_s^0) \rmd s]) . 
  \end{equation}
  In addition, we let $\Pker_t(\infty, f) = f(\infty)$.  Note that
  $(\hat{\Pker}_t)_{t \geq 0}$ defines a strongly continuous semigroup on
  $\hat{\rset}^d$.  Using \eqref{eq:derivative_perturbation} and the fact that
  $\rmc_c^2(\rset^d) \subset \dom(\calA^0)$, where $\calA^0$ is the generator
  associated with $(\bfX_t^0)_{t \geq 0}$, we get that the generator of
  $(\hat{\Pker}_t)_{t \geq 0}$ is given for any $f \in \rmc_c^2(\rset^d)$ and $x \in \rset^d$ by
  \begin{equation}
    \calA (f)(x) = \calA^0(f)(x) - k(x)f(x) .
  \end{equation}
  We conclude the proof upon using \cite[Theorem 1.1, p.157]{ethier2009markov}.
\end{proof}

For an introduction to killed diffusions we refer to
\citep{oksendal2013stochastic}, see also
\citep{karlin1981second,blumenthal2007markov}.

Let us comment on the form of \eqref{eq:feynman-kac-semigroup} in
\Cref{prop:feynm-kac-semigr}. Formally speaking, this formula indicates that in
order to integrate a function $f$ with respect to the killed diffusion we can
consider an integration w.r.t. to the unconstrained diffusion
$(\bfX_t^0)_{t \geq 0}$ with a \emph{loss of mass} given by the exponential term
$ \exp[-\int_0^t k(\bfX_s^0) \rmd s]$. Consider the limit case where
$k = +\infty$ on some domain $\msa$ and $k=0$ otherwise. Then
\eqref{eq:feynman-kac-semigroup} becomes
$\Pker_t(f) = \expeLigne{f(\bfX_t^0) \mathbf{1}_{t < \tau_{\msa}} }$, where
  $\tau_\msa = \inf \ensembleLigne{s \geq 0}{\bfX_s \in \msa}$. 

  Equipped with the notion of Feynman-Kac semigroup, we can rewrite
  \eqref{eq:feynman-kac-formula} in \Cref{prop:feynan-kac-formula} as
  $v(t,x) = \Pker_{T-t}(x, f)$. This remark is at the basis of the backward
  Kolmogorov formula for killed diffusions.

  \subsection{Forward and Backward Kolmogorov Equations}
  In order to derive the time-reversal of a stochastic process it is useful to
  define its time-reversal \citep{haussmann1986time}. In what follows, we denote
  $(\bfX_t)_{t \geq 0}$, the killed diffusion, i.e., the process associated with
  $(\Pker_t)_{t \geq 0}$.

  \begin{proposition}
    \label{prop:backward-kolmogorov}
    Assume \textup{\Cref{assum:positivity_killing}},
    \textup{\Cref{assum:regularity}} and
    \textup{\Cref{assum:existence_pde}}. Let $(\bfX_t)_{t \geq 0}$ be given by
    \Cref{prop:feynm-kac-semigr}. Then, for any $f \in \rmc_0(\rset^d)$,
    $T, t \geq 0$ with $T \geq t$ and $x \in \rset^d$
    \begin{equation}
      \label{eq:backward_forward}
      \partial_t \Pker_t(x, f) = \calA(\Pker_t(\cdot, f))(x) , \qquad \partial_t \Pker_{T-t}(x,f) = - \calA(\Pker_{T-t}(\cdot,f))(x) ,
    \end{equation}
    with $\calA$ given by \eqref{eq:inf_gen_rd_compact}.
  \end{proposition}

  The first part of \eqref{eq:backward_forward} is referred to as the
  \emph{forward Kolmogorov equation} while the second part is referred to as the
  \emph{backward Kolmogorov equation}. The proof is a direct consequence of
  \Cref{prop:feynm-kac-semigr}. Note that under \Cref{assum:existence_pde} we
  have additional regularity assumptions on $(t,x) \mapsto \Pker_{T-t}(x,f)$ and
  therefore its evolution can be made explicit using the expression of $\calA$.

  \subsection{Reversal of the Fokker-Planck Equation}
  In this paragraph, we provide a heuristic derivation of the time-reversal of
  the killed diffusion process. Under mild assumptions, we get that $\Pker_t$
  admits a density w.r.t the Lebesgue measure denoted $p_t$ for any $t >
  0$. Note that $\int_{\rset^d} p_t(x) \rmd x \leq 1$. In addition, we have that
  \begin{equation}
    \partial_t p_t(x) = -\mathrm{div}(b(t, \cdot) p_t)(x) + \tfrac{1}{2} \Delta p_t(x) -k(x)p_t(x) .
  \end{equation}
  Note that the right-hand side of the previous equation can formally be
  identified with the dual operator $\calA^\star$. Now, considering the
  time-reversal of the previous equation, we get
  \begin{equation}
    \partial_t p_{T-t}(x) = -\mathrm{div}(\{-b(t, \cdot) + \nabla \log p_{T-t}\}p_{T-t})(x) + \tfrac{1}{2} \Delta p_{T-t}(x) +k(x)p_{T-t}(x) .
  \end{equation}  
  Note that the \emph{non-negative} term $k(x)p_t(x)$ is turned into a
  \emph{non-positive} term $-k(x)p_{T-t}(x)$. This suggests that the
  \emph{death} behavior of the forward process is turned into a \emph{birth}
  behavior for the backward process. We will make this statement precise in the next
  paragraph. This is in contrast with common results which state that if the
  forward process is in a certain class then so is the backward process. For
  instance, in the case of a reflected forward process, the backward is also
  reflected \citep{lou2023reflected,fishman2023diffusion}.
  
  \subsection{Time Reversal of Diffusion With Births}
  In the previous paragraph, we provided the time-reversal of the \emph{dual}
  generator in order to derive the backward evolution of the density $p_t$. In
  this section, we follow the approach of \cite{haussmann1986time} to derive a
  \emph{pathwise} time-reversal. Doing so we will directly obtain the
  time-reversal of the generator (and not its dual). We start by recalling the
  expression of the extended infinitesimal generator,
  \eqref{eq:inf_gen_extended_rd}. For any $f \in \rmc^2(\hat{\rset}^d)$ and
  $x \in \hat{\rset}^d$ we have
  \begin{equation}
    \hat{\calA}(f)(x) = [\langle b(x), \nabla f(x) \rangle + \tfrac{1}{2} \Delta f(x) - k(x)(f(x) - f(\infty))] \mathbf{1}_{\rset^d}(x) . 
  \end{equation}
  In what follows, we let $(\bfX_t)_{t \geq 0}$ given by
  \Cref{prop:feynm-kac-semigr}. We denote $S_t = \Pbb[\bfX_t \in \rset^d]$ for
  any $t \geq 0$. We will also make the following assumption.
  \begin{assumption}
    \label{assum:density}
    For any $t \geq 0$, the measure $\mu_t$ on $\rset^d$ given for any
    $f \in \rmc_c(\rset^d)$ by
    $\mu_t[f] = \expeLigne{f(\bfX_t) \mathbf{1}_{\rset^d}(\bfX_t)}$ admits a density
    $p_t$ w.r.t. the Lebesgue measure. In addition
    $(t,x) \mapsto p_t(x) \in \rmc_b^\infty(\ooint{0,+\infty} \times \rset^d,
    \ooint{0,+\infty})$. Finally, we assume that $S_t < 1$ for any $t > 0$. 
  \end{assumption}

  The following lemma is central to establish our result.
  \begin{lemma}
    \label{lemma:discrete_ipp}
    Assume \textup{\Cref{assum:positivity_killing}},
    \textup{\Cref{assum:regularity}} and \textup{\Cref{assum:density}}. Then for
    any $t > 0$, $h : \rset^d \times \{\infty\} \to \rset$ and 
    $g: \ \{\infty\} \to \rset$ measurable and bounded, we have
    \begin{equation}
      \textstyle
      \expeLigne{\mathbf{1}_{\rset^d}(\bfX_t) h(\bfX_t, \infty) g(\infty)} = \expeLigne{\mathbf{1}_{\infty}(\bfX_t) \int_{\rset^d} h(x, \bfX_t) p_t(x) \rmd x / (1-S_t)  g(\bfX_t)} . 
    \end{equation}
  \end{lemma}

  \begin{proof}
    Let $t > 0$, $h : \rset^d \times \{\infty\} \to \rset$ and
    $g: \ \{\infty\} \to \rset$ measurable and bounded. First, we have
    \begin{align}
      \textstyle \expeLigne{\mathbf{1}_{\rset^d}(\bfX_t) h(\bfX_t, \infty) g(\infty)} &= \textstyle \int_{\rset^d} h(x, \infty) g(\infty) p_t(x) \rmd x \\
      &= \expeLigne{\mathbf{1}_{\infty}(\bfX_t) \textstyle \int_{\rset^d} h(x, \bfX_t) g(\bfX_t) p_t(x) \rmd x} / (1-S_t) , 
    \end{align}
    which concludes the proof.
  \end{proof}

  Based on this lemma, we are now ready to state the our time-reversal result,
  extending the approach of \cite{haussmann1986time}. We make the following
  assumption, which ensures that integration by part is valid in our setting.

  \begin{assumption}
    \label{assumption:integration_by_part}
    For any $f, g \in \rmc^2(\hat{\rset}^d)$ we have
    \begin{equation}
      \expeLigne{\mathbf{1}_{\rset^d}(\bfX_t) \langle \nabla f(\bfX_t), \nabla g(\bfX_t) \rangle} = -  \expeLigne{\mathbf{1}_{\rset^d}(\bfX_t) g(\bfX_t) (\Delta f(\bfX_t) + \langle \nabla \log p_t(\bfX_t), \nabla f(\bfX_t) \rangle)} . 
    \end{equation}
  \end{assumption}

  We emphasize that \Cref{assumption:integration_by_part} finds sufficient
  condition on the parameters of the unconstrained diffusion such that this
  integration by part formula is true in the unconstrained setting. We leave the
  extension of such results to the killed case for future work.  We are now
  ready to state our time-reversal formula.

  \begin{proposition}
    \label{prop:time_reversal_kill}
    Assume \textup{\Cref{assum:positivity_killing}},
    \textup{\Cref{assum:regularity}}, \textup{\Cref{assum:density}} and
    \textup{\Cref{assumption:integration_by_part}}. Let $(\bfX_t)_{t \geq 0}$
    given by \Cref{prop:feynm-kac-semigr}, $T > 0$ and consider
    $(\bfY_t)_{t \in \ccint{0,T}} = (\bfX_{T-t})_{t \in \ccint{0,T}}$. Then
    $(\bfY_t)_{t \in \ccint{0,T}}$ is solution to the martingale problem
    associated with $\hat{\calR}$, where for any $f \in \rmc^2(\hat{\rset}^d)$,
    $t \in \ooint{0,T}$ and $x \in \hat{\rset}^d$ we have
    \begin{align}
      \textstyle \calR(f)(t,x) &= [\langle -b(x) + \nabla \log p_{T-t}(x), \nabla f(x) \rangle + \tfrac{1}{2} \Delta f(x)] \mathbf{1}_{\rset^d}(x) \nonumber\\
      \label{eq:time_reversal_kill}
      & \qquad \textstyle + \int_{\rset^d} p_{T-t}(\tilde{x}) k(\tilde{x})  (f(\tilde{x}) - f(\infty)) \rmd \tilde{x} / (1-S_{T-t}) \mathbf{1}_{\infty}(x) . 
    \end{align}
  \end{proposition}

  \begin{proof}
    Let $f, g \in \rmc^2(\hat{\rset}^d)$. We are going to show that for any $s, t \in \ccint{0,T}$ with $t \geq s$
    \begin{equation}
      \textstyle \expeLigne{(f(\bfY_t) - f(\bfY_s))g(\bfY_s) } = \expeLigne{g(\bfY_s) \int_s^t \calR(f)(u, \bfY_u) \rmd u } . 
    \end{equation}
    This is equivalent to show that for any $s, t \in \ccint{0,T}$ with $t \geq s$
    \begin{equation}
      \textstyle \expeLigne{(f(\bfX_t) - f(\bfX_s))g(\bfX_t) } = \expeLigne{-g(\bfX_t) \int_s^t \calR(f)(u, \bfX_u) \rmd u } . 
    \end{equation}
    Let $s, t \in \ccint{0,T}$, with $t \geq s$. In what follows, we denote for
    any $u \in \ccint{0,t}$ and $x \in \hat{\rset}^d$,
    $g(u,x) = \CPELigne{g(\bfX_t)}{\bfX_u=x}$. Using
    \Cref{prop:backward-kolmogorov}, we have that for any $u \in \ccint{0,t}$
    and $x \in \hat{\rset}^d$, $\partial_u g(u,x) + \hat{\calA}(g)(u,x) = 0$,
    i.e.~ $g$ satisfies the backward Kolmogorov equation. For any
    $u \in \ccint{0,t}$ and $x \in \hat{\rset}^d$, we have
    \begin{align}
      \hat{\calA}(fg)(u, x) &= \partial_u g(u,x) f(x) + (\langle b(x), \nabla g(u,x) \rangle + \tfrac{1}{2} \Delta g(u,x)) f(x) \mathbf{1}_{\rset^d}(x) \\
                            &\qquad +(\langle b(x), \nabla f(x) \rangle + \tfrac{1}{2} \Delta f(x)) g(u,x) \mathbf{1}_{\rset^d}(x) + \mathbf{1}_{\rset^d}(x) \langle \nabla f(x), \nabla g(u,x) \rangle \\
                            &\qquad + \mathbf{1}_{\rset^d} (f(x)g(u,x) - f(\infty)g(u,\infty)) \\
&= \partial_u g(u,x) f(x) + \hat{\calA}(g)(u,x) f(x)+ \mathbf{1}_{\rset^d}(x) (f(x) - f(\infty))g(u,\infty) \\
                            &\qquad +(\langle b(x), \nabla f(x) \rangle + \tfrac{1}{2} \Delta f(x)) g(u,x) \mathbf{1}_{\rset^d}(x) + \mathbf{1}_{\rset^d}(x) \langle \nabla f(x), \nabla g(u,x) \rangle \\
                            &= (\langle b(x), \nabla f(x) \rangle + \tfrac{1}{2} \Delta f(x)) g(u,x) \mathbf{1}_{\rset^d}(x) + \mathbf{1}_{\rset^d}(x) \langle \nabla f(x), \nabla g(u,x) \rangle \\
& \qquad + \mathbf{1}_{\rset^d}(x) (f(x) - f(\infty))g(u,\infty) . \label{eq:decomposition}
    \end{align}
    Using \Cref{assumption:integration_by_part}, we have that for any $u \in \ccint{0,t}$
    \begin{align}
      &\expeLigne{\mathbf{1}_{\rset^d}(\bfX_u) \langle \nabla f(\bfX_u), \nabla g(u, \bfX_u) \rangle} \\
      & \qquad \qquad = -  \expeLigne{\mathbf{1}_{\rset^d}(\bfX_u) g(u, \bfX_u) (\Delta f(\bfX_u) + \langle \nabla \log p_u(\bfX_u), \nabla f(\bfX_u) \rangle)} . \label{eq:ipp_1}
    \end{align}
    In addition, using \Cref{lemma:discrete_ipp}, we have that for any
    $u \in \ccint{0,t}$
        \begin{equation}
      \textstyle
      \expeLigne{\mathbf{1}_{\rset^d}(\bfX_u) h(\bfX_u, \infty) g(u, \infty)} = \expeLigne{\mathbf{1}_{\infty}(\bfX_u) \int_{\rset^d} h(x, \bfX_u) p_u(x) \rmd x / (1-S_u)  g(u, \bfX_u)} . \label{eq:ipp_2}
    \end{equation}
    Combining \eqref{eq:decomposition}, \eqref{eq:ipp_1} and \eqref{eq:ipp_2}, we get that
    \begin{equation}
      \expeLigne{\hat{\calA}(fg)(u, \bfX_u)} = -\calR(f)(u, \bfX_u)g(u, \bfX_u) . 
    \end{equation}
    In addition, we have
    \begin{align}
      \expeLigne{(f(\bfX_t) -f(\bfX_s))g(\bfX_t)} &= \expeLigne{g(t, \bfX_t)f(\bfX_t) -f(\bfX_s)g(s, \bfX_s)} \\
                                                  &\textstyle = \expeLigne{\int_s^t \calA(fg)(u, \bfX_u) \rmd u } \\
      &\textstyle = - \expeLigne{\int_s^t \calR(f)(u, \bfX_u)g(u, \bfX_u) \rmd u } = - \expeLigne{g(\bfX_t) \int_s^t \calR(f)(u, \bfX_u) \rmd u } , 
    \end{align}
    which concludes the proof.
  \end{proof}

  \Cref{prop:time_reversal_kill} shows that the time-reversal of a diffusion
  with killing is a diffusion with the birth. The update can be split into two
  parts. First, the drift of the diffusion is turned into
  $(t,x) \mapsto -b(x) + \nabla \log p_{T-t}(x)$. This is in accordance with
  classical time-reversal results for unconstrained diffusions. The main novelty
  of \Cref{prop:time_reversal_kill} resides in the change of the killing
  procedure to a birth procedure. The killing rate $x \mapsto k(x)$ is turned
  into a birth density $(t, x) \mapsto k(x) p_t(x) / 1-S_t$. This birth density
  means that, in the time-reversal process, we give birth to a particle near $x$
  if $k(x) p_t(x)$ is large. This means that two could conditions must be met
  for a particle to have a high likelihood to be born
  \begin{enumerate*}[label=(\alph*)]
  \item the killing rate must be large, i.e.~birth can only occur at places
    where particles died in the original process.
  \item the density $p_t$ must be large, i.e.~birth can only occur at places
    which are visited by the original process.
  \end{enumerate*}
  
  \subsection{Time Reversal of Diffusions with Births}
  \label{sec:time-reversal-birth}

  In the Schr\"odinger bridge setting, we iterate on the time-reversal
  procedure. This means that we need to consider the time-reversal of a
  \emph{birth} process. In this section, we provide the heuristics to derive
  such a time-reversal. We assume that there exists a birth process
  $(\bfX_t)_{t \geq 0}$ taking values in $\hat{\rset}^d$ such that
  $(\bfX_t)_{t \geq 0}$ is Markov and solution to the martingale problem
  associated with $\hat{\calA}$ where for any $f \in \rmc^2(\hat{\rset}^d)$ and
  $x \in \hat{\rset}^d$ we have 
  \begin{equation}
    \textstyle \hat{\calA}(f)(x) = (\langle b(x), \nabla f(x) \rangle + \tfrac{1}{2}\Delta f(x)) \mathbf{1}_{\rset^d}(x) + \int_{\rset^d} (f(\tilde{x}) - f(\infty)) q(\tilde{x}) \mathbf{1}_{\infty}(x) . 
  \end{equation}
  Similarly to the previous section, we consider \Cref{assum:density} and
  therefore, we have the existence (and regularity) of a density $p_t$. Under
  similar assumptions to \Cref{prop:time_reversal_kill}, we have that its
  time-reversal is also the solution to a martingale problem with generator $\hat{\calR}$ 
  given for any $f \in \rmc^2(\hat{\rset}^d)$, $t \in \ccint{0,T}$ and
  $x \in \hat{\rset}^d$ by
  \begin{align}
    \hat{\calR}(f)(t,x) &= (\langle -b(x) + \nabla \log p_{T-t}(x), \nabla f(x) \rangle + \tfrac{1}{2}\Delta f(x))  \mathbf{1}_{\rset^d}(x) \nonumber\\
    \label{eq:time_reversal_birth}
    & \qquad - (1-S_{T-t}) q(x) / p_{T-t}(x) (f(x) - f(\infty)) \mathbf{1}_{\rset^d}(x) . 
  \end{align}
  Therefore, the time-reversal of a diffusion process with birth is a diffusion
  process with death. Similarly to \Cref{prop:time_reversal_kill}, the drift is
  updated as in the unconstrained case, i.e., $x \mapsto b(x)$ is replaced by
  $(t,x) \mapsto -b(x) + \nabla \log p_{T-t}(x)$. The birth measure $x \mapsto q(x)$
  is changed into the killing rate $(t,x) \mapsto (1-S_{T-t}) q(x) / p_{T-t}(x)$.
  
  \paragraph{Entropic time-reversal with jumps.} In fact
  \eqref{eq:time_reversal_kill} and \eqref{eq:time_reversal_birth} can be
  inferred from the results of \cite{conforti2022time}. In this work, the
  authors consider a pure jump process, i.e., no diffusion, with jumps in
  $\rset^d$. In that case the infinitesimal generator $\calA$ is given for any
  $f \in \rmc_c^2(\rset^d)$ and $x \in \rset^d$
  \begin{equation}
    \textstyle \calA(f)(x) = \langle b(x), \nabla f(x) \rangle + \int_{\rset^d} (f(y) - f(x)) \Jbb(x, \rmd y) , 
  \end{equation}
  where $\Jbb$ is a Markov kernel. Under mild entropic assumptions, the authors
  prove that the time-reversal is also a process with pure jumps associated with
  the infinitesimal generator $\calR$ is given for any $f \in \rmc_c^2(\rset^d)$,
  $x \in \rset^d$ and $t \in \ccint{0,T}$ by 
  \begin{equation}
    \textstyle \calR(f)(x) = -\langle b(x), \nabla f(x) \rangle + \int_{\rset^d} (f(y) - f(x)) \Rbb(x, \rmd y) , 
  \end{equation}
  where $\Rbb$ is a Markov kernel such that for any $f \in \rmc_c^2(\rset^d \times \rset^d)$
  \begin{equation}
    \textstyle \int_{\rset^d} \int_{\rset^d} f(x,y) p_t(x) \Jbb(x, \rmd y) \rmd x = \int_{\rset^d} \int_{\rset^d} f(x,y) p_t(y) \Rbb(y, \rmd x) \rmd y . 
  \end{equation}
  This equation is sometimes called the \emph{flux equation}, see \cite[Equation
  (1.1)]{conforti2022time}. Our setting is different in that we also consider a
  diffusive part, i.e.~ there is an additional term $\tfrac{1}{2} \Delta f$ in
  our generator and our generator is defined on the extended space
  $\hat{\rset}^d$ in order to properly account for the \emph{killing} and
  \emph{birth} processes.

  However, assuming that we can extend the flux equation to our setting, we have
  in the case of a death process
  $\Jbb(x, \rmd y) = k(x) \updelta_\infty(\rmd y)$. Hence, we have that
  \begin{equation}
   \Rbb(x,\rmd y) (1-S_t) \updelta_\infty(x) = \Jbb(y,  \rmd x) p_t(y) ,
 \end{equation}
 and therefore $\Rbb(x, \rmd y) = k(y) p_t(y) / (1-S_t) \updelta_\infty(y)$,
 which is the update we have in \eqref{eq:time_reversal_kill}. Similarly, in the
 case of a birth process $\Jbb(x, \rmd y) = \updelta_\infty(x) q(y) \rmd y$. Hence, we have that
  \begin{equation}
   \Rbb(x,\rmd y) p_t(x) \updelta_\infty(x) = \Jbb(y,  \rmd x) (1-S_t) ,
 \end{equation}
 and therefore $\Rbb(x, \rmd y) = q(x) (1-S_t) / p_t(x) \updelta_\infty(y)$,
 which is the update we have in \eqref{eq:time_reversal_birth}. Finally, we
 emphasize that obtaining a ratio of the form $p_t(x) / \Pbb(\bfX_t = \infty)$
 is similar to what has been obtained in the case of \emph{discrete} state-space
 time-reversal, see \citep{campbell2022continuous,benton2022denoising} for instance.

\subsection{Duality}

\newcommand{\R}{\mathbb{R}}
\newcommand{\dx}{\rmd x}

In this section, we show that the system \eqref{eq:system_unrolled} can be understood as a pair of Kolmogorov forward and backward equations. 

Define $\textcolor{MyBlue}{\chi}, \textcolor{MyRed}{\hat{\chi}}: \ \ccint{0,T} \times \hat{\rset}^d \to \rset$, for any $t \in \ccint{0,T}$ and $x \in \hat{\rset}^d$, as
\begin{equation}
  \textcolor{MyBlue}{\chi_t(\infty)} = \textcolor{MyBlue}{\Psi_t} , \quad \textcolor{MyBlue}{\chi_t(x)} = \textcolor{MyBlue}{\varphi_t(x)} \qquad \text{and} \qquad \textcolor{MyRed}{\hat{\chi}_t(\infty)} =  \textcolor{MyRed}{\hat{\Psi}_t} , \quad \textcolor{MyRed}{\hat{\chi}_t(x)} = \textcolor{MyRed}{\hat{\varphi_t}(x)} . 
\end{equation}Recall the formula of a generator with killing:
\begin{equation}
(\hat{\calK}^0 f )(x) \coloneqq \left(\langle b(x), \nabla f(x) \rangle + \frac{1}{2}\Delta f(x) - k(x)\left(f(x)-f(\infty) \right)  \right) \mathbf{1}_{\rset^d}(x).
\end{equation}
Then it is evident that the equations governing $ \textcolor{MyBlue}{\Psi_t}$ and $\textcolor{MyBlue}{\varphi_t(x)}$ can be rewritten as:
\begin{equation}
\partial_t \textcolor{MyBlue}{\chi_t} = - \hat{\calK}^0 \textcolor{MyBlue}{\chi_t}
\end{equation}which is the usual Kolmogorov \emph{backward} equation. On the other hand, we will show below that equations governing $\textcolor{MyRed}{\hat{\Psi}_t}$ and $\textcolor{MyRed}{\hat{\varphi_t}(x)}$ can be interpreted as the Kolmogorov \emph{forward} equation. To this end, for any two functions $f,g: \hat{\rset}^d \to \R$, define the usual $L_2$-inner product as:
\begin{equation}
\langle g,  f \rangle \coloneqq \int_{\R^d} f(x)g(x) \dx + f(\infty)\cdot g(\infty).
\end{equation}
For any operator $\calK$, its dual $\calK^\star$ with respect to $\langle\cdot,\cdot\rangle$ is defined as the operator satisfying the relation:
\begin{equation}
\label{eq:dual-def}
\forall f,g, \quad \langle g, \calK  f \rangle  = \langle  \calK^\star g,  f \rangle .
\end{equation}

Using the standard integration by part formulas, we get
\begin{align}
\langle g, \hat{\calK}^0  f \rangle &= \int_{\R^d} g(x)\left(\langle b(x), \nabla f(x) \rangle + \frac{1}{2}\Delta f(x) - k(x)\left(f(x)-f(\infty) \right)  \right) \dx \\
&=  \int_{\rset^d}  \left( \langle g(x)b(x), \nabla f(x) \rangle + \frac{1}{2} g(x)\Delta f(x) - k(x)g(x)f(x) + k(x)g(x)f(\infty) \right)\dx \\
&= \int_{\rset^d}  \left( -\mathrm{div} \big(b(x)g(x) \big)  + \frac{1}{2} \Delta g(x) - k(x)g(x) \right)f(x) \dx +   \left(\int_{\rset^d} k(x)g(x) \dx \right) \cdot f(\infty) \\
&\eqqcolon  \langle  \hat{\calK}^{0,\star} g,  f \rangle 
\end{align}where the dual of $\hat{\calK}^0$ is given by
\begin{equation}
(\hat{\calK}^{0,\star}g)(x) \coloneqq  \left( -\mathrm{div} \big(b(x)g(x) \big)  + \frac{1}{2} \Delta g(x) - k(x)g(x) \right) \mathbf{1}_{\rset^d} + \left(\int_{\rset^d} k(x)g(x) \dx \right)\mathbf{1}_{\infty}.
\end{equation}
With these definitions, it holds that $\textcolor{MyRed}{\hat{\chi}}$ satisfies the \emph{Forward Kolmogorov} equation:
\eqref{eq:system_unrolled} is equivalent to
\begin{equation}
\textstyle
  \partial_t \textcolor{MyRed}{\hat{\chi}_t} = \hat{\calK}^{0,\star} \textcolor{MyRed}{\hat{\chi}_t} .
\end{equation}

\section{Unbalanced IPF}
\label{app:sec:unbalanced_ipf}

In this section, we state the \acp{SDE} describing the evolution of log-potentials $\log \varphi$ and $\log \hat \varphi$ along backward and forward trajectories of the optimal solution to the unbalanced \ac{SB} problem. We then further comment on the sampling strategies used for birth processes.
We provide formulas which generalize the prior process \eqref{eq:diffusion} to diffusions with arbitrary (scalar) diffusivity $\diffusivity: [0,1] \rightarrow \rset_+$.
 
\subsection{SDEs for Log-Potentials}
To simplify the notation, in this and subsequent sections, we denote the Neural Networks parametrizing $\log \varphi$ and $\log \hat \varphi$ as $\Ynn{F}$ and $\Ynn{B}$, and their score as:
\begin{align}
    \Znn{F}(t, x_t) = \diffusivity \gradx \Ynn{F}(i, x_t) \\
    \Znn{B}(t, x_t) = \diffusivity \gradx \Ynn{B}(i, x_t) 
\end{align}

\paragraph{Forward Dynamics}
When accounting for time-varying diffusivity, the forward diffusion process \eqref{eq:diffusion} reads: 
\begin{equation}
\label{eq:gen_f_diffusion}
    \diff{\fvers{X}_t} = (\drift + \diffusivity \Znn{F}) \diff{t} + \diffusivity \diff{W_t}
\end{equation}

Along paths $(\fvers{X}_t)_t$, the log-potentials evolve as:
\begin{align}
\label{eq:usb_sde_y_f}
    \begin{cases}
        \diff{\Y{F}}_t = \left( \frac{1}{2} \norm{\Z{F}_t}^2 + \killing_t \left(1 - \frac{\discrfact{F}}{\exp{\Y{F}_t}} \right) \right) \diff{t} + \ip{\Z{F}_t}{\diff{W}_t} \\
        \diff{\Y{B}}_t = \left( \frac{1}{2} \norm{\Z{B}_t}^2 + \divx(-\drift + \diffusivity \Z{B}_t) + \ip{\Z{B}_t}{\Z{F}_t} - \killing_t \right) \diff{t} + \ip{\Z{B}_t}{\diff{W}_t} 
    \end{cases}
\end{align}

\paragraph{Backward Dynamics}
The time-reversal of \eqref{eq:gen_f_diffusion} is:
\begin{equation}
\label{eq:gen_b_diffusion}
    \diff{\bvers{X}_t} = (\drift - \diffusivity \Znn{B}) \diff{t} + \diffusivity \diff{W_t}
\end{equation}
The log-potentials along $(\bvers{X}_t)_t$ follow, instead, the SDEs:
\begin{align}
\label{eq:usb_sde_y_b}
    &\begin{cases}
        \diff{\Y{F}}_s = \left( \frac{1}{2} \norm{\Z{F}_s}^2 + \diffusivity \divx \Z{F}_s + \ip{\Z{F}_s}{\Z{B}_s} + \killing_s \left( \frac{\discrfact{F}}{\exp{\Y{F}_s}} - 1 \right) \right) \diff{s} + \ip{\Z{F}_s}{\diff{W}_s} \\
        \diff{\Y{B}}_s = \left(\divx\drift + \frac{1}{2} \norm{\Z{B}_s}^2 + \killing_s  \right) \diff{s} + \ip{\Z{B}_s}{\diff{W}_s}
    \end{cases}
\end{align}

\subsection{Discretization of Diffusions with Killing and Birth}
In \cref{alg:usb_sampling}, we need to sample both from the forward (function \textsc{Sample-Forward}) and backward dynamics (function \textsc{Sample-Backward}). We discretize Eqs. \eqref{eq:gen_f_diffusion} and \eqref{eq:gen_b_diffusion} using the standard Euler-Maruyama discretization ($i \in \{0, I\}$ is the discrete time index): 
\begin{equation}
\label{eq:usb_algo_q_paths}
    \Delta \fvers{X}_i = (\drift + \diffusivity \Znn{F}) \Delta t + \diffusivity \Delta W_i
\end{equation}
and
\begin{equation}
\label{eq:usb_algo_n_paths}
    \Delta \bvers{X}_i = (\drift - \diffusivity \Znn{B}) \Delta t + \diffusivity \Delta W_i
\end{equation}

The quantities $\Delta \Y{B}$ and $\Delta \Y{F}$ are obtained analogously by discretizing Eqs. (\ref{eq:usb_sde_y_f}.2) and (\ref{eq:usb_sde_y_b}.1).

Transitions to and from the coffin state are, instead, simulated using the technique of \textit{shadow paths}. The exact mechanism is provided in \cref{alg:usb_sampling}.

\section{Related Work}
\label{app:sec:related-work}

\paragraph{Unbalanced optimal transport.} We start by briefly recalling two different approaches to unbalanced OT.  First, the \emph{hard} marginal constraints can be relaxed and replaced by \emph{soft} ones. We refer to \citet{chizat2018interpolating,liero2018optimal,yang2018scalable,kondratyev2016new}
and the references therein. Second, the measures of interest can be extended to
satisfy the \emph{conservation of mass} constraint. This is usually done by
adding a \emph{coffin} state $\{\infty\}$. After this operation, it is possible
to apply standard (although defined on this extended space) tools from OT
\citep{pele2009fast,caffarelli2010free,gramfort2015fast,ekeland2010existence}.
These works also include entropic relaxations of the OT objective. Moving to the
dynamic and static formulations of unbalanced \aclp{SB},
\citet{chen2022most} have obtained the main properties of the Schr\"odinger
Bridge in the unbalanced setting. To the best of our knowledge, our study of the
iterates of the Iterate Proportional Fitting procedure is new.

\paragraph{Diffusion models.} The time reversal of the continuous part
of the dynamics has been investigated in
\citet{anderson1982reverse,haussmann1986time,cattiaux2021time}. It is at the
basis of diffusion models \citep{ho2020denoising,song2021maximum}. Such
approaches represent the state-of-the-art generative models in many domains,
from text-to-image generation \citep{saharia2022photorealistic} to protein
modeling \citep{watson2022broadly}. To the best of our knowledge, no existing
diffusion model includes birth and death mechanisms in the forward and backward
processes. Hence, the results we obtain are also new from a diffusion model
perspective. We highlight that the updates we obtain on the birth and death rates
resemble closely the update obtained for the time-reversal of discrete
state-space diffusion models such as \citet{shi2023diffusion}. We refer to
\citet{benton2022denoising} for a treatment of general diffusion models through
the lens of the infinitesimal generator. An alternative approach to our derivations
would have been to use these results to establish
\begin{enumerate*}[label=(\roman*)]
\item a variational lower-bound defining a loss function,
\item a time-reversal formula.
\end{enumerate*}
We leave such a study for future work.

\paragraph{Diffusion Schr\"odinger bridges and extensions.}
The Diffusion Schr\"odinger Bridge (DSB)
\citep{de2021diffusion,vargas2021solving,chen2021likelihood} is a new paradigm to
solve transport problems efficiently in high dimensions. It relies on the advent
of diffusion models. Several extensions of \acp{DSB} have been proposed. Closest to
our work is \citet{liu2022deep}, which studies a \emph{generalized} version of the
\acs{SB} problem. In that case, an extra functional is added to the
quadratic control corresponding to the \acl{SB}. Even though the
updated equations share some similarities with ours, we highlight some key differences:
\begin{enumerate*}[label=(\roman*)]
\item there is an extra term in the Partial Differential Equation system we consider, which can be identified with the fact that we consider \emph{death},
\item the dynamics in \citet{liu2022deep} do not include birth and death
  mechanisms. As a result, the obtained procedure corresponds to a numerical
  scheme for \citet[Section 5]{chen2022most}, which describes a \emph{reweighted}
  \acs{SB}.
\end{enumerate*}
While the reweighted approach can also account for the loss of mass, it does not
update the particles in a meaningful way from a transport point of view, since
the death and birth rate are never updated. Theoretical properties of the
solutions to Schr\"odinger bridge problems are investigated in
\citet{chen2022most}. Finally, we highlight that recent approaches \citep{somnath2023aligned,liu20232,shi2023diffusion} introduce
new schemes to solve \acp{SB} based on the ideas developed
in \citet{lipman2022flow}.

\section{Algorithmic Details}
\label{app:sec:algorithmic_details}

In this section, we clarify the design of our algorithm \usbferryman{}, which we introduce to mitigate the issues of \usbalgo{}. We present an approximation to the computation of the posterior killing rate which is conducive to stability and scalability. We highlight the changes to the training and sampling procedures that this approximation entails and how it allows us to tackle forward dynamics with both killing and births.

\subsection{Estimating Extremal values of Log-Potentials}

Sampling procedures in \cref{alg:usb_sampling} iteratively compute the values of $\Y{F}$ and $\Y{B}$ by starting from one extreme (either $\Y{B}_0$ or $\Y{F}_I$) and then adding (or subtracting) the increments $\Delta \Y{F}$ and $\Delta \Y{B}$ at each time-step. The computation of $\Y{B}_0$ or $\Y{F}_I$, however, requires some care.
These extreme values can be approximated by:
\begin{align}
    \Y{B}_0 = \log \frac{\marg{F}}{\densfact{F}_0} \approx \log\marg{F} - \Ynn{F}_0 \\
    \Y{F}_I = \log \frac{\marg{B}}{\densfact{B}_I} \approx \log\marg{B} - \Ynn{B}_I,
\end{align}
but to evaluate these expressions we need access to the log-densities $\log\marg{F}$ and $\log\marg{B}$.
Obviously, we should not rely on the availability of exact, closed-form expressions, since that would make our algorithm not applicable to empirical data. We, therefore, resort to coarse approximations computed once before starting the training. We can, for instance, fit Bayesian Gaussian Mixture models but different solutions may be preferable when working with specific types of data.

 We note that the need for estimates of marginal densities is a burdensome requirement, which is not placed by solvers of balanced SBs and its removal motivates our algorithm \usbferryman{}.

\begin{algorithm}[h!]
    \caption{\usbalgo{} sampling (death forward process)}
    \label{alg:usb_sampling}
    \begin{algorithmic}[1]
        \item[] \textbf{Input:} drift $\drift$, diffusivity $\diffusivity$, initial mass $M_0$, final mass $M_1$, killing function $\killing$
        \item[]
        \Function{Sample-Forward}{$\sfYnn, \sbYnn, \Psi$}
            \State Sample position $X_0 \sim \marg{F}$
            \State $A_0 \gets 1$ \Comment{The particle is initially alive}
            \State $\Y{B}_0 \gets$ \Call{Compute-Initial-y-hat}{$\sfYnn, \sbYnn$}
            \For{step $i$ in $\{1, ..., I\}$}
                \State $X_{i} \gets X_{i-1} + \Delta \fvers{X_i}$
                \State $\Y{B}_{i} \gets \Y{B}_{i-1} + \Delta \Y{B}_i$
                \If{$A_{i-1}=1$} \Comment{The particle is alive}
                    \State Flip coin $D \sim \text{Bernoulli}\left(1 - \Psi \frac{\killing(i, X_i)}{\exp{\sfYnn(i, X_{i})}} \Delta t\right)$
                    \State $A_{i+1} \gets D$ \Comment{Store particle's fate}
                \EndIf
            \EndFor
            \State \Return $(X_i, \Y{B}_i, A_i)_i$
        \EndFunction
        
        \State
        \Function{Sample-Backward}{$\sfYnn, \sbYnn, \Psi$}
            \State Sample position $\X{F}_I \sim \marg{B}$
            \State Sample status $A_I \sim \text{Bernoulli}\left(\min\left(1, \frac{M_1}{M_0}\right)\right)$ \Comment{Is the particle alive at $k=K$?}
            \State $\Y{F}_I \gets$ \Call{Compute-Initial-y}{$\sfYnn, \sbYnn$}
            \For{step $i$ in $\{I-1, ..., 0\}$}
                \State $\X{F}_{i} \gets \X{F}_{i+1} - \Delta \bvers{\X{F}}_i$
                \State $Y_{i} \gets Y_{i+1} - \Delta Y_i$
                \If{$A_{i+1}=0$} \Comment{The particle is dead}
                    \State $R \gets$ Fraction of dead particles at time $k+1$: 
                    \State Flip coin $D \sim \text{Bernoulli}\left(\Psi\frac{\killing(i, X_i)}{R \exp\sfYnn(i, X_{i})} \Delta t\right)$
                    \State $A_{i} \gets D$ \Comment{Store particle's fate}
                \EndIf
            \EndFor
            \State \Return $(\X{F}_i, Y_i, A_i)_i$
        \EndFunction
    \end{algorithmic}
\end{algorithm}

\subsection{Training Objectives}
In this section, we explicitly state the training objectives used by \usbalgo{}.

\subsubsection{MM Losses}
\paragraph{Mean-Matching Objective.}
It is easy to adapt the \textit{mean-matching} objective~\citep{de2021diffusion} to UDSBs since it is computed from paths of live particles --which follow a standard diffusion process-- and is therefore clearly independent from deaths. We use the following first-order approximation (in $\Delta t$) of its original expression ($\tilde{b}$ represents indifferently any forward or backward drift):
\begin{align}
    \mathcal{L}_{\text{mm}} &\approx \Delta t\sum_i \evd{X_i}{\norm{( X_{i+1}-X_i) - \tilde{b} \Delta t}^2}.
\end{align}
When evaluated on forward $(\fvers{X}_i)_i$ and backward $(\bvers{X}_i)_i$ paths, this loss becomes: 
\begin{align}
    \mathcal{L}_{\text{mm}}((\bvers{X}_i)_i; \params{F}) &= \Delta t \sum_i \,\norm{\left( \bvers{X}_{i+1}-\bvers{X}_i\right) - (b + \diffusivity^2\gradx\Ynn{F}) \Delta t}^2 \\
    \mathcal{L}_{\text{mm}}((\fvers{X}_i)_i; \params{B}) &= \Delta t \sum_i \,\norm{\left( \fvers{X}_{i+1}-\fvers{X}_i \right) - (b - \diffusivity^2\gradx\Ynn{B}) \Delta t}^2.
\end{align}
Minimizing these quantities allows, in our case, to learn the forward/backward trajectories of particles that do not jump to the coffin state. However, it is not sufficient to fully characterize the magnitude of $\Ynn{F}$ and $\Ynn{B}$, since only their gradients appear in it. $\mathcal{L}_{\text{mm}}$ is in fact invariant to constant shifts: i.e., there is no difference between $\Ynn{F}$ and $(\Ynn{F}-c)$ during training.

\paragraph{Divergence Objective.}
An alternative $\mathcal{L}_\text{MM}$ candidate is given by the \textit{divergence-based objective} $\mathcal{L}_{\text{div}}$ by~\citet{chen2021likelihood}. Its original formulation in the context of balanced \acsp{SB} depends only on the scores $\Z{F}$ and $\Z{B}$.
Surprisingly, the same is not true for \acsp{USB} since this objective is affected by the presence of deaths. We extend it to the unbalanced case by recalling~\citep{liu2022deep} that $\mathcal{L}_{\text{div}}$ can be also expressed as:
\begin{align}
\label{eq:usb_ipf_abstract}
    \mathcal{L}_{\text{div}}(\params{F}) &= \evd{X_s \sim \bvers{X}}{\int_0^1 {\left(\diff{\Ynn{F}_s}(X_s) + \diff{\Ynn{B}_s}(X_s) \right)}} \\
    \mathcal{L}_{\text{div}}(\params{B}) &= \evd{X_t \sim \fvers{X}}{\int_0^1 {\left(\diff{\Ynn{F}_t}(X_t) + \diff{\Ynn{B}_t}(X_t)\right)}},
\end{align}
where the first expectation runs over the backward dynamics (from $N$) while the second uses the forward dynamics (from $Q$).
The terms related to jumps that appear in the differentials of $\Y{B}$ and $\Y{F}$ (Eqs. \ref{eq:usb_sde_y_f} and \ref{eq:usb_sde_y_b}) do not completely cancel out when summed (i.e., the term $\killing\discrfact{F}$ from $\diff{\Y{F}}$ remains). The objective $\mathcal{L}_{\text{div}}(\params{F})$, computed as the sum of the equations in \eqref{eq:usb_sde_y_b}, is therefore:
\begin{equation}
\label{eq:usb_ipf_f}
    \begin{split}
        \mathcal{L}_{\text{div}}(\params{F}) = \int_0^1 &\evd{\X{F}_s \sim \bvers{X}}{\frac{1}{2}\norm{\Znn{F}_s(\X{F}_s)}^2 + \diffusivity \divx\Znn{F}_s(\X{F}_s) + \ip{\Znn{F}_s(\X{F}_s)}{\Znn{B}_s(\X{F}_s)} - {\color{MyRed} \frac{\killing\Psi}{e^{\Ynn{F}_s(\X{F}_s)}}}}\diff{s},
    \end{split}
\end{equation}
in which the term in \textcolor{MyRed}{blue} is unique to \acsp{USB}. Note that this new death-related term depends on the parameter that is being optimized ($\params{F}$) and cannot, therefore, be ignored when training the network $\Ynn{F}$. The same term also appears in the objective $\mathcal{L}_{\text{div}}(\params{B})$, which is obtained by summing the differentials in \eqref{eq:usb_sde_y_f}:
\begin{equation}
\label{eq:usb_ipf_b}
    \begin{split}
        \mathcal{L}_{\text{div}}(\params{B}) = \int_0^1 &\evd{\X{F}_t \sim \fvers{X}}{\frac{1}{2}\norm{\Znn{B}_t(\X{F}_t)}^2 + \diffusivity \divx\Znn{B}_t(\X{F}_t) + \ip{\Znn{B}_t(\X{F}_t)}{\Znn{F}_t(\X{F}_t)} - {\color{MyRed} \frac{\killing\Psi}{e^{\Ynn{F}_{1-t}(\X{F}_t)}}}}\diff{t}.
    \end{split}
\end{equation}
In this case, however, the \textcolor{MyRed}{blue} quantity is irrelevant since it does not depend on $\params{B}$. By discretizing the two expressions for $\mathcal{L}_{\text{div}}$ and removing unnecessary terms we obtain the losses:
\begin{equation}
\label{eq:usb_algo_div_loss}
    \begin{split}
        \mathcal{L}_{\text{div}}\left((\bvers{X}_i)_i; \params{F}\right) &= \Delta t \sum_i \Biggl( \frac{1}{2}\norm{\Znn{F}_i(\bvers{X}_i)}^2 + \diffusivity \divx\Znn{F}_i(\bvers{X}_i) \\
        &\qquad \qquad \qquad + \ip{\Znn{F}_i(\bvers{X}_i)}{\Znn{B}_i(\bvers{X}_i)} - {\color{MyRed} \frac{\killing\Psi}{e^{\Ynn{F}_i(\bvers{X}_i)}}} \Biggr) \\
        \mathcal{L}_{\text{div}}\left((\fvers{X}_i)_i; \params{B}\right) &= \Delta t \sum_i \left(\frac{1}{2}\norm{\Znn{B}_i(\fvers{X}_i)}^2 + \diffusivity \divx\Znn{B}_i(\fvers{X}_i) + \ip{\Znn{B}_i(\fvers{X}_i)}{\Znn{F}_i(\fvers{X}_i)} \right).
    \end{split}
\end{equation}

\paragraph{Choosing the MM loss.}
Either one of the losses $\mathcal{L}_\text{mm}$ and $\mathcal{L}_\text{div}$ presented above can be used as $\mathcal{L}_\text{MM}$ in \cref{alg:usb_training}. However, the choice determines the type of admissible reference drifts --which must be linear if the mean-matching loss $\mathcal{L}_\text{mm}$ is selected-- and the computational cost of training --which is higher for $\mathcal{L}_\text{div}$ due to the computation of the divergence.

\subsubsection{TD Loss}
While sufficient to ensure that $\gradx \Ynn{F}$ and $\gradx \Ynn{B}$ point ``in the right direction'', $\mathcal{L}_\text{MM}$ objectives described in the previous section do not constrain the magnitude of $\Ynn{B}$ (which always appears as $\Znn{B}=\diffusivity\nabla\Ynn{B}$). Furthermore, the only objective ($\mathcal{L}_{\text{div}}(\theta)$) which depends on the magnitude of $\Ynn{F}$ is likely insufficient to learn it in practice: the \textcolor{MyRed}{blue} term in \eqref{eq:usb_algo_div_loss} disappears when the prior killing rate is 0 or when $\discrfact{F}$ becomes exceedingly small, leading to a very weak signal on $\Ynn{F}$.
This motivates our quest for a second type of loss ($\mathcal{L}_\text{TD}$) which allows learning the magnitudes of $\Ynn{F}$ and $\Ynn{B}$. 

We adopt the TD loss introduced by~[Liu Deep]. Its expressions for both networks are given by:
\begin{align}
    \mathcal{L}_{\text{TD}}\left(\params{F}\right) &= \int_0^1 \evd{X_s\sim\bvers{X}}{\lvert\Ynn{F}_s(X_s) - \Y{F}_s \rvert}\diff{s} \\
    \mathcal{L}_{\text{TD}}\left(\params{B}\right) &= \int_0^1 \evd{X_t\sim\fvers{X}}{\lvert \Ynn{B}_t(\X{F}_t) - \Y{B}_t \rvert} \diff{t},
\end{align}
where $(\Y{F}_s)_s$ is computed using \eqref{eq:usb_sde_y_b} and $(\Y{B}_t)_t$ using \eqref{eq:usb_sde_y_f}. In words, the TD loss measures the L1 distance between the values of $\Y{F}$ and $\Y{B}$ --which we compute on paths from $N$ and $Q$ respectively-- and the output of neural networks $\Ynn{F}$ and $\Ynn{B}$.

\subsection{Computing \texorpdfstring{$\Psi$}{Psi}}
The previous section describes the losses designed to optimize $\Ynn{F}$ and $\Ynn{B}$ and we now describe how to update $\discrfact{F}$, which is the third quantity estimated by \usbalgo{} (being required to compute the posterior killing rate $\discrfact{F} \killing/\densfact{F}$). Its value is refreshed at the beginning of every training epoch (line~\ref{alg:line:psi_update} of \cref{alg:usb_training}) by \textsc{Update-Psi()}. $\discrfact{F}$ can be computed from the closed-form expression:
\begin{equation}
\label{eq:usb_psi_continuous}
    \discrfact{F} = \left( \int_{\mathcal{X}} \marg{F} \diff{x} - \int_{\mathcal{X}} \marg{B} \diff{x} \right) \left( \int_0^1 \diff{t} \int_{\real^d} \diff{x} \;\killing\densfact{B}(t,x) \right)^{-1}
\end{equation}
which directly follows from the Schr\"odinger Problem [Chen Unbalanced]. We use the following two approximations.

\paragraph{Approximation 1}
The first approximation is the discrete counterpart of the update formula \eqref{eq:usb_psi_continuous} derived from the \schrodinger{} system, where the integration over $\rset^d$ is replaced by a sum over paths $(X_i)_i$:
\begin{equation}
     \discrfact{F} \approx \left( m_0 - m_1 \right) \left( \sum_i \Delta t \sum_n \killing(i\Delta t, X^n_i) \exp{\Ynn{B}(i\Delta t, X^n_i)} \right)^{-1}.
\end{equation}
In the above formula, the superscript $n$ indexes members of the batch.

\paragraph{Approximation 2}
From \eqref{eq:usb_psi_continuous}, by multiplying and dividing by $\densfact{F}$ inside the integral and then applying the decomposition $\densfact{F}_t\densfact{B}_t = \sbproc_t$, we obtain instead:
\begin{equation}
    \begin{split}
         \discrfact{F} &= \left( m_0 - m_1 \right) \left( \int_0^1 \diff{t} \int_{\real^d} \diff{x} \; \frac{\killing}{\densfact{F}}\densfact{B}\densfact{F}(t,x) \right)^{-1} \\
         &= \left( m_0 - m_1 \right) \left( \int_0^1 \diff{t} \; \evd{X_t}{\frac{\killing}{\densfact{F}}(t, X_t)} \right)^{-1} \\
         &\approx \left( m_0 - m_1 \right) \left( \sum_i \frac{\Delta t}{N} \sum_n \frac{\killing(i\Delta t, X^n_i)}{\exp{\Ynn{F}(i\Delta t, X^n_i)}} \right)^{-1},
    \end{split}
\end{equation}
which uses the network $\Ynn{F}$ rather than $\Ynn{B}$ (which is the one queried by approximation 1).

\subsection{Revising the Update of \texorpdfstring{$\Psi$}{Psi}}
\label{app:sub:revising_psi}
The above formulas to update $\Psi$ are heavily dependent on good guesses of $\sfYnn$ and $\sbYnn$, since they both appear inside exponential terms. In practice, however, reliable estimates may not always be available, for several reasons.
First, the user-defined approximations of $\Y{F}$ and $\Y{B}$ at the extremes may not be good, especially with high dimensional state spaces and complex data distributions.
Second, even if learned approximations of log-potentials $\densfact{F}$ and $\densfact{B}$ improve by the end of training, they can still be inadequate at the beginning.

Unreliable values of $\sfYnn$ and $\sbYnn$ hamper convergence, making it slower or, in the worst case, precluding it. We, therefore, re-parametrize the posterior killing rate $\killing^\star = \killing \Psi/\densfact{F}$ with the help of an additional neural network $\ferrymannn$ as:
\begin{equation}
    \killing^\star \approx \killing(t,x) \ferrymannn_t(x).
\end{equation}
We learn $\ferrymannn$ using a novel loss, called \textit{Ferryman} loss, which we present next.

\paragraph{The Ferryman loss.}
At equilibrium, the amount of mass that reaches the coffin state $\coffin$ is given by:
\begin{equation}
    \int_0^1 \int_{\rset^d} \killing^\star(t,x) p_t(x) \diff{x}\diff{t}
\end{equation}
and should be equal to the difference of mass observed at the marginals $\marg{F}(\rset^d) - \marg{B}(\rset^d)$.
By matching these two expressions, we get the loss:
\begin{equation}
\label{eq:ferryman_death_naive}
    \mathcal{L}(\zeta) = \int_0^1\ev{k(\bfX_t)g_{\zeta,t}(\bfX_t)} \rmd t  - \mu_1(\rset^d) + \mu_0(\rset^d),
\end{equation}
where the expectation runs over paths. We stress that this loss does not guarantee the convergence of $\killing \ferrymannn$ to the optimal form of posterior killing rate $\killing \discrfact{F}/\densfact{F}$ but achieves good-enough results in practice.

The loss function \eqref{eq:ferryman_death_naive} tries to capture a mass constraint at the extremes, i.e., for $t=0$ and $t=1$. However, we can extend it to the more general case in which the size of the population $m_t$ is also known at several intermediate times $t_i, i \in I$. This increases the amount of information that \acsp{USB} are able to incorporate and allows learning better dynamics in many real-world applications.

We call $(X_t)_t$ a trajectory and define binary random variables $A_t$ which specify the status of the particle, by taking the value 1 when $X_t \neq \coffin$. For the set of mass measurements $m_t$, it should hold that $\prob{A_t=1} = \evd{A_t}{A_t} = m_t$. This constraint is, however, ill-suited to be directly optimized, since $A_t$ is a discrete value and we should back-propagate through the expected value.
We, therefore, replace it with the quantity $\evd{(X_t, A_t)}{\killing^\ferrymanparams(t, X_t) \mathbf{1}\{A_t = 1\}}$.
Furthermore, we relax the requirement that the prior process is a death-only diffusion and consider a prior birth function $q$, together with the usual killing function $\killing$.

The loss function used to train $\ferrymannn$ then becomes:
\begin{equation}
\label{eq:ferryman_loss}
    \mathcal{L}_F(\ferrymanparams) = \evd{(X_t, A_t)}{ \sum_{i \in I} \left| \int_0^{t_i} \left( \left(1 - A_t\right)q\ferrymannn(t) - A_t\killing\ferrymannn(t) \right)\diff{t} - \frac{m_{t_i} - m_0}{M}\right|},
\end{equation}
where the dependency of $\killing$, $q$ and $\ferrymannn$ on $X_t$ is hidden and $M=\max_i m_{t_i}$ is a normalization constant.
In words, $\mathcal{L}_F$ computes a soft count of deaths and births at each timestep $t$, by summing the probabilities of sampled particles to transition from/to the coffin state. Clearly, these transition probabilities depend on the status $A_t$ of each particle since, e.g., only alive particles can die. For each time interval $[0, t_i]$, $\mathcal{L}_F$ ensures that the approximated change of mass stays close to the observed variation $(m_{t_i}-m_0)/M$.

To use \eqref{eq:ferryman_loss} during training, we should also prevent the value of $\ferrymannn$ from exploding. In our experiments, we, therefore, use a regularized version of this loss, which we illustrate next.

\paragraph{Discretizing the Ferryman loss.}
\label{app:subsec:ferryman_discr}
The loss \eqref{eq:ferryman_loss} may be unstable in practice because $\ferrymannn$ can grow without bounds. This would translate, upon discretization, into invalid transition probabilities $\killing \Delta{t}$ to the coffin state, i.e., probabilities bigger than 1. We, therefore, modify the loss to prevent this. We define the following quantities:
\begin{align}
    \killing_{\rightarrow \infty}(x_n) &= \prob{X_{n+1} = \infty | X_n=x_n} = \killing \ferrymannn(n, x_n) \Delta{t} \\
    \birth_{\leftarrow \infty}(x_n) &= \prob{X_{n+1} = x_n | X_n = \infty} = \birth \ferrymannn(n, x_{n+1}) \Delta{t},
\end{align}
where $(x_n)_n$ is a sampled trajectory and which are, respectively, the conditional probabilities of death $(\killing_{\rightarrow \infty})$ and birth $(\birth_{\leftarrow \infty})$ given the position at step $n$.
The revised Ferryman loss can then be written as:
\begin{equation}
\label{eq:ferryman_loss_discr_reg}
\begin{split}
        \mathcal{L}_F(\ferrymanparams) = \mathbb{E}_{(X_t, A_t)} \Bigg [ \sum_{i \in I} \Bigg | \sum_{n=0}^{n_i} \left( \left(1 - A_n\right) \lceil \birth_{\leftarrow \infty}(X_n) \rfloor - A_n \lceil \killing_{\rightarrow \infty}(X_n) \rfloor \right) - \frac{m_{n_i} - m_0}{M} \Bigg | +\\
        + \sum_{n=0}^{n_i} \Bigl |  \birth_{\leftarrow \infty}(X_n) - \lceil \birth_{\leftarrow \infty}(X_n) \rfloor \Bigr | + \sum_{n=0}^{n_i} \Bigl |  \killing_{\rightarrow \infty}(X_n) - \lceil \killing_{\rightarrow \infty}(X_n) \rfloor \Bigr | \Bigg ],
\end{split}
\end{equation}
where $\lceil x \rfloor$ denotes clipping of $x$ to the unit interval.
\eqref{eq:ferryman_loss_discr_reg} differs in two ways from \eqref{eq:ferryman_loss}: (i) it uses clipped transition probabilities in the first sum and (ii) contains two regularization terms, which penalize values of $\birth_{\leftarrow \infty}$ and $\killing_{\rightarrow \infty}$ bigger than 1.

\subsection{A Revised Algorithm}
We detail the revised training procedure of \usbferryman{} in \cref{alg:usb2_training}. The associated sampling procedures are given in \cref{alg:usb2_sampling} in the specific case of a death-only forward process, to allow a direct comparison with the original sampling in \cref{alg:usb_sampling}.

\begin{algorithm}[h!]
    \caption{\usbferryman{} training}
    \label{alg:usb2_training}
    \begin{algorithmic}[1]
        \item[] \textbf{Input:} $\Znn{F}, \Znn{B}, \ferrymannn$
        \item[] \textbf{Output:} $\params{F}, \params{B}, \ferrymanparams$
        \item[]
        
        \For{epoch $e \in \{0, ..., E\}$}

            \State $( \bvers{X}_i, A_i )_i \gets $ \Call{Sample-Backward-F}{$\sfZnn, \sbZnn, \ferrymannn$}
    
            \While{reuse paths}
                \State $L_\text{MM}(\params{F}) \gets \mathcal{L}_\text{MM}\left((\bvers{X}_i); \params{F}\right)$ 
    
                \State Update $\params{F}$ using $\nabla_\params{F}L_\text{MM}$ \Comment{Train forward score}
            \EndWhile
    
            \State $( \fvers{X}_i, A_i )_i \gets $ \Call{Sample-Forward-F}{$\sfZnn, \sbZnn, \ferrymannn$}
    
            \While{reuse paths}
                \State $L_\text{MM}(\params{B}) \gets \mathcal{L}_\text{MM}\left((\fvers{X}_i); \params{B}\right)$
                
                \State Update $\params{B}$ using $\nabla_\params{B} L_\text{MM}$ \Comment{Train backward score}
                
                \State $L_F(\ferrymanparams) \gets \mathcal{L}_F \left( (\fvers{X}_i, A_i); \ferrymanparams \right)$
                \State Update $\ferrymanparams$ using $\nabla_\ferrymanparams L_F$  \label{alg:line:ferryman_update} \Comment{Train transition function}
            \EndWhile
        \EndFor
    \end{algorithmic}
\end{algorithm}
\vspace{3cm}
\newpage
\begin{algorithm}[H]
    \caption{\usbferryman{} sampling (death-only prior)}
    \label{alg:usb2_sampling}
    \begin{algorithmic}[1]
        \item[] \textbf{Input:} drift $\drift$, diffusivity $\diffusivity$, initial mass $M_0$, final mass $M_1$, killing function $\killing$
        \item[]
        \Function{Sample-Forward-F}{$\sfZnn, \sbZnn, \ferrymannn$}
            \State Sample position $X_0 \sim \marg{F}$
            \State $A_0 \gets 1$ \Comment{The particle is initially alive}
            \For{step $i$ in $\{1, ..., I\}$}
                \State $X_{i} \gets X_{i-1} + \Delta \fvers{X_i}$ \label{alg:line:x_f}
                \If{$A_{i-1}=1$} \Comment{The particle is alive}
                    \State Flip coin $D \sim \text{Bernoulli}\left(1 - \killing(i, X_i) \ferrymannn(i) \Delta t\right)$
                    \State $A_{i+1} \gets D$ \Comment{Store particle's fate}
                \EndIf
            \EndFor
            \State \Return $(X_i, A_i)_i$
        \EndFunction
        
        \State
        \Function{Sample-Backward-F}{$\sfZnn, \sbZnn, \ferrymannn$}
            \State Sample position $\X{F}_I \sim \marg{B}$
            \State Sample status $A_I \sim \text{Bernoulli}\left(\min\left(1, \frac{M_1}{M_0}\right)\right)$ \Comment{Is the particle alive at $k=K$?}
            \For{step $i$ in $\{I-1, ..., 0\}$}
                \State $\X{F}_{i} \gets \X{F}_{i+1} - \Delta \bvers{\X{F}}_i$ \label{alg:line:x_b}
                \If{$A_{i+1}=0$} \Comment{The particle is dead}
                    \State Flip coin $D \sim \text{Bernoulli}\left(\killing(i, X_i) \ferrymannn(i) \Delta t\right)$
                    \State $A_{i} \gets D$ \Comment{Store particle's fate}
                \EndIf
            \EndFor
            \State \Return $(\X{F}_i, A_i)_i$
        \EndFunction
    \end{algorithmic}
\end{algorithm}

\section{Experiments}
\label{app:sec:experiment_details}
In this section, we provide further details on the model architecture, training parameters, and metrics used. We also discuss two additional experiments:
The first (§\ref{app:subsub:alg_comparison}) compares the performances of the two \ac{USB} solvers which we in \cref{sec:algo}.
The second (§\ref{app:sub:covid_experiment}) involves modeling the emergence of the Delta variant during the COVID pandemic.

\subsection{Toy Experiments}
This section contains details about the experiments which involve synthetically generated datasets.

\subsubsection{Comparison between UDSB-TD and UDSB-T}
\label{app:subsub:alg_comparison}
\begin{figure*}[t]
    \centering
    \includegraphics[width=\textwidth]{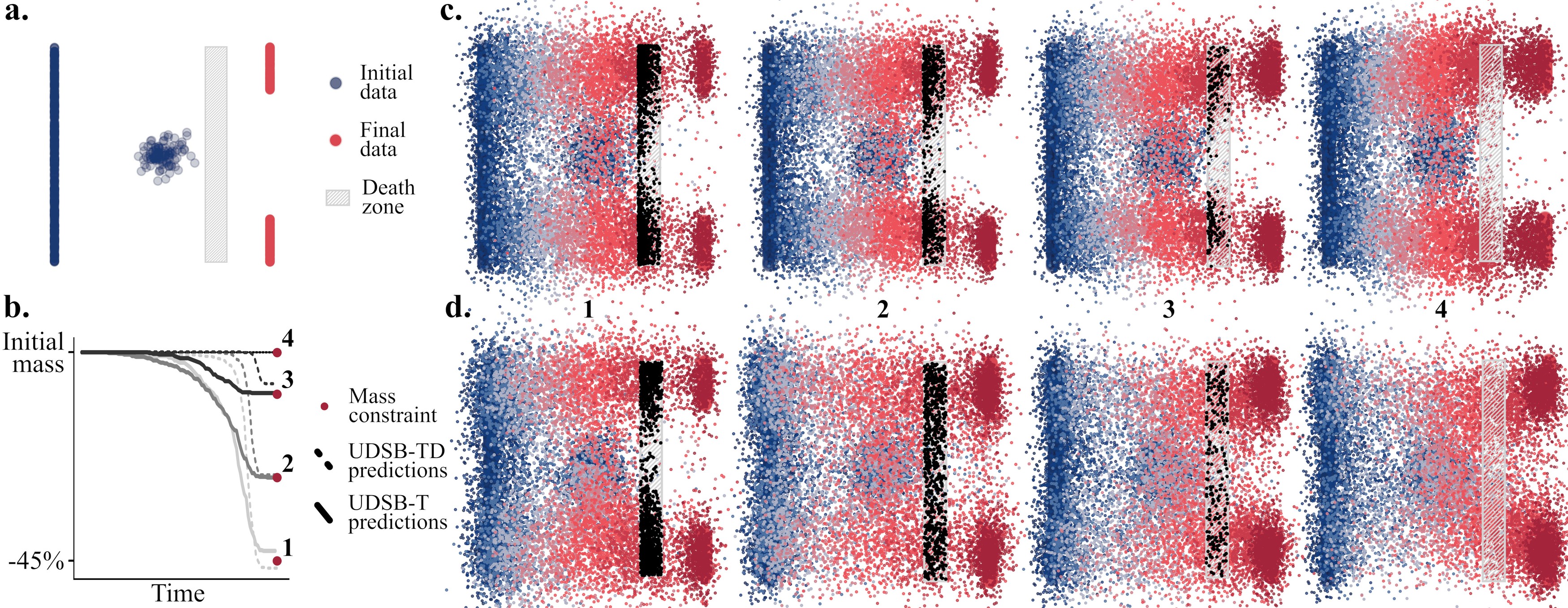}
    \caption{Comparison between \usbalgo{} and \usbferryman{}. (\textbf{a}) Points, initially distributed according to the \textit{blue} marginal, move towards the \textit{red} one. In doing so, they cross a region with non-zero killing probability (\textit{gray} rectangle). Assuming 4 different amounts of end mass (1-4), we use (\textbf{c}) \usbalgo{} and (\textbf{d}) \usbferryman{} to compute particle paths. (\textbf{b}) The algorithms find similar, but not identical, trajectories, which respect the end mass constraints in all cases. However, the predicted amounts of live particles at intermediate times differ, owing to the different updates of $\Psi$ used: deaths are clustered by \usbalgo{} at the top and bottom of the killing zone and appear, instead, more uniformly distributed in (\textbf{d}).}
    \label{fig:results_comparison}
\end{figure*}

We aim to test the ability of our two solvers, \usbalgo{} and \usbferryman{}, to compute valid \acp{SB} while respecting arbitrary mass constraints.
To this end, we run both algorithms on the 2-dimensional toy dataset displayed in Fig. \ref{fig:results_comparison}a. 

Points are initially drawn from a mixture of (i) a uniform distribution (left, \textit{blue} segment) and (ii) an isotropic Gaussian (center). Their final distribution is instead uniform and supported on two segments (right, in \textit{red}). Given the constraints on their endpoints, particles --which follow a Brownian prior-- should travel from left to right. In the process, however, they cross the area marked with a rectangle in the picture. It denotes a region of the state space in which deaths can occur, i.e., a region in which $k$ is non-zero.

We test our \acs{SB} solvers under multiple mass scenarios, by specifying four different amounts of live particles at the end (Fig. \ref{fig:results_comparison}b). We picture the trajectories found by \usbalgo{} in Fig. \ref{fig:results_comparison}c,  and those computed by \usbferryman{} in Fig. \ref{fig:results_comparison}d. We observe that, in both cases, the paths constitute valid \acp{SB}, i.e., correctly match the marginals. Furthermore, the predicted end distributions are similar in quality in all scenarios (Table \ref{app:tab:results_algo_comp}), and no algorithm consistently achieves better scores. 

The trajectories computed by \usbalgo{} and \usbferryman{} are nevertheless not identical. In particular, differences emerge when comparing the positions of deaths (\textit{black} dots), which are concentrated around the top and bottom ends of the killing region in Fig. \ref{fig:results_comparison}c, while appearing more uniform in Fig. \ref{fig:results_comparison}d. This discrepancy is a direct consequence of different ways of computing $\discrfact{F}$. 

\looseness -1\usbalgo{} updates $\discrfact{F}$ using formulas that depend on log-potentials ($\log \varphi$ and $\log \hat\varphi$) which, as discussed in §\ref{app:sub:revising_psi}, are usually not well-approximated at the beginning.
Inaccurate values of potentials lead, in the initial phase of training, to excessive deaths and therefore push the trajectories away from the death region, i.e., drive particles above or below the rectangle.
When the estimates of $\log \varphi$ and $\log \hat\varphi$ gradually improve, the algorithm readjusts the trajectories to reach the desired amounts of particle deaths. However, the initial steering persists for the particles starting closer to the death zone (central \textit{blue} points in Fig. \ref{fig:results_comparison}a) since their trajectories underwent the most significant distortion.
Thanks to the improved stability of the Ferryman loss, instead, the trajectories computed by \usbferryman{} evolve more gradually during training, and particles are therefore free to cross the killing zone (and die).
\begin{table}
    \centering
        \begin{tabular}{lcccc}
        \toprule
         & \multicolumn{4}{c}{\textbf{Algorithm Comparison}} \\
        \cmidrule(lr){2-5}
         & \multicolumn{2}{c}{MMD $\downarrow$} & \multicolumn{2}{c}{$\text{W}_\varepsilon \downarrow$} \\
         \textbf{Final mass} & \usbalgo{} & \usbferryman{} & \usbalgo{} & \usbferryman{} \\
        \midrule
         $45\%$ & \textbf{8.92e-3} & 12.60e-3 & \textbf{1.33} & 1.70 \\
         $56\%$ & 9.42e-3 & \textbf{6.74e-3} & 1.61 &  \textbf{1.28}\\
         $67\%$ & \textbf{3.28e-3} & 5.65e-3 & 1.14 & \textbf{1.08} \\
         $78\%$ & \textbf{5.78e-3} & 6.58e-3 & 1.52 & \textbf{1.28} \\
         $89\%$ & \textbf{5.43e-3} & 8.03e-3 & \textbf{0.96} &  1.53 \\
         $100\%$ & \textbf{1.74e-3} & 5.96e-3 & 1.24 & \textbf{1.06} \\
         \bottomrule \\
        \end{tabular}
        \caption{\looseness -1\textbf{Prediction quality on toy dataset.} MMD and Wasserstein distance between the predicted end marginal and the ground truth. The two algorithms \usbalgo{} and \usbferryman{} perform comparably, with the former often achieving better MMD scores and the latter faring better with $W_\varepsilon$.}
        \label{app:tab:results_algo_comp}
\end{table}

\subsubsection{Models and Training}
All models used in experiments on (low-dimensional) synthetic datasets share similar architectures and training procedures.

The architecture of networks $f^\params{F}$ and $\hat{f}^\params{B}$ consists of:
\begin{itemize}
    \item \texttt{x\_encoder}: a 3-layer MLP, with $32$-dimension wide hidden layers, which takes points in the state space as input;
    \item \texttt{t\_encoder}: a 3-layer MLP, with $32$ hidden dimensions, which takes the sinusoidal embedding (over 16 dimensions) of the time $t$ as input;
    \item \texttt{net}: a 3-layer MLP, with $32$-dimension wide hidden layers, which receives the concatenation of the outputs of the two previous modules as input and outputs the score.
\end{itemize}
All the above networks use the \texttt{SiLU} activation function.

We parametrize $\ferrymannn$ with a 5-layer MLP with $64$ hidden dimensions and \texttt{Leaky-ReLU} as non-linearity.

We run 15 iterations of our unbalanced IPF algorithm and update weights using the \textsc{AdamW} optimizer with gradient clipping. We set the initial learning rate to 1e-3 for $f$ and $\hat{f}$, and to 1e-2 for $q$.

\subsection{Cell Drug Response}
We further comment on the pre-processing of the cell evolution dataset and provide details on the model architecture and training procedure.

\subsubsection{Dataset}
We start with the dataset collected by \citet{bunne2021learning}, which captures the temporal evolution of melanoma cells treated with a mix of the cancer drugs \textit{Trametinib} and \textit{Erlotinib}.
The drug is given at time $t=0$, and the first measurement, which happens at time $t=8h$, examines a population of 2452 cells.
Each cell is characterized by 78 features which are a combination of \textit{morphological} features derived from microscopy (such as cell shapes) and detailed
information on the abundance and location of proteins –-obtained via the powerful Iterative Indirect Immunofluorescence Imaging (4i) technique \citep{bunne2021learning}-- and which we refer to as \textit{intensity} features.
Intensity features can be, in turn, categorized as measuring either the sum or average intensity. We remove the 28 features belonging to the former group and are, therefore, left with a 50-dimensional state space. We split the dataset into training and test sets according to an 80/20 split.

Two subsequent measurements of cells in the population take place at times $t=24h$ and $t=48h$. Each captures a different population, owing to cell death and birth. In particular (see Fig. \ref{fig:results_cell}d), the population is found to have grown by $35\%$ after $24h$ while shrinking down to $75\%$ of its original size at time $t=48h$.

To compute the unbalanced \ac{DSB} in Fig. \ref{fig:results_cell}e, we assume a Brownian prior motion of cells in the feature space. Furthermore, we consider a prior killing rate proportional to the distance from the first-order spline interpolation ($g$) of the empirical means of the train set. More precisely, this killing function penalizes cell statuses that deviate substantially ($> 2\sigma$) from $g$ in more than $20\%$ of features.
The birth rate is, instead, proportional to a Gaussian \acf{KDE} computed on the train set.

\subsubsection{Model and Training}
TThe algorithm \usbferryman{} requires 3 networks: $f^\params{F}$ and $\hat{f}^\params{B}$ and $\ferrymannn$.

The architecture of the first two consists of:
\begin{itemize}
    \item \texttt{x\_encoder}: a 3-layer MLP, with $300$-dimension wide hidden layers, which takes the cell status $x$ as input;
    \item \texttt{t\_encoder}: a 3-layer MLP, with $32$ hidden dimensions, which takes the sinusoidal embedding (over 16 dimensions) of the time $t$ as input;
    \item \texttt{net}: a 3-layer MLP, with $300$-dimension wide hidden layers, which receives the concatenation of the outputs of the two previous modules as input and outputs the score.
\end{itemize}
All the above networks use the \texttt{SiLU} activation function.

To parametrize $\ferrymannn$, instead, we use a 5-layer MLP with $64$ hidden dimensions and \texttt{Leaky-ReLU} as non-linearity.

We run 10 iterations of our unbalanced IPF algorithm and update weights using the \textsc{AdamW} optimizer with gradient clipping. We set the initial learning rate to 1e-3 for $f$ and $\hat{f}$, and to 1e-2 for $q$.
We use a batch size of 512.

\subsection{COVID Variants Spread}
\label{app:sub:covid_experiment}
To further test unbalanced \acp{SB} on real-world phenomena, we model the global evolution of the COVID pandemic over a 4-month period. We aim to reconstruct how multiple COVID variants propagate across countries between April 5, 2021, and August 9, 2021. We choose this time window because it encompasses the appearance and quick spread of the Delta variant, at the expense of the once-dominant Alpha variant.

It is important to emphasize that our task focuses on reconstructing the historical trajectory of the pandemic based on its known initial and final statuses. This differs from the more common practice \citep{caoCOVID19ModelingReview2021,nixonEvaluationProspectiveCOVID192022} of predicting how viruses spread solely based on present epidemiological data.
Although our approach may not be directly applicable to guiding policies and public health responses during an ongoing outbreak, it can nonetheless provide valuable insights into the transmission and mutation of a pathogen, when its evolution at intermediate times is not known.

\begin{figure*}[t]
    \centering
    \includegraphics[width=\textwidth]{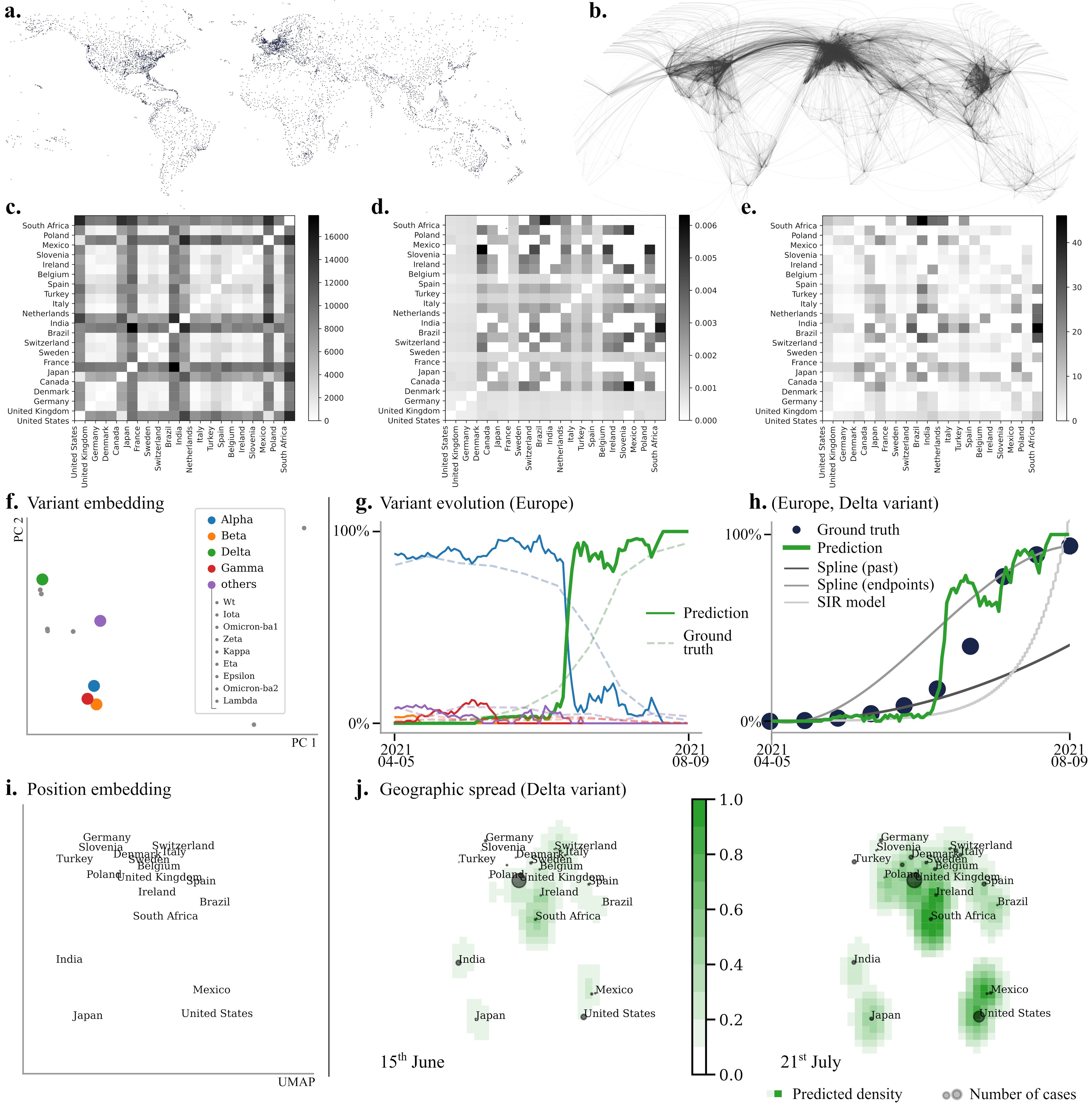}
    \caption{\textbf{COVID evolution experiment}. We model variants spreading across countries, by diffusing over a state space $\mathcal{X}$ endowed with location- and variant-dependent information. We represent (\textbf{i}) the location of COVID cases, by leveraging the notion of \textit{effective} distance between countries. It is based on (\textbf{b}) the number of flights that connect airports (\textbf{a}) in different countries. We compute (\textbf{e}) \textit{effective} distances as the product of (\textbf{c}) \textit{flight} distances $D^F$ and (\textbf{d}) \textit{geographic} distances: for the sake of clarity, we only plot pairwise between the 21 countries with the highest number of COVID cases in 2021. (\textbf{f}) We encode different COVID variants by means of low-dimensional representations of their spike proteins obtained from Protein Language Models. (\textbf{g}) By running \usbferryman{} on $\mathcal{X}$, we can recover the evolution of COVID variants in Europe more accurately than the baselines (\textbf{h}). (\textbf{j}) Furthermore, we approximately reconstruct the geographic distribution of cases around the globe at intermediate times.}
    \label{fig:results_covid}
\end{figure*}

\paragraph{Results.}
By running our algorithm \usbferryman{} on two snapshots of the COVID pandemic, we are able to model it at intermediate times. We can reconstruct the change in variant prevalence both in time and space. We can, for instance, successfully model the proliferation of cases of Delta variant in Europe: Fig. \ref{fig:results_covid}g plots the predictions for each variant against the ground truth, which is only known by our algorithm at the initial and final times. Remarkably, \usbferryman{} reconstructs the spread of the Delta variant better than the baselines (Fig. \ref{fig:results_covid}h) and achieves the smallest Mean Square Error (MSE) with respect to the observations (Table \ref{app:tab:results_covid}). Besides offering variant counts for each continent, our predictions also provide time-resolved density estimates of COVID cases for each country (Fig. \ref{fig:results_covid}j).
\begin{table}
    \centering
        \begin{tabular}{lc}
        \toprule
         \textbf{Method}&{MSE $\downarrow$} \\
        \midrule
         Spline interpolation (past) & 3.07e-1 \\
         Spline interpolation (endpoints) & 1.60e-1 \\
         SIR model & 2.54e-1 \\
         \usbferryman{} & \textbf{1.12e-1} \\
         \bottomrule \\
        \end{tabular}
        \caption{\textbf{Prediction quality on COVID dataset.} MSE between the predicted and observed presence of Delta variant cases in Europe. \usbferryman{} beats the baselines consisting of a SIR infection model and of two cubic spline interpolations, using respectively the first 4 measurements (\textit{past}), and the first and last 2 (\textit{endpoints}).}
        \label{app:tab:results_covid}
\end{table}

\bigskip
We now provide further details regarding the dataset and how we represent the location and variant of COVID cases. 

\subsubsection{Dataset}
We start from bi-weekly records\footnote{Dataset available at: \url{https://www.kaggle.com/datasets/gpreda/covid19-variants}} of sequenced viral samples around the world. 
They provide the number of cases linked to different COVID variants which are found in each country on a given day. 
We only consider samples belonging to the 4 most prevalent variants in our time range, i.e., \textit{Alpha}, \textit{Beta}, \textit{Gamma}, and \textit{Delta}, and represent all the other variant with a fifth category, named \textit{others}.
Each of the datapoints referring to the period between 2021-04-05 and 2021-08-09 therefore contains the following entries:
\begin{itemize}
    \item \texttt{location}: country for which the variants information is provided;
    \item \texttt{date}: the date of the measurement;
    \item \texttt{[variant]}: fraction of sequences belonging to the given variant;
    \item \texttt{num\_sequences}: number of sequences processed in the country.
\end{itemize}
When training our model, we only use the first and last measurements and leave the other ones for testing.

We represent COVID cases that spread among countries as vectors $\mathbf{c}^T = (\mathbf{x}^T; \mathbf{v}^T) \in \mathcal{X}$, in which the first block of coordinates ($\mathbf{x}$) describes the geographic location of the virus and the second one ($\mathbf{v}$) identifies its variant. 
The use of standard \acp{SB} would require that points move \textit{continuously} in $\mathcal{X}$. While we can choose both kinds of embedding to make this assumption approximately true, it is hard to strictly enforce it. This issue emphasizes the need for \acp{USB}, which accounts for jumps in the diffusion. In this case, jumps allow (i) new variants to appear in locations where they were initially not present and (ii) local outbreaks to suddenly disappear.

\paragraph{Location embedding.}
To (approximately) respect the continuity hypothesis, we cannot directly use country coordinates to encode the location $\mathbf{x}$ of viruses. In fact, COVID cases do not preferentially propagate to neighboring countries, and using this representation would then render \acp{SB} a poor fit to model such dynamics.
We, therefore, set to leverage the intuition that the virus tends to spread among countries that are either (i) close to each other or (ii) well-connected, especially via air transportation.

Inspired by the approach pioneered by \citet{brockmann2013hidden}, we compute an \textit{effective} distance between countries, which better represents the ease with which COVID moves between them.

To compute effective distances, we rely on flight data, obtained by OpenFlights\footnote{https://openflights.org}.
Starting with 66934 commercial routes (Fig. \ref{fig:results_covid}b), between 7698 airports (Fig. \ref{fig:results_covid}a), we count the number of flights $C_{ij}$ linking airports of country $i$ with those in country $j$. We first compute the \textit{flight distance} $D^F_{ij}$ between $i$ and $j$ as $D^F_{ij} = 1/(C_{ij} + \epsilon)$ (where $\epsilon$ is a small constant) (Fig. \ref{fig:results_covid}c).
In words, a small flight distance implies the existence of frequent connections between two countries.
By multiplying flight distances $D^F_{ij}$ by the geographic distance between countries (Fig. \ref{fig:results_covid}d), we obtain the matrix $D^E = (D^E_{ij})_{ij}$ of \textit{effective} distances (Fig. \ref{fig:results_covid}e).
$D^E$ is a connectivity matrix on the graph of nations and we can measure the shortest paths ($S_{ij}$) between nodes. We then embed in 2 dimensions the manifold induced by the metric $(S_{ij})_{ij}$ via UMAP \citep{mcinnes2020umap} and therefore obtain a planar representation of the \textit{effective} positions of countries (Fig. \ref{fig:results_covid}i). This layout resembles the world map, e.g., it clusters European nations at the top, but also reflects the ease of traveling between countries, e.g., India is equally far from the US and from the UK.

\paragraph{Variant embedding.}
Having described how we determine the location coordinates $(\mathbf{x})$, we now turn to the representation of COVID variants $(\mathbf{v})$. We aim to capture the biological similarity between variants based on the similarity of corresponding Spike proteins, which are critical in determining their infectiousness.

We focus on the mutations occurring in the receptor-binding domains (RBDs), a key part of the Spike protein.
More specifically, we encode the amino-acid sequences describing RBDs using ESM-2 \citep{lin2023evolutionary}, a state-of-the-art language model optimized for proteins.
When projected along their biggest 2 principal components, these embeddings organize in the plane as shown in Fig. \ref{fig:results_covid}f. We construct the representation of the \textit{other} variant type, by averaging the embeddings of all the variants excluded by our analysis (\textit{gray} dots in Fig. \ref{fig:results_covid}f).
As expected, the Beta variant shares more similarities with Alpha than with Delta, which emerged later in time and features additional mutations in its RBD.

\paragraph{Prior death/birth rates.}
To complete the description of the \acs{SB} problem at hand, we need to specify the prior killing and birth functions.
We assume a positive prior probability that Delta cases emerge in India since early Delta samples were detected in that country. Furthermore, we introduce birthplaces (for all variants) close to every country and killing zones everywhere else. This prior allows \usbferryman{} to strike a balance between spreading the virus by making it move across countries (initial infections) and growing native clusters (local proliferation).

\subsubsection{Model and Training}
We require 3 networks to parametrize the forward and backward scores ($f$ and $\hat{f}$) and the multiplicative reweighting factor of the posterior killing function ($g$).

We use the same model architecture for $f^\params{F}$ and $\hat{f}^\params{B}$, which consists of:
\begin{itemize}
    \item \texttt{x\_encoder}: a 3-layer MLP, with 64-dimensional hidden layers, which takes $c \in \mathcal{X}$ as input;
    \item \texttt{t\_encoder}: a 3-layer MLP, with $32$ hidden dimensions, which takes the sinusoidal embedding (over 16 dimensions) of the time $t$ as input;
    \item \texttt{net}: a 5-layer MLP, with $64$-dimension wide hidden layers, which receives the concatenation of the outputs of the two previous modules as input and outputs the score.
\end{itemize}
All the above networks use the \texttt{SiLU} activation function.

To parametrize $\ferrymannn$, instead, we use a 5-layer MLP with $64$ hidden dimensions and \texttt{Leaky-ReLU} as non-linearity.

We run \usbferryman{} for 10 iterations with a batch size of 1024 and update the weights using the \texttt{AdamW} optimizer, with initial learning rates of 1e-3, for $f$ and $\hat f$, and 1e-2 for $\ferrymannn$.

\paragraph{Variant assignment.}
Given a point $\mathbf{c}^T = (\mathbf{x}^T, \mathbf{v}^T)$ in the statespace $\mathcal{X}$, we map it to one of the five COVID variants by using a 1-Neighbor Classifier on $\mathbf{v}$, from the \texttt{scikit-learn} library \citep{scikit-learn}, trained on the representations in Fig. \ref{fig:results_covid}f.

\paragraph{Baselines.}
To assess the performance of our method, we first restrict to the sub-task of predicting how the Delta variant spreads in Europe (Fig. \ref{fig:results_covid}h), in the time period under consideration. We consider the following 3 different baselines:
\begin{itemize}
    \item Spline interpolation (\textit{past}): a cubic spline interpolation of the first 4 available measurements, i.e., those happening between April 5, 2021 and May 17, 2021. In this period, the Delta variant was not yet widespread in Europe, and, as a consequence, the fitted curve underestimates the number of COVID registered in later months.
    \item Spline interpolation (\textit{endpoints}): a cubic spline interpolation of the first 2 and last 2 known measurements.
    \item SIR model: a popular ODE-based model \citep{kermack1927contribution} of infectious diseases.
\end{itemize}

\subsection{Evaluation Metrics}
In this section, we detail the evaluation metrics used to benchmark our algorithms.

\paragraph{Wasserstein-2 distance.}
We measure accuracy of the predicted target population $\hat{\nu}$ to the observed target population $\nu$ using the entropy-regularized Wasserstein distance \citep{cuturi2013sinkhorn} provided in the \texttt{OTT} library \citep{jax2018github,cuturi2022optimal}.

\paragraph{Maximum mean discrepancy.}
Kernel maximum mean discrepancy~\citep{gretton2012kernel} is another metric to measure distances between distributions, i.e., in our case between predicted population $\hat{\nu}$ and observed one $\nu$.
Given two random variables $x$ and $y$ with distributions $\hat{\nu}$ and $\nu$, and a kernel function $\omega$, \citet{gretton2012kernel} define the squared MMD as:
\begin{equation}
    \text{MMD}(\hat{\nu},\nu; \omega) = \mathbb{E}_{x,x^\prime}[\omega(x, x^\prime)] + \mathbb{E}_{y,y^\prime}[\omega(y, y^\prime)] - 2\mathbb{E}_{x,y}[\omega(x, y)].
\end{equation}
We report an unbiased estimate of $\text{MMD}(\hat{\nu},\nu)$, in which the expectations are evaluated by averages over the population particles in each set. We utilize the RBF kernel, and as is usually done, report the MMD as an average over the length scales: $2, 1, 0.5, 0.1, 0.01, 0.005$.

\section{Code Availability}

The code used in this work can be found at \url{https://github.com/matteopariset/unbalanced_sb}.

\end{document}